# On Reinforcement Learning for the Game of 2048

Student: Hung Guei
Advisor: Dr. I-Chen Wu

A Dissertation
Submitted to
Institute of Computer Science and Engineering
College of Computer Science

National Yang Ming Chiao Tung University
in partial Fulfillment of the Requirements

for the Degree of
Doctor of Philosophy
in
Computer Science

January 10, 2023
Taiwan, Republic of China

# Abstract

2048 is a single-player stochastic puzzle game. This intriguing and addictive game has been popular worldwide and has attracted researchers to develop game-playing programs. Due to its simplicity and complexity, 2048 has become an interesting and challenging platform for evaluating the effectiveness of machine learning methods. This dissertation conducts comprehensive research on reinforcement learning and computer game algorithms for 2048. First, this dissertation proposes optimistic temporal difference learning, which significantly improves the quality of learning by employing optimistic initialization to encourage exploration for 2048. Furthermore, based on this approach, a state-of-the-art program for 2048 is developed, which achieves the highest performance among all learning-based programs, namely an average score of 625377 points and a rate of 72% for reaching 32768-tiles.

Second, this dissertation investigates several techniques related to 2048, including the $n$-tuple network ensemble learning, Monte Carlo tree search, and deep reinforcement learning. These techniques are promising for further improving the performance of the current state-of-the-art program. Finally, this dissertation discusses pedagogical applications related to 2048 by proposing course designs and summarizing the teaching experience. The proposed course designs adopt 2048-like games as materials for beginners to learn reinforcement learning and computer game algorithms. The courses have been successfully applied to graduate-level students and received well by student feedback.

# Contents

# List of Figures

# List of Tables

# Chapter 1    Introduction

*2048* is a single-player stochastic puzzle game introduced by Cirulli [1] as a variant of *Threes!* and *1024*. This intriguing and even addictive game has been popular worldwide since it is non-trivial to master despite the simple rules [2], and it has also attracted researchers to develop game-playing programs [3]. Due to its simplicity and complexity [4], 2048 is considered to be an interesting and challenging platform for evaluating the effectiveness of machine learning methods [5]–[7].

In recent years, many *reinforcement learning (RL)* methods have been proposed for 2048. Szubert and Jaśkowski [2] applied the *temporal difference (TD) learning* with $n$-*tuple networks* to 2048. In their approach, 2048-tiles were achieved at a rate of 97%. Yeh *et al.* [3] introduced *multistage TD (MS-TD) learning* which improved the training by separating an entire episode into several stages. Their 5-stage TD method reached 32768-tiles with a rate of 31.75% and even achieved the first-ever 65536-tile. Matsuzaki [8] presented a systematic analysis of $n$-tuples, identified several well-performed $n$-tuple network configurations. Jaśkowski [5] improved the performance with *temporal coherence (TC) learning*, which accelerated the convergence of the training by adaptively reducing the learning rate. The 16-stage TC method significantly enhanced the reaching rate of 32768-tiles, even up to 70%. Matsuzaki [9] showed that *deep convolutional neural networks (DCNN)* also performed well and demonstrated that DCNNs outperformed $n$-tuple networks when the network sizes are similar. Antonoglou *et al.* extended the well-known *MuZero* [10] to stochastic environments, named *Stochastic MuZero* [11], and achieved results comparable to other high-performance learning algorithms. On the other hand, 2048 is also a good tool for educational purposes [12]. As a teaching tool, 2048's popularity can increase student engagement, while the existing machine learning methods provide a well-established basis for education.

This dissertation focuses on reinforcement learning for the game of 2048. The rest of this chapter is organized as follows. Section 1.1 introduces the rules of 2048. Then, Section 1.2 summarizes the motivation, goals, and achievements of this dissertation. Finally, Section 1.3 describes the organization of the contents.

## 1.1 The Game of 2048

2048[1] is a single-player stochastic slide-and-merge puzzle game with the objective of sliding the puzzle to merge small *tiles* into large tiles to create a *2048-tile*. The game is played on a puzzle with 4×4 cells, starting with two randomly placed tiles. Cells on the puzzle are either empty cells or tiles numbered with powers of 2, such as 2-tiles, 4-tiles, and 65536-tiles.

Whenever the player slides the puzzle up, down, left, or right, all tiles are moved in the chosen direction as far as possible, i.e., until they reach either the border or another tile [2]. Illustrations are shown in Figure 1, such as sliding up from (a) to (b) and sliding left from (c) to (d). Upon sliding the puzzle, two adjacent tiles with the same value in the chosen direction, say both $v$-tiles, will be merged into a single $2v$-tile, and the player receives $2v$ points as the reward. Note that a direction is prohibited if the puzzle remains unchanged after sliding.

After the player slides the puzzle and merges the tiles, the environment randomly adds a new tile on an empty cell, as in Figure 1 (c) from (b). The newly added tile is either a 2-tile or a 4-tile, with probabilities of 0.9 or 0.1, respectively [2]. Then the player continues to slide the puzzle, repeating the above process until there is no possible direction to slide. The player wins the game when achieving a 2048-tile; nevertheless, the game continues until there is no possible direction. The final score of a game is the total cumulative rewards of merging tiles.

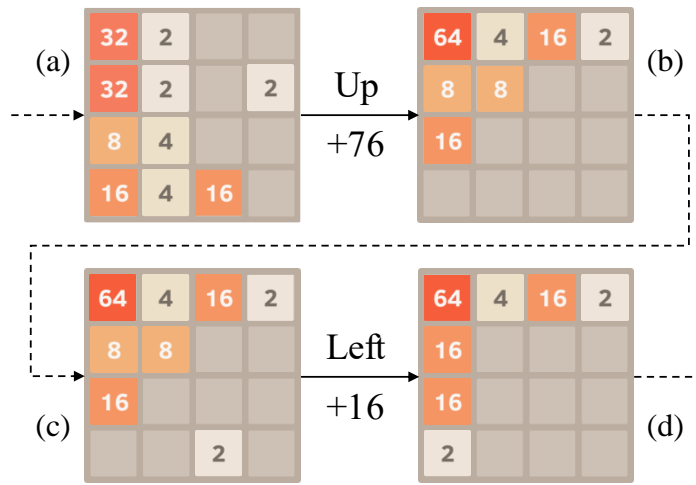

Figure 1. Example of the gameplay of 2048.
The player slides the puzzle up from (a) to (b), merging tiles and receiving 76 points.
After the environment generates a new tile from (b) to (c), the player continues to play by sliding the puzzle left from (c) to (d), which receives 16 points.

---

[1] Available at https://gabrielecirulli.github.io/2048/

## 1.2 Motivation and Goals

Reinforcement learning (RL) is a machine learning method that trains an *agent* to interact with an *environment* to maximize the total outcome [13]. For 2048, the agent continuously slides the puzzle to merge small tiles; the environment provides the reward and then generates a new tile on the puzzle. This dissertation conducts comprehensive research on reinforcement learning and related computer game algorithms for the game 2048. The goal includes but is not limited to developing high-performance programs for 2048, investigating and designing new methods applicable to 2048, and applying 2048 to pedagogical applications.

First, this dissertation proposes *optimistic temporal difference learning* to improve the exploration for 2048. Researchers have tried other well-known exploration techniques for years, such as $\epsilon$-greedy or softmax, while neither works for 2048. Since the training seems to be well, it is hypothesized that the learning process of 2048 does not require additional exploration as the stochastic environment already provides enough for it. For this topic, we propose to use *optimistic initialization* for 2048, which shows that stochastic games such as 2048 still benefit from explicit exploration. Based on the optimistic temporal difference learning with additional tunings such as expectimax search, multistage learning, and tile-downgrading technique, our design achieves state-of-the-art performance, namely an average score of 625377 and a rate of 72% for reaching 32768-tiles. In addition, 65536-tiles are reached at a rate of 0.02%.

Second, this dissertation investigates several methods related to 2048, including *ensemble learning*, *Monte Carlo tree search*, and *deep reinforcement learning*. For ensemble learning, we leverage *stochastic weight averaging* to improve the generalization by averaging multiple $n$-tuple network snapshots, thereby significantly improving performance. For Monte Carlo tree search, we demonstrate that it functions well with $n$-tuple networks and show that it is promising for both testing and training by analyzing different hyperparameters. For deep reinforcement learning, we successfully combine *Gumbel MuZero* and *Stochastic MuZero*, which significantly reduces the training cost of using deep neural networks for 2048. These related methods have shown promise in preliminary experiments and may become essential for the next state-of-the-art of 2048.

Finally, this dissertation also focuses on pedagogical applications. Due to its simplicity and popularity, 2048 and similar games can be good tools for teaching reinforcement learning and computer game algorithms. We summarize the experience of using 2048-like games as teaching materials, which includes course designs for beginners to learn reinforcement learning and computer game algorithms using 2048, 2584, and Threes!. The proposed *lightweight course* guides students to familiarize temporal difference learning in a few hours. Furthermore, the proposed *comprehensive course* guides students to improve their programs while learning new knowledge, with a clear goal of developing a strong AI program. Therefore, the courses not only provide beginners with an intuitive way to learn reinforcement learning and computer games effectively but also motivate them to challenge higher performances.

## 1.3 Organization

This dissertation is organized as follows. Chapter 1 briefly introduces this dissertation. Chapter 2 reviews background techniques for 2048. Chapter 3 presents the optimistic temporal difference learning and the current state-of-the-art 2048 program. Chapter 4 investigates other topics related to 2048 and reinforcement learning. Chapter 5 demonstrates the course designs with 2048-like games as teaching materials. Finally, Chapter 6 summarizes this dissertation and makes concluding remarks.

# Chapter 2    Related Techniques

This chapter reviews the methods and techniques related to 2048 as organized as follows. Sections 2.1 to 2.3 review the foundations and implementations of TD methods. Section 2.4 reviews the $n$-tuple networks. Sections 2.5 and 2.6 review the expectimax search and the Monte Carlo tree search. Section 2.7 reviews Stochastic MuZero. Section 2.8 reviews the bitboard design.

## 2.1   Reinforcement Learning

*Reinforcement learning (RL)* is a machine learning method that trains an *agent* on how to respond to an *environment* with the objective of maximizing the total outcome [13]. As illustrated in Figure 2, the agent continuously interacts with the environment by performing the *actions* $a_t$ in the current *states* $s_t$, and the environment responds to actions by providing the corresponding *rewards* $r_t$ and the new states $s_{t+1}$.

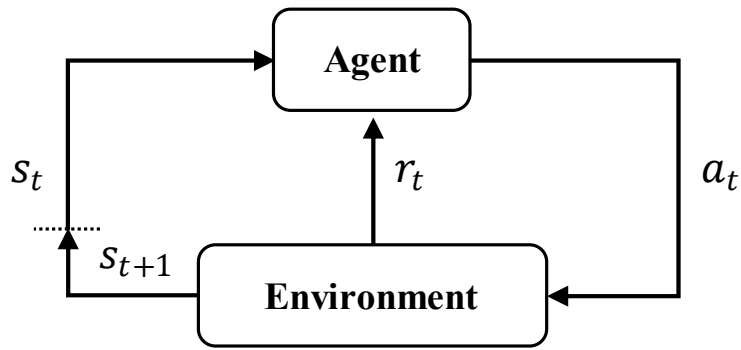

Figure 2. The framework of reinforcement learning.

*Markov decision process (MDP)* is a mathematical framework for decision-making problems, which is commonly used in reinforcement learning [13]. An MDP is constructed by $\langle \mathcal{S}, \mathcal{A}, \mathcal{P}, \mathcal{R} \rangle$, where $\mathcal{S}$ denotes the finite set of states; $\mathcal{A}$ denotes the finite set of actions of the state; $\mathcal{P}: \mathcal{S} \times \mathcal{A} \to \mathcal{S}$ denotes the state transition function; and $\mathcal{R}: \mathcal{S} \times \mathcal{A} \to \mathbb{R}$ denotes the immediate reward function. The MDP models problems of how an agent interacts with the environment through a sequence of actions with respect to states and rewards.

The game of 2048 can be well modeled as an MDP, in which the player is considered an agent who takes actions to the states and receives rewards from the environment. For example, the puzzles illustrated in Figure 1 (a), (b), (c), (d) can be expressed as $s_t$, $s'_t$, $s_{t+1}$, $s'_{t+1}$ respectively, in an episode from $s_0$ to $s_T$, as follows.

$$s_0 \cdots \dashrightarrow s_t \xrightarrow[r_t]{a_t} s'_t \dashrightarrow s_{t+1} \xrightarrow[r_{t+1}]{a_{t+1}} s'_{t+1} \dashrightarrow \cdots s_T \qquad (1)$$

The episode starts with an *initial state* $s_0 \in \mathcal{S}$. At steps $t$, the agent performs *actions* $a_t \in \mathcal{A}(s_t)$ on *states* $s_t \in \mathcal{S}$ to transform $s_t$ into *afterstates* $s'_t = \psi(s_t, a_t)$, where $\psi$ denotes the transition function from states to afterstates, i.e., slides the puzzle. The environment responds *rewards* $r_t = \mathcal{R}(s_t, a_t)$, and then changes $s'_t$ to next states $s_{t+1} \sim \mathcal{P}(s'_t, s_{t+1})$, i.e., adds a new tile. This process repeats until a *terminal state* $s_T \in \mathcal{S}$ that $\mathcal{A}(s_T) = \emptyset$ is reached, where $T$ refers to the end of the episode.

The objective of the problem of MDPs is to find a *policy* $\pi$ that decides which action to take for any given state and maximizes the cumulative rewards [13]. The *state value function* is defined as $V(s_t) = \mathbb{E}[r_t + r_{t+1} + \cdots]$. Therefore, the policy $\pi : \mathcal{S} \to \mathcal{A}$ can be derived from the function as

$$\pi(s_t) = \underset{a_t}{\mathrm{argmax}} \left( r_t + \sum_{\forall s_{t+1}} \mathcal{P}(s'_t, s_{t+1}) V(s_{t+1}) \right), \qquad (2)$$

where $a_t \in \mathcal{A}(s_t)$ and $s'_t = \psi(s_t, a_t)$.

## 2.2 Temporal Difference Learning

*Temporal difference (TD) learning* is a kind of reinforcement learning method that adjusts the state estimations based in part on other learned estimations [13]. This method has been widely applied to many game-playing programs [14]–[21] and was first applied to 2048 by Szubert and Jaśkowski [2], resulting in the first-ever learning-based program that can reach 2048-tiles.

*TD(0)* is the simplest form of TD learning that adjusts the estimation with only one subsequent reward and estimation [13]. After the agent performs an action and receives a reward $r_t$, the environment will provide the next state $s_{t+1}$ whose value is $V(s_{t+1})$. Thus, $r_t + V(s_{t+1})$, known as the *TD target*, is an estimation of $V(s_t)$. Therefore, the estimation error for $s_t$, called the *TD error*, is calculated as

$$\delta_t = r_t + V(s_{t+1}) - V(s_t). \tag{3}$$

Note that $V(s_{t+1})$ is weighted by a *discount factor* $\gamma$ [13] that is disregarded for simplicity, i.e., $\delta_t = r_t + \gamma V(s_{t+1}) - V(s_t)$ where $\gamma = 1$. Then, the estimation of $s_t$, $V(s_t)$, is adjusted with the TD error and the learning rate parameter $\alpha$ as

$$V(s_t) \leftarrow V(s_t) + \alpha \delta_t. \tag{4}$$

TD(0) adjusts the estimation at the current step based on the current reward and the next estimation. Its general form, $n$-*step TD*, adjusts the estimation based on $n$ subsequent rewards and a more distant estimation [13]. The TD error of $n$-step TD is calculated as

$$\delta_t = R_t^{(n)} - V(s_t), \tag{5}$$

where $R_t^{(n)}$ is known as the $n$-*step return*, defined by

$$R_t^{(n)} = \sum_{k=0}^{n-1} \gamma^k r_{t+k} + \gamma^n V(s_{t+n}). \tag{6}$$

Furthermore, $n$-step TD can be generalized to *TD($\lambda$)*, which adjusts the current estimation based on all subsequent estimations [13], [22]. Its TD error is then written as

$$\delta_t^\lambda = R_t^\lambda - V(s_t), \tag{7}$$

where $R_t^\lambda$ is known as the $\lambda$-*return*, defined by

$$R_t^\lambda = (1 - \lambda) \sum_{n=1}^{T-t-1} \lambda^{n-1} R_t^{(n)} + \lambda^{T-t-1} R_t, \text{ in which} \tag{8}$$

$$R_t = \sum_{k=0}^{T-t-1} \gamma^k r_{t+k}. \tag{9}$$

In TD($\lambda$), the *trace decay parameter* $\lambda$ controls the adjustments to be propagated from the distant states to the current state, i.e., a higher $\lambda$ increases the proportion of adjustments from more distant states. For 2048, $\lambda$ is usually set to 0.5 when TD($\lambda$) is applied [3], [5]. Note that the common *eligibility traces* [13] implementation for TD($\lambda$) is infeasible for 2048, since the used $n$-tuple networks have millions of features [3], [5]. Instead, the simplified *n-step TD($\lambda$)* is applied in practice, which limits the adjustment from only $n$ subsequent states.

The above learning framework is to evaluate the state values, i.e., $V(s_t)$. However, from the perspective of taking actions, it is more efficient for 2048 to evaluate the afterstate values, i.e., $V(s'_t)$, instead, called the *afterstate learning framework* [2], [5]. With afterstate values, the policy function can be more efficient than that in (2), as described as follows.

$$\pi(s_t) = \underset{a_t}{\mathrm{argmax}}\big(r_t + V(s'_t)\big). \tag{10}$$

Similarly, the TD error is then calculated as

$$\delta_t = r_{t+1} + V(s'_{t+1}) - V(s'_t). \tag{11}$$

Note that the reward $r_t$ corresponding to $s_t$ and $a_t$ is not counted in (11) when calculating the TD target for $V(s'_t)$. That is, $V(s_t) = \mathbb{E}[r_t + V(s'_t)]$ and $V(s'_t) = \mathbb{E}[r_{t+1} + \cdots]$ for each state-afterstate pair.

Depending on the order of adjusting afterstate values within an episode, *forward update* and *backward update* are both common implementations [23], [24]. Forward update here refers to the scheme where all states are updated in order of an episode from the initial state to the terminal state; backward update simply reverses this order. Note that adjusting with other strategies is possible, e.g., with shuffled order, but it is not required for 2048.

In addition, *Q-learning* [13], [25] can also be employed to 2048, in which the *Q-value* of each action, $Q(s_t, a_t) = r_t + V(s'_t)$, are trained for each state. Thus, the policy is derived as

$$\pi(s_t) = \underset{a_t}{\mathrm{argmax}}\, Q(s_t, a_t). \tag{12}$$

And the Q-learning network is trained similarly to the TD learning as follows.

$$Q(s_t, a_t) \leftarrow Q(s_t, a_t) + \alpha\left(r_t + \underset{a_{t+1}}{\max}\, Q(s_{t+1}, a_{t+1}) - Q(s_t, a_t)\right), \tag{13}$$

where $a_{t+1} \in \mathcal{A}(s_{t+1})$. Note that the discount factor $\gamma = 1$ applied to $\max Q(s_{t+1}, a_{t+1})$ is also disregarded for simplicity. However, for 2048 programs with $n$-tuple networks, learning with Q-learning requires four times as much memory to store the feature weights for each state-action pair. Even more, the performance of Q-learning is significantly worse than TD learning [2], so it has not been widely used in the research of 2048.

## 2.3 Advanced TD Methods

*Multistage temporal difference (MS-TD) learning* proposed by Yeh *et al.* [3] is a kind of hierarchical TD learning [26] that divides the entire episode into multiple *stages*, in which each stage has an independent value function, used in [3], [5], [27], [28]. MS-TD learning improves the performance at the cost of additional storage for stages. The work [3] applied this method with 5-stage to 2048 and obtained the first-ever 65536-tile.

*Temporal coherence (TC) learning* proposed for 2048 by Jaśkowski [5] is a TD variant with adaptive learning rates [29]. Instead of adjusting the learning rate $\alpha$ directly, this method introduces the *coherence* parameters $\beta_i$ for the $i$th feature weight, denoted by $\theta_i$, to modulate the adjustments as

$$\theta_i \leftarrow \theta_i + \alpha \beta_i \delta_t, \tag{14}$$

where $\beta_i$ is calculated from two parameters $E_i$ and $A_i$ for each weight, as

$$\beta_i = \begin{cases} \dfrac{|E_i|}{A_i}, & \text{if } A_i \neq 0 \\ 1, & \text{otherwise.} \end{cases} \tag{15}$$

Both $E_i$ and $A_i$ are initialized with 0 and adjusted by the TD error $\delta_t$

$$E_i \leftarrow E_i + \delta_t \text{ and } A_i \leftarrow A_i + |\delta_t|. \tag{16}$$

Therefore, the amount of adjustment is automatically reduced by coherence $\beta_i$ once the TD error $\delta$ starts to oscillate between positive and negative values. TC learning is an effective learning rate decay method with an overhead of more required memory for storing $E_i$ and $A_i$. This technique can be integrated with MS-TD, e.g., a 16-stage TC method achieved 609104 points and a 70% chance of reaching 32768-tiles on average [5].

## 2.4 N-Tuple Networks

A straightforward method to estimate state values $V(s)$ is to use a tabular implementation for the whole state space. However, the state space requires $18^{4\times4} \approx 1.2 \times 10^{20}$ for 2048, too large to be implemented. Therefore, a *function approximator* is applied in practice. *N-tuple network* is a function approximator that has been successfully applied to applications such as Connect4 [20], Othello [21], pattern recognition [30], as well as 2048 [2], [3], [5], [8].

For $n$-tuple networks for 2048, an $n$-*tuple* $\phi$ is a sequence of distinct *features*, each representing a tuple of $n$ designated cells on the puzzle. To be more specific, the $i$th $n$-tuple $\phi_i$ consists of $n$ predefined locations $\varphi_{i,1}, \varphi_{i,2}, \ldots, \varphi_{i,n}$; and $s[\varphi_{i,j}]$ refers to the cell value at the location $\varphi_{i,j}$ of state $s$. The extracted features $\phi_i(s)$ is defined as follows.

$$\phi_i(s) = (s[\varphi_{i,1}], s[\varphi_{i,2}], \ldots, s[\varphi_{i,n}]). \tag{17}$$

For example, let $\phi_{R1}$ be a 4-tuple that denotes features from the first row, i.e., the associated locations $\varphi_{R1,1}, \varphi_{R1,2}, \varphi_{R1,3}, \varphi_{R1,4}$ are predefined as index numbers 0, 1, 2, 3, respectively. For a puzzle $s$, say the one in Figure 3 (a), a feature $\phi_{R1}(s)$ refers to (32, 2, 0, 0). Similarly, for $s'$, say the one in Figure 3 (b), a feature $\phi_{R1}(s')$ refers to (64, 4, 16, 0).

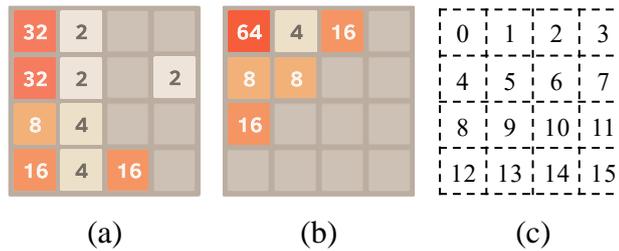

(a)  (b)  (c)

Figure 3. Examples for feature extraction with $n$-tuples for 2048.
(a) and (b) are two 2048 puzzles; (c) is the index of each cell location.
For a 4-tuple (0, 1, 2, 3) denoted by the index in (c), the extracted features for (a) and (b)
are (32, 2, 0, 0) and (64, 4, 16, 0), respectively.

An $n$-tuple network is an implementation for a set of weights of $n$-tuple features. In order to access these *feature weights*, let $\phi$ be an $n$-tuple associated with a lookup table LUT in which the feature weight of $\phi(s)$ is stored at a distinct $\text{LUT}[\phi(s)]$. An illustration of the implementation for $\phi_{R1}$ is as follows. The lookup table $\text{LUT}_{R1}$ for $\phi_{R1}$ needs to contain $c^4$ distinct feature weights, since $\phi_{R1}$ consists of 4 cells, each with $c$ distinct cell values.

Intrinsically, $c$ is 18, i.e., from empty cell to 131072-tile, but is usually set to 16 for efficiency, since 65536-tile and 131072-tile are rarely obtained.

In this work, we define an *m×n-tuple network* to be $m$ different $n$-tuples $\phi_1, ..., \phi_m$ with their corresponding lookup tables $\text{LUT}_1, ..., \text{LUT}_m$. Given a state $s$, the state value estimation $V(s)$, is calculated by summing all the $m$ feature weights $\text{LUT}_i[\phi_i(s)]$ of state $s$ as

$$V(s) = \sum_{i=1}^{m} \text{LUT}_i[\phi_i(s)]. \tag{18}$$

When adjusting a state estimation $V(s)$ by a TD error $\delta$, the adjustment is equally distributed to $m$ feature weights of state $s$. Equation (19) shows how $\delta$ is distributed to a feature weight $\text{LUT}_i[\phi_i(s)]$ in terms of TD(0) with learning rate $\alpha$:

$$\text{LUT}_i[\phi_i(s)] \leftarrow \text{LUT}_i[\phi_i(s)] + \frac{\alpha}{m}\delta. \tag{19}$$

Note that only the $m$ feature weights corresponding to state $s$ need to be adjusted, with the same adjustment $(\alpha/m)\delta$.

For 2048, *symmetric sampling* is a widely used technique that shares the feature weights of tuples eight times by rotating and mirroring [2], [3], [5], [8]. A *symmetrically sampled m×n-tuple network* involves $8m$ $n$-tuples actually, which improves the overall performance without additional lookup tables. Continue the above example, let $\phi_{R1'}$ be a 4-tuple produced by rotating $\phi_{R1}$ counterclockwise, i.e., $\varphi_{R1',1}, \varphi_{R1',2}, \varphi_{R1',3}, \varphi_{R1',4}$ becomes 12, 8, 4, 0, respectively. Then, the feature $\phi_{R1'}(s)$ is (16, 8, 32, 32). Both $\phi_{R1}$ and $\phi_{R1'}$ share the same lookup table $\text{LUT}_{R1}$. Note that unless specified otherwise, an *m×n*-tuple network refers to a symmetrically sampled *m×n*-tuple network for simplicity.

Designing an effective *m×n*-tuple network is not trivial, and has been investigated as a research topic for 2048 [2], [3], [8], [31]. Figure 4 (a), (b), (d), and (e) illustrate *the 4×6-tuple network* proposed by Yeh et al. [3], [28]; Figure 4 (a)–(e) illustrate *the 5×6-tuple network* used by Jaśkowski [5]. Figure 5 illustrates a full set of 6-tuples proposed by Matsuzaki [8], where the best $k\times6$-tuple network consists of the first $k$ listed 6-tuples, e.g., *the 8×6-tuple network* contains all from Figure 5 (a) to (h).

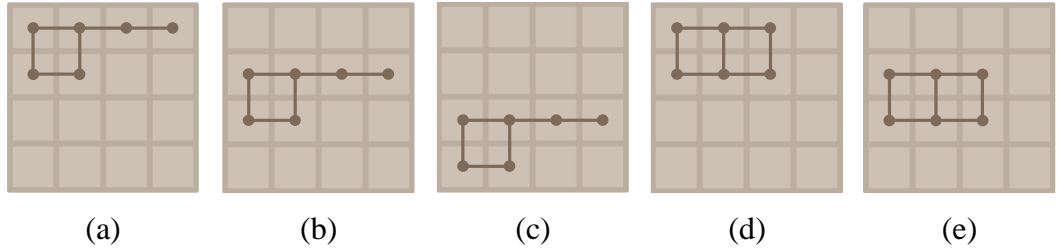

(a) (b) (c) (d) (e)

Figure 4. Yeh's 4×6-tuple network and the 5×6-tuple network.
(a), (b), (d), and (e): the 4×6-tuple network proposed by Yeh *et al.* [3];
(a)–(e): the 5×6-tuple network proposed by Jaśkowski [5].

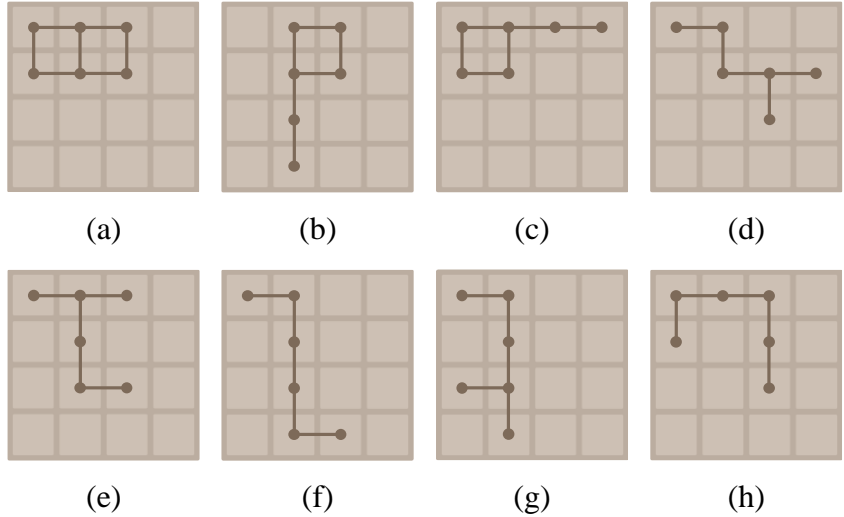

(a) (b) (c) (d)

(e) (f) (g) (h)

Figure 5. Matsuzaki's 8×6-tuple network.
(a)–(h): the 8×6-tuple network proposed by Matsuzaki [8].

$N$-tuple networks are usually implemented using random access memory (RAM); thus, they are also known as *RAM-based neural networks* [32], [33]. Besides $n$-tuple networks, *deep neural networks (DNN)* and *convolutional neural networks (CNN)* may also be applied to 2048 [9], [11], [34]–[38]. DNNs consist of fewer weights overall, but the feature weights of the whole network are involved for each output. As a result, DNNs achieve comparable or even better performance to $n$-tuple networks for 2048 when the number of weights is limited [9], [36], [37]. However, the biggest issue of using DNNs is the high computational overhead for training, which makes it difficult to increase the network size for performance.

## 2.5 Expectimax Search

*Expectimax search* is a technique for stochastic games whose game tree is composed of *max nodes* and *chance nodes* [39]–[41], corresponding to states and afterstates respectively.

Namely, a max node of the state $s_t$ searches all its afterstates $s'_t$ to find the best action; and a chance node of afterstate $s'_t$ is evaluated based on either the expected value of next states, or the TD value $V(s'_t)$ when the ply limit is reached. For 2048 with afterstate learning framework, a *p-ply fixed-depth search* tree has at most $p$ layers of chance nodes.

For a given state $s_t$ and a search limit $p$, a max node searches all available afterstates $s'_t$ for the best action that corresponds to the best value $V_{\max}$, as

$$V_{\max}(s_t, p) = \begin{cases} \max_{\forall a_t}\{r_t + V_{\text{chance}}(s'_t, p-1)\}, & \text{if } s_t \text{ is not terminal} \\ 0, & \text{if } s_t \text{ is terminal}, \end{cases} \quad (20)$$

where $a_t \in \mathcal{A}(s_t)$ is the legal action; $r_t = \mathcal{R}(s_t, a_t)$ and $s'_t = \psi(s_t, a_t)$ are the reward and the afterstate corresponding to $s_t$ and $a_t$; and $V_{\text{chance}}$ is the value function for the chance node. For the afterstate $s'_t$, the value of the chance node is calculated by using the expected value $\mathbb{E}[V(s_{t+1})]$ when the depth limit is not reached; otherwise, using the TD value $V(s'_t)$, as

$$V_{\text{chance}}(s'_t, p) = \begin{cases} \mathbb{E}[V_{\max}(s_{t+1}, p)], & \text{if } p > 0 \\ V(s'_t), & \text{if } p = 0, \end{cases} \quad (21)$$

$$\mathbb{E}[V_{\max}(s_{t+1}, p)] = \sum_{s_{t+1}} \mathcal{P}(s'_t, s_{t+1}) V_{\max}(s_{t+1}, p). \quad (22)$$

Note that $\mathbb{E}[V_{\max}(s_{t+1}, p)]$ expands all possible successors of this afterstate to calculate the expected value. Finally, the agent takes action with the highest value $V_{\max}$ at the root node after the search is completed.

Given a fixed time budget, an expectimax search with *iterative deepening* may perform better than the fixed-depth search since it allows the tree to be deeper at the endgame [5]. To avoid redundant searches, a *transposition table* is applied to cache previously seen states and associated values. Transposition tables use the hash value of states to access the corresponding values. For 2048, *Zobrist hashing* [42] and *MurmurHash* [43] are both effective hashing algorithms for implementing transposition tables.

## 2.6 Monte Carlo Tree Search

*Monte Carlo tree search (MCTS)* is a best-first search algorithm that has been successfully applied to many applications [44], [45], including the game of 2048 [46]. MCTS applies the *upper confidence bounds (UCB)* function to select moves to evaluate, which balances exploration and exploitation, leading the search to explore promising moves more frequently and unpromising moves less frequently.

The typical Monte Carlo tree search uses multiple iterations to determine the best action to play, where each iteration consists of four phases: selection, expansion, simulation, and backpropagation [44], [45]. (i) Selection: Traverse the tree from the root node to recursively select child nodes until reaching a leaf node. (ii) Expansion: Generate a new node for the selected leaf node. (iii) Simulation: Play the game state of the generated node until the game ends with an outcome. (iv) Backpropagation: Update the statistics of all traversed nodes with the outcome of the simulated game. After all the iterations, the action with the highest visit count becomes the best action.

For single-player stochastic games such as 2048, max nodes and chance nodes require different selection policies, in which max nodes select the child node using the UCB function, and chance nodes randomly select the child node using the transition probability. The standard UCB function for selecting child nodes is defined as

$$\underset{a}{\mathrm{argmax}}\left\{Q(s,a) + c\sqrt{\frac{\ln N(s)}{N(s,a)}}\right\}, \tag{23}$$

where $Q(s,a)$ and $N(s,a)$ are the win rate and the visit count of action $a$; $N(s)$ is the visit count of state $s$, i.e., the summation of visit counts of all child nodes of state $s$; and $c$ is the exploration constant.

## 2.7 Stochastic MuZero

Recently, Google DeepMind proposed a series of deep reinforcement learning algorithms, including the well-known *AlphaGo* [47], *AlphaGo Zero* [48], *AlphaZero* [49], and *MuZero* [10]. These algorithms collect high-quality training data using Monte Carlo tree search (MCTS) and optimize the planning model using the collected data. The improvement between MuZero and its ancestors is that it learns the dynamics of the environment and plans with the learned model, i.e., unlike its ancestors, MuZero does not require a perfect simulator during the planning. This significant breakthrough allows MuZero to be applied to more environments where working with perfect simulators is too costly, thus making MuZero to be even more promising for complex real-world problems.

For planning, MuZero performs MCTS with a learned model consisting of three functions: the *representation function h*, the *prediction function f*, and the *dynamics function g*.

$$s^0 = h(o_0, \ldots, o_t), \tag{24}$$

$$p^k, v^k = f(s^k), \tag{25}$$

$$s^{k+1}, r^k = g(s^k, a^k). \tag{26}$$

The representation function $h$ maps previous *observations* $o_0, \ldots, o_t$ to a *hidden state* $s^0$. The prediction function $f$ predicts the *policy* $p^k$ and the *value* $v^k$ of the hidden state $s^k$, where the policy $p^k$ represents the probabilities of taking each action; the value $v^k$ represents the win rate of the state $s^k$. The dynamics function $g$ transforms the state $s^k$ and an action $a^k$ sampled from $p^k$ to the next state $s^{k+1}$ and provides the reward $r^k$ corresponding to action $a^k$.

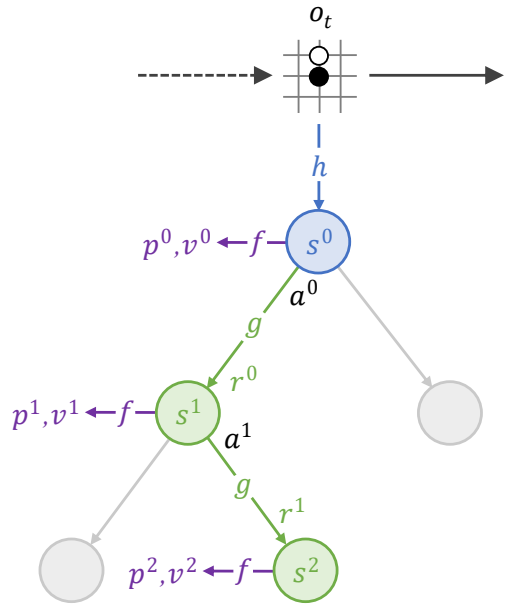

Figure 6. MCTS in MuZero.

With these functions, MuZero performs MCTS at each timestep $t$. Figure 6 illustrates an example of MCTS in MuZero. First, an observation $o_t$ is obtained from the environment. The representation function $h$ maps previous observations $o_0, \ldots, o_t$ to a hidden state $s^0$ to start the MCTS with a number of iterations that consists of the following three procedures: (i) Selection: To select an action $a^k$ to visit, MuZero uses a variant of the PUCT function [50] that takes the policy $p^k$. The selection repeats until a leaf state $s^l$ is reached. (ii) Expansion: A new state $s^{l+1}$ is expanded by using the dynamics function $g$. The policy $p^{l+1}$ and the value $v^{l+1}$ is also predicted by the prediction function $f$. (iii) Backup: The predicted value $v^{l+1}$ of the newly-expanded state is used for backpropagation to update the statistics of the search tree.

At each timestep $t$, MCTS constructs the *search policy* $\pi_t$ by applying softmax to the visit counts of actions at the root state after completing all iterations. Then, an action $a_t$ sampled from the search policy $\pi_t$ is applied to the environment, which provides the immediate reward $u_t$ and the next observation $o_{t+1}$. The planning procedure is illustrated in Figure 7. On the other hand, by using the game-play data collected in the trajectories, the three model functions are jointly trained to predict three quantities: $p^k \approx \pi_{t+k}$, $v^k \approx z_{t+k}$, and $r^k \approx u_{t+k}$, where $z_{t+k}$ is either the outcome of the episode or the $n$-step return.

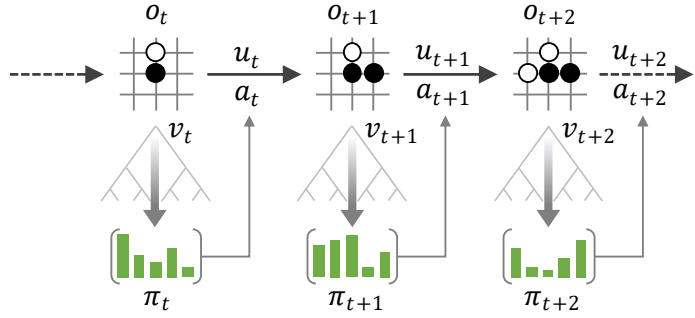

Figure 7. Planning by MCTS in MuZero.

*Stochastic MuZero* [11] extends MuZero to stochastic environments by introducing the concepts of afterstates; it has been successfully applied to 2048 and achieved an average score of approximately 510000 points, comparable to other learning algorithms. To support planning in stochastic environments, the model consists of five functions: the representation function $h$, the prediction function $f$, the *afterstate dynamics function* $\phi$, the *afterstate prediction function* $\psi$, and the dynamics function $g$.

$$s^0 = h(o_0, \dots, o_t), \tag{27}$$
$$p^k, v^k = f(s^k), \tag{28}$$
$$as^k = \phi(s^k, a^k), \tag{29}$$
$$\sigma^k, Q^k = \psi(as^k), \tag{30}$$
$$s^{k+1}, r^k = g(as^k, c^k). \tag{31}$$

The representation function $h$ and the prediction function $f$ are the same as those in MuZero. The afterstate dynamics function $\phi$ transforms the state $s^k$ and an action $a^k$ sampled from $p^k$ to the afterstate $as^k$. The afterstate prediction function $\psi$ predicts the *chance outcome* $\sigma^k$ and the value $Q^k$ of the afterstate $as^k$, where the chance outcome $\sigma^k$ represents the distribution over possible future chance events. The dynamics function $g$ transforms the afterstate $as^k$ and a chance outcome $c^k$ sampled from $\sigma^k$ to the next state $s^{k+1}$ and provides the reward $r^k$ corresponding to action $a^k$ and chance $c^k$. An example of MCTS in Stochastic MuZero is depicted in Figure 8.

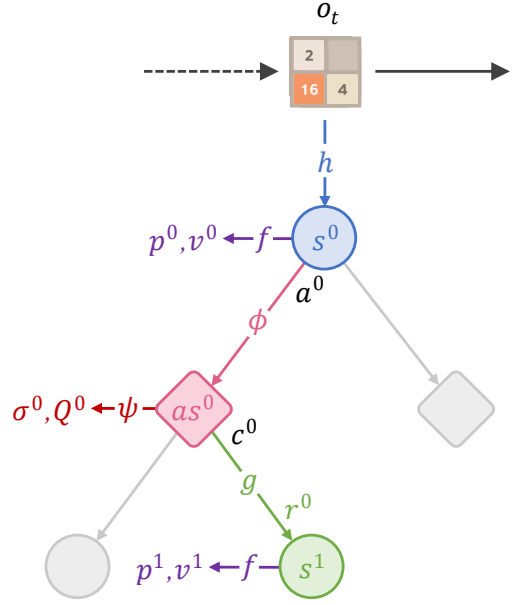

Figure 8. MCTS in Stochastic MuZero.

To model the chance outcomes $\sigma$, Stochastic MuZero proposes to use a novel variant of *Vector Quantized Variational AutoEncoder (VQ-VAE)* [51], which adopts a fixed codebook of one-hot vectors. Thus, the encoder embedding $c$ can be modeled as a categorical variable. Note that a decoder is not required. The chance outcomes are jointly trained with other functions with training targets: $\sigma \approx \text{onehot}(c)$ and $c \approx \text{onehot}(c)$, where $\text{onehot}(c)$ produces the closest one-hot vector of the encoder output.

## 2.8 Bitboard

*Bitboard* is an optional improvement for storing states. For 2048, each tile only needs 4 bits, and the whole puzzle requires 64 bits, which exactly fits into a 64-bit register. Note that 65536-tiles and 131072-tiles are nearly impossible to reach and can be safely ignored in most cases. As the sliding result is deterministic, a lookup table that pre-calculates the sliding results of every 16-bit row can be built when the program first executes. During runtime, the program can simply find the sliding results row by row efficiently. Also, the generation of isomorphic states through reflection or transposition can be implemented with bitwise operations. Since bitboards are efficient in both memory usage and calculation time, most of the efficient 2048 programs, e.g., [52]–[55], are implemented with bitboards.

# Chapter 3  Optimistic Temporal Difference Learning

*Temporal difference (TD) learning* and its variants, such as *multistage TD (MS-TD) learning* and *temporal coherence (TC) learning*, have been successfully applied to 2048. These methods rely on the stochasticity of the environment of 2048 for exploration. In this chapter, we propose to employ *optimistic initialization (OI)* to encourage exploration for 2048, and empirically show that the learning quality is significantly improved. This method has been published in IEEE Transaction on Games [56]. This approach optimistically initializes the feature weights to very large values. Since weights tend to be reduced once the states are visited, agents tend to explore those states which are unvisited or visited a few times. Our experiments show that both TD and TC learning with OI significantly improve performance. As a result, the network size required to achieve the same performance is significantly reduced. With additional tunings such as expectimax search, multistage learning, and *tile-downgrading technique*, our design achieves state-of-the-art (SOTA) performance, namely an average score of 625377 and a rate of 72% to reach 32768-tiles. In addition, for sufficiently large tests, 65536-tiles are reached at a rate of 0.02%.

## 3.1  Motivation

Although various methods have been continually proposed and improved, these TD methods for 2048, however, were still based on the greedy policy [13] to deterministically choose actions with maximum estimations, and thus relied on the stochastic environment to provide enough randomness for exploration. In addition, it is observed that TD learning tends not to reach large tiles when training is saturated in terms of the average score, namely when the average score increases to nearly the highest [3], [28]. This reflects a potential problem of exploration deficiency. On the other hand, recent works [3], [8], [31] tended to employ larger $n$-tuple networks for higher performance. Hence, as the networks become larger, the issue of exploration deficiency becomes non-negligible. Moreover, in addition to the size of the network, the learning rate is another factor related to exploration deficiency. TC learning effectively improved the performance but converged rather fast [5], [29], possibly resulting in less exploration. Researchers [2], [5] noticed the exploration issue and tried some exploration mechanisms, such as $\epsilon$-greedy and softmax, but neither worked for 2048. Thus, it was simply assumed that the stochastic environment provided enough randomness, and left the efficient exploration for 2048 as an open question.

In this work, we propose to use optimistic initialization (OI) to improve the TD methods for 2048. The approach is to optimistically initialize feature weights to large values in order to encourage exploration [13], [57]. Namely, those feature weights rarely adjusted or visited tend to be high, and therefore the value adjustments are often negative, i.e., these weights tend to be reduced. Thus, agents tend to explore the less visited states next time. All feature weights eventually converge after sufficient visits.

Our experiments show that OI significantly improves the performance of TD methods. With additional tunings such as expectimax search, multistage learning, and tile-downgrading technique, our design outperforms the previous SOTA result [5] and achieves new SOTA performance, namely an average score of 625377 and a rate of 72% to reach 32768-tiles. Even more, our method requires only 20% of network weights compared with the previous SOTA method. In addition, for sufficiently large tests, 65536-tiles are reached at a rate of 0.02%.

The chapter is organized as follows. Section 3.2 introduces the optimistic initialization and the proposed optimistic methods for 2048. Section 3.3 and 3.4 conduct experiments and analyses of the optimistic methods. Section 3.5 summarizes the results with concluding remarks, and discusses potentially related techniques with directions for possible future research.

## 3.2 Optimistic Methods

In this section, optimistic methods for 2048 are introduced. Section 3.2.1 reviews the potential problems of insufficient exploration in previous works. Sections 3.2.2 and 3.2.3 present TD learning with optimistic initialization for 2048. Finally, Section 3.2.4 describes how to determine initial values and use them in $n$-tuple networks.

### 3.2.1 Insufficient Exploration

In reinforcement learning, the *exploration-exploitation dilemma* has been intensively studied by researchers for decades [13], [57], [58]. Many methods, including UCB, $\epsilon$-greedy, softmax, as well as optimistic initialization, are proposed to balance between exploration and exploitation [59] for better learning performance.

Most previous RL-related works for 2048 [2], [3], [5], [6], [8], [23], [27], [28], [31] are based on simple TD methods or some of its variants. Among these works, the agents simply follow the greedy policy with respect to the estimations, i.e., always select an action with maximum value during training. Past works [2], [5] have already noticed this issue, and have tried some explicit exploration techniques such as $\epsilon$-greedy and softmax. However, they failed to have these techniques work for 2048. As a result, these TD methods involve no explicit exploration to choose actions and entirely rely on the stochasticity of the environment for exploration.

For example, MS-TD learning is proposed to cope with the issue that TD learning tends not to reach large tiles [3], [28]. It improves performance by employing additional networks, while the issue of potential exploration deficiency of each stage remains not addressed. Another effective technique is TC learning, which reduces weight adjustments to accelerate network convergence [5]. This approach is not designed to improve exploration but to improve exploitation with fast convergence. However, fast convergence exacerbates the exploration issue, especially when applying a larger $n$-tuple network. An example that TC converges with insufficient exploration is provided in Appendix A.

### 3.2.2 Optimistic initialization

Optimistic initialization (OI) is an approach that employs optimistic initial estimations to encourage exploration [13], [57]. This approach has been widely applied to RL applications and is considered to have good convergence in practice [13]. Instead of setting the estimations to zero or random, the technique optimistically initializes them to a large value to encourage the agent to explore. Due to the large value, the estimations tend to be reduced once the corresponding states are visited, leading an agent to select unexplored or rarely explored actions the next time it revisits the same state. This process repeats until all the actions are sufficiently explored, even if the greedy policy is always applied. As the actions are progressively explored, the learning algorithm can seamlessly switch from exploration to exploration.

The estimations will eventually converge and may even converge to a near-optimal policy when the initial value is set sufficiently large [58], [60], namely, the upper bound of the value function. Thus, the learning algorithm must explore unvisited or rarely visited states and reduce the corresponding estimations before exploiting the best one [61]. Consequently, OI may significantly increase the training time in exchange for encouraging exploration. However, the advantage of OI is that it relies only on a properly set initial value, i.e., it does not require a scheduler to reduce the level of exploration. Compared with OI, other standard exploration techniques need not only an initial hyperparameter for exploration (e.g., $\epsilon$ in $\epsilon$-greedy; $T$ in softmax) but also a schedular to reduce the exploration.

### 3.2.3 Optimistic TD Methods

Past works summarized that non-greedy behaviors, e.g., $\epsilon$-greedy or softmax, significantly inhibit learning performance [2], [5]. Therefore, we propose to use OI as an exploration mechanism for 2048. The proposed OI methods perform exploration while conserving the greedy behavior, therefore mitigating the previous inhibition phenomenon. In addition, since the explicit exploration technique is independent of some existing learning methods, such as multistage and TC learning, OI can be easily applied together with these methods.

The first objective is to improve the existing TD and TC methods for 2048 by using OI, which forms the *optimistic TD (OTD)* and the *optimistic TC (OTC) learning*, respectively. Second, we propose a two-phase method using OI, called *OTD+TC learning*, a hybrid learning paradigm that combines the advantages of both OTD and OTC learning. OTD+TC learning first performs OTD learning with a fixed learning rate to further encourage exploration for a while, and then, in the second phase, continues with TC fine-tuning for exploitation.

When compared to TD or TC, OTD+TC learning includes two new hyperparameters, $V_{init}$ and $P_{TC}$. $V_{init}$ is the initial value of the function approximator; and $P_{TC}$ is the proportion of the fine-tuning phase to the total phases. If $V_{init}$ is set to 0, OTD+TC becomes non-optimistic. If $P_{TC}$ is set to 0% (or 100%), OTD+TC becomes pure OTD (or OTC).

### 3.2.4 Initial Values for OI

Based on the proof of the *optimistic Q-learning* in [60], the *initial value* $V_{init}$ should be set to the theoretical maximum to ensure that the network converges to a near-optimal policy. However, the theoretical maximum for 2048 is an extremely large number that is nearly impossible to obtain. Using such a large $V_{init}$ wastes too much time, it is non-trivial to choose $V_{init}$ such that training result and training time are balanced.

The initial value $V_{init}$ is chosen as illustrated as follows. Consider using an $m \times n$-tuple network for training with the non-optimistic method as a dry run, and then estimate $V_{init}$ based on the result of the dry run. To initialize the network feature weights, we evenly distribute the value $V_{init}$ over these feature weights as

$$\text{LUT}[i] \leftarrow \frac{V_{init}}{m} \text{ for all } i. \tag{32}$$

## 3.3 Experiments and Analysis

In this section, experiments are presented to analyze the effectiveness of optimistic TD learning for 2048. Common training settings are described as follows. For convergence, each $n$-tuple network was trained with 100M episodes by using the afterstate learning framework. The learning performance was evaluated every 1M training episodes, and each performance evaluation consists of 100k testing episodes. For statistics, each method was trained with five runs, each with one individual trained network with a different initial random seed. The experiments were performed on workstations with Intel Xeon E5 processors. To speed up the training process, *lock-free optimistic parallelism* with 20 threads was applied as in [5]. Specific settings will be described in each experiment below.

This chapter presents the results in tables and figures, in which each value represents the average of five trained networks. Tables show the final results after training, while figures show the training progresses. For tables (e.g., Table 1), a value indicates the average result after 100M training episodes. For figures (e.g., Figure 9), a point at time $t$ indicates the average result after $t$M training episodes. Both tables and figures are accomplished with 95% confidence intervals.

The experiments are organized as follows. Section 3.3.1 analyzes the effectiveness of initial values on OTD and OTC learning. Section 3.3.2 analyzes OTD+TC learning with different fine-tuning proportions on networks of different sizes.

### 3.3.1 OTD and OTC Learning

In order to demonstrate the effectiveness of OI and obtain an appropriate initial value, we analyze how OTD learning and OTC learning perform with various initial values. Two different $n$-tuple networks, Yeh's 4×6-tuple (Figure 4) and Matsuzaki's 8×6-tuple (Figure 5), were used. Based on their average scores with non-optimistic TD learning, we test the initial value $V_{init}$ to 0, 40k, 80k, 160k, 320k, 640k, respectively. For OTD learning, the learning rate $\alpha$ was initially set to 0.1 and was reduced to 0.01 and 0.001 after completing 50% and 75% of training; for OTC learning, a larger initial learning rate is required, so $\alpha$ was set to 1.0 without manual decay as in [5]. The results are organized as follows. OTD and OTC learning for the 4×6-tuple will be discussed first, and then followed by OTD and OTC learning for the 8×6-tuple.

For the 4×6-tuple network, OTD learning did not obviously improve the learning quality, as shown in Table 1. On the other hand, OTC learning significantly improved the performance, especially for the reaching rate of 32768-tiles. As the results summarized in Table 2 and Figure 9, OTC with 320k outperformed non-optimistic TC ($V_{init} = 0$), and even achieved a rate of 5.30% reaching 32768-tiles, which is the first-ever reaching rate with such a 1-stage network. As shown in Figure 10, many methods reached maximum scores of around 385000. This implicitly indicates to achieve states with 16384-tile, 8192-tile, 4096-tile, and 2048-tile. Note that a method can continue to obtain much more scores if a 32768-tile is reached. Only OTC with 320k and 640k provided enough exploration to achieve 32768-tiles. Interestingly, as shown in Figure 10, OTC with 640k reached 32768-tiles earlier than OTC with 320k, but eventually ended at a worse average score. In general, a large $V_{init}$ needs more time to converge. Our conjecture is that 640k is too large to converge. Thus, we prefer 320k for obtaining higher average scores.

Table 1. Performance of OTD learning in the 4×6-tuple network.

| $V_{init}$ | Average Score | 8192 [%] | 16384 [%] | 32768 [%] |
|---|---|---|---|---|
| 0 | 251920 ± 11217 | 92.87 ± 0.90% | 65.23 ± 2.86% | 0.00 ± 0.00% |
| 40k | **254784 ± 13034** | **93.20 ± 1.08%** | **65.89 ± 3.60%** | 0.00 ± 0.00% |
| 80k | 253393 ± 11732 | 92.52 ± 1.13% | 65.25 ± 3.35% | 0.00 ± 0.00% |
| 160k | 244875 ± 10590 | 91.40 ± 1.41% | 62.29 ± 3.27% | 0.00 ± 0.00% |
| 320k | 239075 ± 14781 | 89.68 ± 2.11% | 59.73 ± 5.00% | 0.00 ± 0.00% |
| 640k | 228032 ± 1056 | 87.64 ± 0.44% | 55.63 ± 0.48% | 0.00 ± 0.00% |

Table 2. Performance of OTC learning in the 4×6-tuple network.

| $V_{init}$ | Average Score | 8192 [%] | 16384 [%] | 32768 [%] |
|---|---|---|---|---|
| 0 | 265481 ± 3318 | 93.36 ± 1.07% | **68.33 ± 0.52%** | 0.00 ± 0.00% |
| 40k | 263284 ± 2445 | **93.68 ± 0.32%** | 66.72 ± 1.79% | 0.00 ± 0.00% |
| 80k | 265703 ± 2471 | 93.60 ± 0.26% | 67.93 ± 1.49% | 0.00 ± 0.00% |
| 160k | 264058 ± 1412 | 93.24 ± 0.28% | 67.47 ± 0.35% | 0.00 ± 0.00% |
| 320k | **280281 ± 5267** | 93.11 ± 0.53% | 67.83 ± 1.50% | **5.30 ± 0.58%** |
| 640k | 252254 ± 6836 | 89.18 ± 1.48% | 60.22 ± 1.98% | 3.25 ± 0.38% |

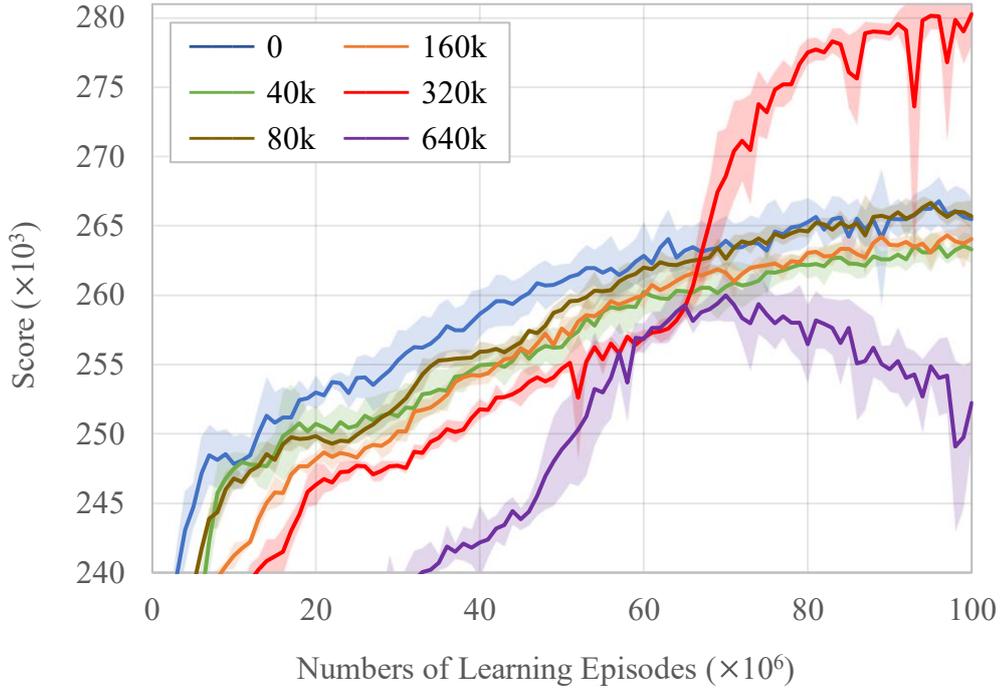

Figure 9. Average scores of OTC learning in the 4×6-tuple network.

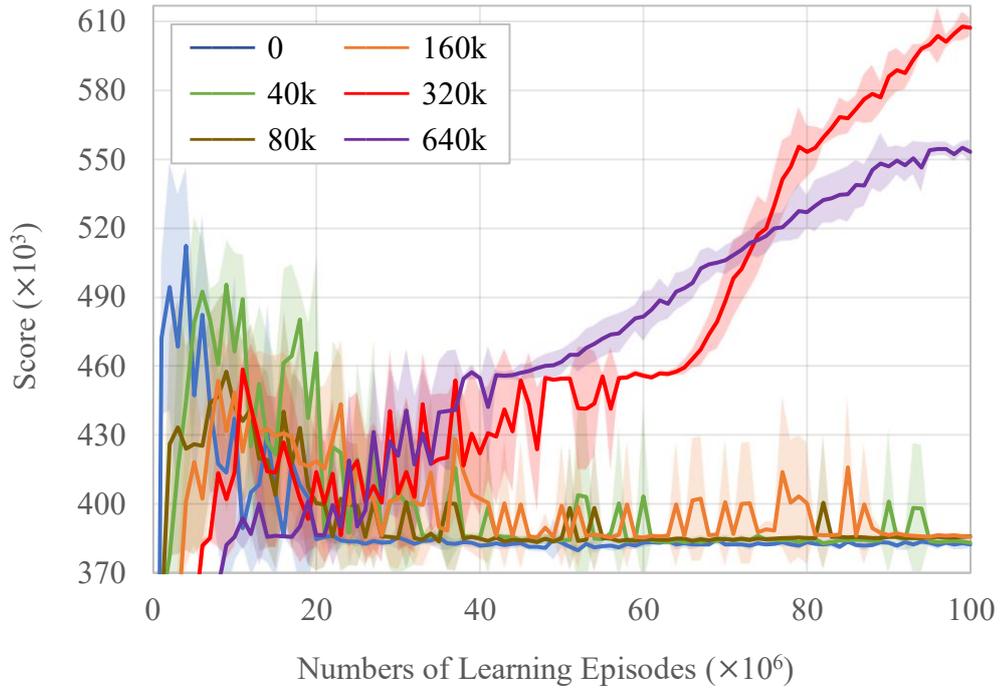

Figure 10. Maximum scores of OTC learning in the 4×6-tuple network.

For the 8×6-tuple network, in contrast to the 4×6-tuple, OTD learning significantly improves the performance. As shown in Table 3 and Figure 11, networks initialized with appropriate values (80k ≤ $V_{init}$ < 640k) achieved higher scores, and even reached 32768-tiles well, especially for $V_{init}$ = 160k and 320k. As shown in Figure 12, networks trained with low exploration ($V_{init}$ < 80k) still stuck at the barrier of 32768-tiles; and networks trained with 640k still did not converge. On the other hand, the results of OTC learning are summarized in Table 4. OTC learning also performed well, but it was slightly worse than OTD for the 8×6-tuple network. Interestingly, for $V_{init}$ = 80k on the 8×6-tuple network, OTD achieved 32768-tiles, while OTC did not. Since such a phenomenon did not occur on the 4×6-tuple, it can be derived from this observation that exploration may be restricted by TC learning since the weight adjustments drop too fast; therefore, a pure TD method plays an important role for larger networks.

Table 3. Performance of OTD learning in the 8×6-tuple network.

| $V_{init}$ | Average Score | 8192 [%] | 16384 [%] | 32768 [%] |
| --- | --- | --- | --- | --- |
| 0 | 309208 ± 5788 | **97.24 ± 0.58%** | 85.13 ± 0.78% | 0.00 ± 0.00% |
| 40k | 306438 ± 2400 | 97.27 ± 0.11% | 84.11 ± 0.44% | 0.00 ± 0.00% |
| 80k | 365364 ± 4856 | 97.16 ± 0.17% | 85.59 ± 0.67% | 21.23 ± 1.04% |
| 160k | **369172 ± 3341** | 97.18 ± 0.30% | **85.70 ± 0.63%** | **22.47 ± 1.31%** |
| 320k | 361471 ± 5473 | 96.77 ± 0.28% | 84.41 ± 0.96% | 21.75 ± 0.74% |
| 640k | 339492 ± 6747 | 95.79 ± 0.28% | 81.80 ± 1.07% | 18.11 ± 2.12% |

Table 4. Performance of OTC learning in the 8×6-tuple network.

| $V_{init}$ | Average Score | 8192 [%] | 16384 [%] | 32768 [%] |
| --- | --- | --- | --- | --- |
| 0 | 310259 ± 4056 | 96.54 ± 0.13% | **84.81 ± 0.43%** | 0.00 ± 0.00% |
| 40k | 310347 ± 6201 | 96.72 ± 0.71% | 83.59 ± 1.38% | 0.00 ± 0.00% |
| 80k | 311214 ± 2205 | **97.07 ± 0.22%** | 83.45 ± 0.23% | 0.00 ± 0.00% |
| 160k | **361298 ± 4833** | 96.34 ± 0.20% | 84.12 ± 0.52% | 22.26 ± 2.12% |
| 320k | 360228 ± 18829 | 95.73 ± 0.39% | 82.28 ± 2.25% | **22.74 ± 2.38%** |
| 640k | 325357 ± 33416 | 92.76 ± 2.55% | 75.88 ± 6.10% | 17.71 ± 4.89% |

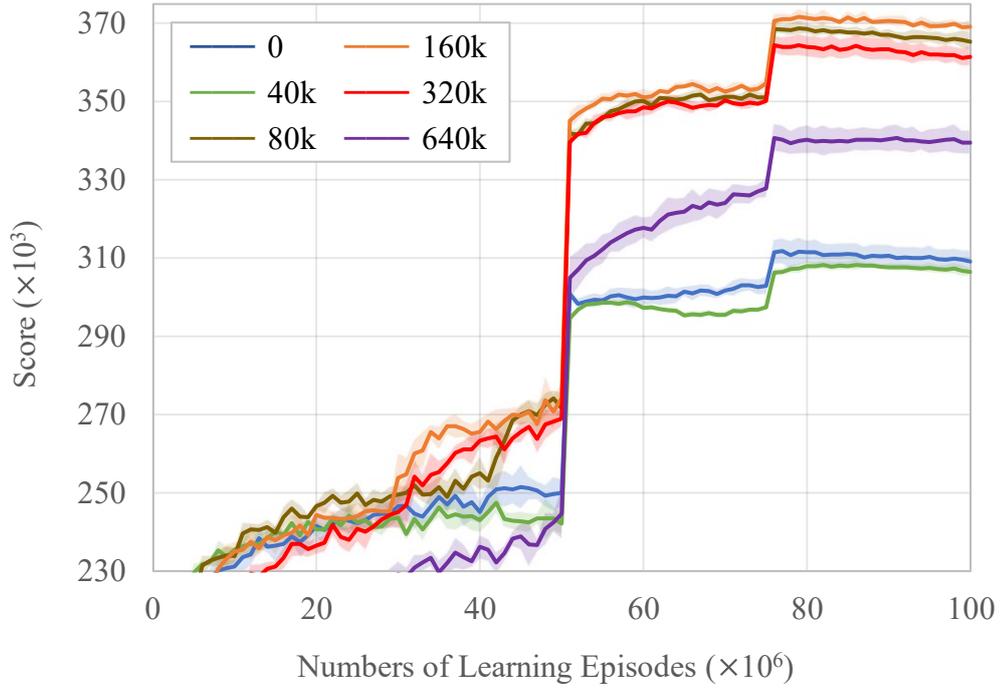

Figure 11. Average scores of OTD learning in the 8×6-tuple network.

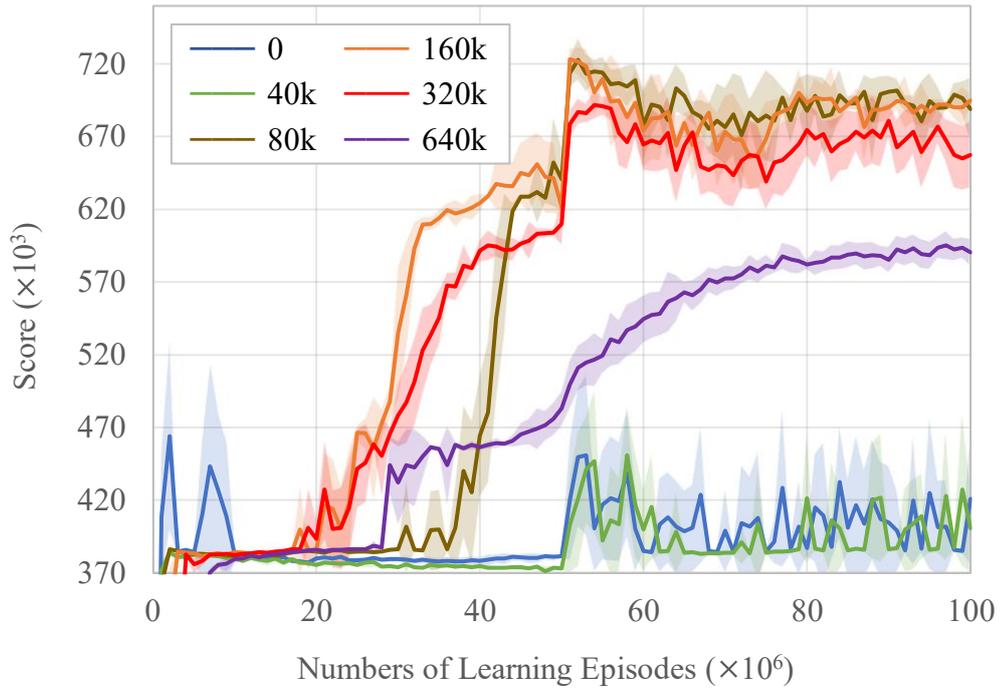

Figure 12. Maximum scores of OTD learning in the 8×6-tuple network.

### 3.3.2 OTD+TC Learning

This experiment analyzes OTD+TC learning. In order to analyze the correlation between OI and network size in detail, five experimental network architectures were chosen and divided into two classes: the smaller networks, including Yeh's 4×6-tuple and 5×6-tuple (Figure 4); and the larger networks, including Matsuzaki's 6×6-tuple, 7×6-tuple, and 8×6-tuple (Figure 5). Since $V_{init}$ = 320k performed well in most cases as described above, $V_{init}$ is set to 320k in the rest of the experiments.

First, we would like to analyze the effectiveness of OTD+TC learning. The average scores and the 32768-tile reaching rates of OTD, OTC, and OTD+TC with $P_{TC}$ = 10% in the five networks with different sizes are summarized in Table 5. OTD+TC outperforms OTD in all cases, and slightly outperforms OTC in most cases of using larger networks. In the cases of using smaller networks, OTC outperforms OTD+TC, while OTD+TC hardly obtained 32768-tiles. In brief, OTC learning performed well regardless of the network size, and OTD+TC learning outperforms OTC learning in most larger networks.

Table 5. Comparison of OTD, OTC, and OTD+TC learning in $n$-tuple networks with different architectures.

| Network | OTD | OTD+TC | OTC |
| --- | --- | --- | --- |
| 4×6-tuple | 239075 ± 14781 (0.00 ± 0.00%) | 261433 ± 3143 (0.02 ± 0.02%) | **280281 ± 5267 (5.30 ± 0.58%)** |
| 5×6-tuple | 272904 ± 7788 (0.01 ± 0.01%) | 279653 ± 8707 (0.47 ± 1.08%) | **313948 ± 2433 (11.48 ± 0.97%)** |
| 6×6-tuple | 314324 ± 8883 (8.33 ± 5.58%) | **337951 ± 4655 (14.39 ± 0.96%)** | 324714 ± 17688 (12.88 ± 5.88%) |
| 7×6-tuple | 352986 ± 3553 (18.43 ± 0.69%) | **360286 ± 2332 (19.18 ± 0.70%)** | 335042 ± 23023 (16.56 ± 6.61%) |
| 8×6-tuple | 361471 ± 5473 (21.75 ± 0.74%) | **370907 ± 1630** (22.18 ± 0.53%) | 360228 ± 18829 **(22.74 ± 2.38%)** |

The values outside and inside parentheses represent average scores and 32768-tile reaching rate, respectively. Networks initialized with $V_{init}$ = 320k; OTD+TC adopted $P_{TC}$ = 10%.

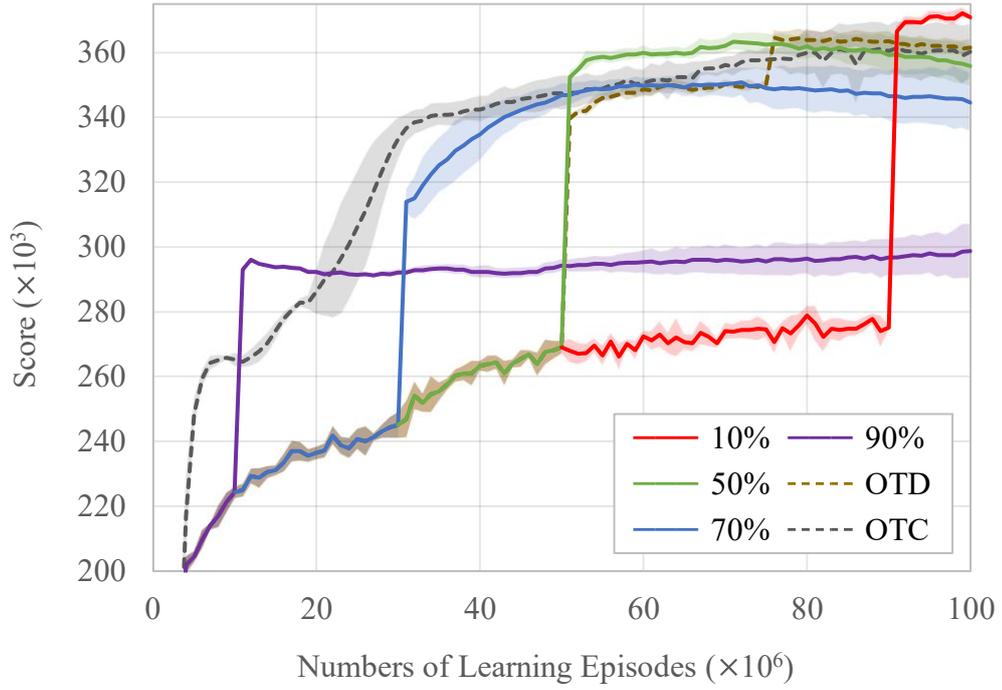

Figure 13. Average scores of OTD+TC learning in the 8×6-tuple network. Networks initialized with $V_{init}$ = 320k.

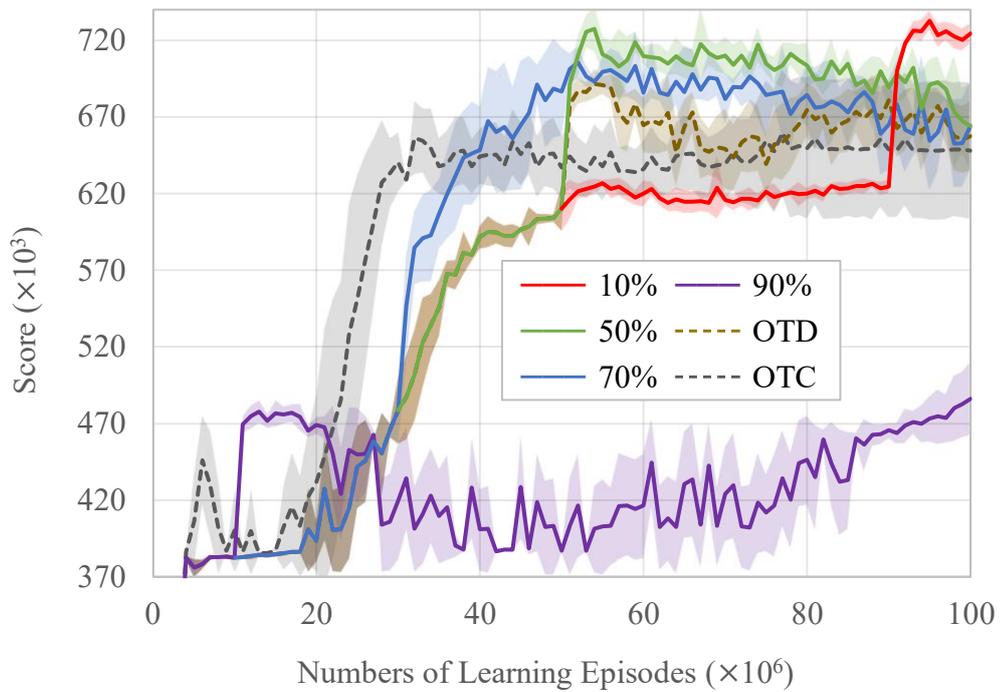

Figure 14. Maximum scores of OTD+TC learning in the 8×6-tuple network. Networks initialized with $V_{init}$ = 320k.

Second, it is interesting to investigate how the fine-tuning proportions $P_{TC}$ affects the performance. Table 6 summarizes the results of more $P_{TC}$ of 10%, 20%, 30%, 50%, 70%, and 90% for the 8×6-tuple network. Figure 13 and Figure 14 illustrate the learning curves of 10%, 50%, 70%, 90%, and OTD and OTC as baselines. Based on the results, it is observed that high proportions (e.g., 70% and 90%) limit the ability to explore, and lead to worse results. Therefore, OTD+TC learning should be performed with low fine-tuning proportions, such as 10% or 20%, to provide sufficient exploration and prevent potential overfitting issues in practice.

Table 6. Performance of OTD+TC learning in the 8×6-tuple network.

| $P_{TC}$ | Average Score | 8192 [%] | 16384 [%] | 32768 [%] |
|---|---|---|---|---|
| 10% | 370907 ± 1630 | **97.26 ± 0.16%** | 85.43 ± 0.34% | 22.18 ± 0.53% |
| 20% | **371198 ± 8301** | 97.19 ± 0.53% | **85.55 ± 1.21%** | **22.78 ± 1.39%** |
| 30% | 366868 ± 7767 | 97.07 ± 0.45% | 84.82 ± 1.28% | 22.28 ± 1.38% |
| 50% | 355806 ± 13994 | 96.32 ± 0.73% | 83.64 ± 1.97% | 20.61 ± 2.24% |
| 70% | 344570 ± 20316 | 95.99 ± 1.06% | 82.04 ± 3.58% | 17.78 ± 2.83% |
| 90% | 298673 ± 18995 | 95.25 ± 1.64% | 80.05 ± 4.60% | 0.86 ± 2.00% |

Networks initialized with $V_{init} = 320$k.

## 3.4 Further Fine-Tuning

To further improve the performance, we demonstrate how to apply optimistic methods together with other established techniques. The fine-tuning experiments are organized as follows. Section 3.4.1 improves the performance by adding the expectimax search. Section 3.4.2 extends the optimistic methods with multistage learning. Section 3.4.3 proposes to use tile-downgrading to improve the performance further and analyze its effect.

### 3.4.1 Expectimax Search

As also described in [2], [5], the expectimax search can be used to improve the performance further. We applied the expectimax search to two cases, the 4×6-tuple network with OTC and the 8×6-tuple with OTD+TC, since both generally performed well, as shown above. The performance of three different depths, 1-ply, 3-ply, and 5-ply, were evaluated for the tests of 1M, 10k, and 100 episodes, respectively. Each evaluation is also repeated five times by using the five trained networks. Transposition tables were used to speed up the search.

Table 7 lists the performance of the 4×6-tuple and the 8×6-tuple network with 1-ply, 3-ply, and 5-ply modified expectimax search. Benefiting from OI, both average scores and 32768-tile reaching rates achieved SOTA for the same 1-stage networks and search depths, to the best of our knowledge. However, the performance was saturated around 5-ply. In addition, iterative deepening, although considered a better strategy for 2048, did not improve the optimistic methods in our experiments.

Table 7. Performance of optimistic learning with expectimax search in the 4×6-tuple and the 8×6-tuple networks.

| Network | Search | Average Score | 8192 [%] | 16384 [%] | 32768 [%] |
|---|---|---|---|---|---|
| 4×6-tuple | 1-ply | 280123 ± 4920 | 93.00 ± 0.61% | 67.79 ± 1.39% | 5.32 ± 0.43% |
| | 3-ply | 417712 ± 2988 | 99.66 ± 0.10% | 94.95 ± 0.60% | 37.18 ± 1.62% |
| | 5-ply | **445085 ± 8843** | **99.80 ± 0.89%** | **96.60 ± 5.02%** | **51.20 ± 3.58%** |
| 8×6-tuple | 1-ply | 370194 ± 4366 | 97.23 ± 0.18% | 85.30 ± 0.46% | 22.18 ± 0.92% |
| | 3-ply | 475126 ± 6407 | 99.73 ± 0.08% | 96.88 ± 0.78% | 50.73 ± 1.56% |
| | 5-ply | **500098 ± 8590** | **100.00 ± 0.00%** | **98.20 ± 3.85%** | **57.80 ± 2.97%** |

The 4×6-tuple and the 8×6-tuple networks were trained with OTC and OTD+TC, respectively. Networks initialized with $V_{init}$ = 320k; OTD+TC learned with $P_{TC}$ = 10%.

In addition, we observed that afterstate values $V(s'_t)$ may become negative after long training. Negative values may seriously influence the correctness of the tree search. Therefore, a workaround that rectified negative values was applied to the search to mitigate the issue. More descriptions of the negative values and the workaround are in Appendix B. To summarize in short, we consider the negative values are caused by the feature-sharing mechanism of $n$-tuple network when excessive long training is applied. Since long training is necessary for OI to converge, we leave the issue of negative values as an open question.

### 3.4.2 Multistage Learning

We further evaluate optimistic TD learning together with the multistage paradigm [3], named *multistage OTD+TC (MS-OTD+TC) learning*. As in Section 3.3, we also use the two cases, the 4×6-tuple network with OTC and the 8×6-tuple network with OTD+TC. For both, the first stages started with initial states, and the second stages started with states with 16384-tile. Note that the first stages are directly copied from trained networks, and the second stages also received 100M training episodes. All stages were trained using their appropriate optimistic methods and hyperparameters, $V_{init}$ and $P_{TC}$, as mentioned in previous experiment.

Table 8 presents the performance of the 2-stage 4×6-tuple and 8×6-tuple network with 1-ply, 3-ply, and 5-ply expectimax search. For comparison, the non-optimistic 2-stage networks are also provided as baselines. In the second stage, average scores were improved by around 10 ± 5%. For further improvement, we attempted to add the third stage to the 2-stage 8×6-tuple network by starting with 16384-tile + 8192-tile. However, such a 3-stage 8×6-tuple network only improved the average scores by less than 2%, as presented in Table 8. We conjecture that the hyperparameters $V_{init}$ and $P_{TC}$ for the 8×6-tuple, tuned for the first stage, did not fit the third stage. As the performance of the 2-stage network was sufficient, we did not tune the 3-stage network using different hyperparameters.

Table 8. Performance of multistage optimistic learning with expectimax search in the 4×6-tuple and the 8×6-tuple networks.

| Network | Search | Average Score | 8192 [%] | 16384 [%] | 32768 [%] |
|---|---|---|---|---|---|
| 2-stage | 1-ply | 291428 ± 5229 | 93.30 ± 1.21% | 68.58 ± 1.05% | 7.72 ± 1.80% |
| 4×6-tuple | 3-ply | 463784 ± 12569 | 99.72 ± 0.15% | 96.16 ± 0.36% | 46.97 ± 1.28% |
| ($V_{init}$=0) | 5-ply | 478831 ± 13174 | **100.00 ± 0.00%** | 98.40 ± 1.10% | 53.80 ± 3.29% |
| 2-stage | 1-ply | 290549 ± 4105 | 93.03 ± 0.62% | 67.88 ± 1.36% | 8.42 ± 0.44% |
| 4×6-tuple | 3-ply | 474167 ± 7087 | 99.70 ± 0.15% | 95.41 ± 0.98% | 46.02 ± 2.53% |
| ($V_{init}$=320k) | 5-ply | 485588 ± 30802 | **100.00 ± 0.00%** | 97.00 ± 4.00% | 53.00 ± 9.70% |
| 2-stage | 1-ply | 386078 ± 17104 | 97.37 ± 0.23% | 85.25 ± 1.23% | 26.67 ± 1.64% |
| 8×6-tuple | 3-ply | 506486 ± 50356 | 99.66 ± 0.11% | 96.89 ± 0.32% | 55.50 ± 3.07% |
| ($V_{init}$=0) | 5-ply | 516761 ± 67952 | **100.00 ± 0.00%** | 97.40 ± 3.35% | 56.80 ± 8.53% |
| 2-stage | 1-ply | 404288 ± 2583 | 97.25 ± 0.11% | 85.37 ± 0.39% | 30.17 ± 0.48% |
| 8×6-tuple | 3-ply | 538582 ± 5692 | 99.70 ± 0.12% | 96.83 ± 0.32% | 57.58 ± 0.81% |
| ($V_{init}$=320k) | 5-ply | 581896 ± 14765 | 99.80 ± 0.89% | **98.60 ± 2.68%** | 66.40 ± 6.42% |
| 3-stage | 1-ply | 412492 ± 3666 | 97.23 ± 0.13% | 85.34 ± 0.35% | 33.55 ± 0.91% |
| 8×6-tuple | 3-ply | 545231 ± 6479 | 99.66 ± 0.12% | 96.90 ± 0.44% | 60.48 ± 0.19% |
| ($V_{init}$=320k) | 5-ply | **588426 ± 30723** | 99.80 ± 0.89% | 98.20 ± 2.19% | **72.80 ± 8.17%** |

The 4×6-tuple and the 8×6-tuple networks were trained with OTC and OTD+TC, respectively. Networks initialized with $V_{init}$ = 0 or 320k; OTD+TC learned with $P_{TC}$ = 10%.

### 3.4.3 Tile-Downgrading

*Tile-downgrading expectimax search* is a technique that searches states by translating them into downgraded states. Let us illustrate it by an example in Figure 15, where (a) is an original state and (b) is its *downgraded state*. The downgraded state (b) is derived by halving all tile values larger than a missing tile from (a), i.e., halving tiles larger than 256-tile. Then, we evaluate the downgraded state instead of the original state. Since afterstates with large tiles have relatively less chance to be trained, their evaluations are relatively less accurate. For the downgraded states, their afterstates have smaller tiles, implying higher chances to be trained and higher accuracy.

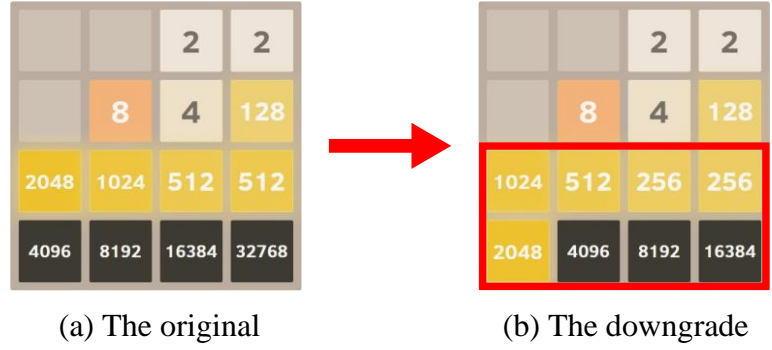

(a) The original    (b) The downgrade

Figure 15. Example for applying tile-downgrading to a state with 32768-tile. The original state (a) is translated to its downgraded state (b), where the highlighted tiles indicate that the values have been halved by tile-downgrading.

The tile-downgrading can only be applied to the root state of a search tree when the following two conditions hold in the root: there exists a missing tile not exceeding the largest tile, and the state is not smaller than the *activation threshold* $\rho$. For Figure 15 (a), the 256-tile is the largest missing tile that does not exceed the largest tile. The tile-downgrading will be activated when $\rho$ = 32768-tile, but will not when $\rho$ = 65536-tile. For a state $s = (c_0, c_1, \dots, c_{15})$ where $c_i$ are tile values, the downgraded state $\tilde{s}$ is produced as follows.

$$\tilde{s} \leftarrow \left(\tilde{c}_0, \tilde{c}_1, \dots, \tilde{c}_{15} \middle| \tilde{c}_i = \begin{cases} c_i \div 2, & \text{if } c_i > c_{\nexists} \\ c_i, & \text{otherwise} \end{cases}\right), \tag{33}$$

where $c_{\nexists} = \max\{c \mid c \notin s \land 0 < c < \max(s)\}$ is the largest missing tile that does not exceed the largest tile of state $s$.

Let a root state $s$ satisfy the conditions. The tile-downgrading expectimax search is to translate $s$ into a downgraded state $\tilde{s}$; then, use the expectimax search to choose the best action for the downgraded state $\tilde{s}$. The downgraded state $\tilde{s}$ will be used as the root node of the search to recommend an action $a$ that with the highest $V_{\max}(\tilde{s}, p)$, where $V_{\max}$ is the expectimax procedure and $p$ is the search depth limit. Note that no more tile-downgrading is required inside the search. Since this technique preserves the puzzle structure while halving some tile values, the best action of $\tilde{s}$ can be directly applied to the original state $s$.

The tile-downgrading expectimax search was evaluated on the 2-stage 8×6-tuple network trained with OTD+TC and $V_{init}$ = 320k. Table 9 summarizes the results of using different

activation thresholds $\rho$, including 16384-tile + 8192-tile, 32768-tile, 32768-tile + 8192-tile, and 32768-tile + 16384-tile. The performance was even worse with low thresholds, e.g., $\rho <$ 32768-tile, which indicates that states with 16384-tile as the largest tile had been sufficiently trained, thereby downgrading them did not help. With large thresholds, e.g., $\rho >$ 32768-tile + 8192-tile, the improvement was limited since only a few episodes could reach it. For the 2-stage 8×6-tuple trained with OTD+TC, the most effective thresholds were between 32768-tile and 32768-tile+8192-tile, with more than 600000 points and about a 70% rate of reaching 32768-tiles on average. Interestingly, this technique also significantly improved the chance of creating 65536-tiles to roughly 0.02% with the 3-ply search, which is the highest 65536-tile reaching rate to the best of our knowledge. Note that 65536-tiles were not reached for the 5-ply search since insufficient episodes were evaluated for it.

Table 9. Performance of multistage optimistic learning with tile-downgrading expectimax search in the 2-stage 8×6-tuple network.

| $\rho$ | Search | Average Score | 8192 [%] | 16384 [%] | 32768 [%] |
|---|---|---|---|---|---|
| 16384-tile +8192-tile | 1-ply | 352951 ± 2620 | 97.24 ± 0.10% | 85.36 ± 0.42% | 7.97 ± 0.82% |
| | 3-ply | 437012 ± 3499 | 99.70 ± 0.11% | 96.78 ± 0.45% | 19.64 ± 1.30% |
| | 5-ply | 432141 ± 45177 | **100.00 ± 0.00%** | 97.80 ± 4.34% | 16.20 ± 13.22% |
| 32768-tile | 1-ply | 412785 ± 2208 | 97.24 ± 0.12% | 85.39 ± 0.35% | 30.16 ± 0.38% |
| | 3-ply | 563316 ± 5650 | 99.63 ± 0.19% | 96.88 ± 0.12% | 57.90 ± 1.63% |
| | 5-ply | **608679 ± 42177** | 99.80 ± 0.89% | 97.80 ± 1.67% | **67.40 ± 12.21%** |
| 32768-tile +8192-tile | 1-ply | 411724 ± 2945 | 97.26 ± 0.14% | 85.36 ± 0.38% | 30.19 ± 0.42% |
| | 3-ply | 560366 ± 7306 | 99.73 ± 0.09% | 96.86 ± 0.23% | 57.55 ± 1.45% |
| | 5-ply | 603254 ± 32902 | **100.00 ± 0.00%** | **98.80 ± 0.89%** | 65.60 ± 4.60% |
| 32768-tile +16384-tile | 1-ply | 409164 ± 3122 | 97.24 ± 0.14% | 85.37 ± 0.42% | 30.20 ± 0.43% |
| | 3-ply | 554005 ± 3494 | 99.73 ± 0.09% | 96.76 ± 0.34% | 57.20 ± 1.14% |
| | 5-ply | 580322 ± 50336 | **100.00 ± 0.00%** | 98.60 ± 1.10% | 59.60 ± 16.04% |

The 2-stage 8×6-tuple network analyzed in this experiment was the same as in Table 8.

## 3.5 Chapter Conclusion

We have demonstrated that optimistic methods perform significantly well for 2048. In the final section of this chapter, we will compare our methods to other work, discuss related methods, and make a conclusion about the optimistic methods. Section 3.5.1 compares this work with previous state-of-the-art. Section 3.5.2 discusses the related methods. Section 3.5.3 summarizes the optimistic method proposed in this work.

### 3.5.1 Comparison to SOTA

In order to compare with the previous SOTA [5], [54], we used the same network, namely the 5×6-tuple network (Figure 4) with 2-stage OTC and tile-downgrading search. Additionally, we used a larger network, the 8×6-tuple network (Figure 5) with 2-stage OTD+TC and tile-downgrading search. We tested 1-ply, 3-ply, and 5-ply searches for these networks, and the results are shown in Table 10. From Table 10, our methods outperformed the 16-stage 5×6-tuple network with TC in [5] in terms of fixed-depth search.

Table 10. Comparison of the SOTA optimistic method and the previous SOTA method.

| Author | Method | Weights | Search | Average Score | 32768 [%] |
|---|---|---|---|---|---|
| Jaśkowski [5] | MS-TC 16-stage 5×6-tuple | 1342.2M | 1-ply | 324710 ± 11043 | 19% |
| | | | 3-ply | 511759 ± 12021 | 50% |
| | | | 5-ply | 545833 ± 21500 | 54% |
| | | | 1000ms | 609104 ± 38433 | 70% |
| This work | MS-OTC 2-stage 5×6-tuple with tile-downgrading | 167.8M | 1-ply | 331073 ± 6699 | 15.38 ± 0.62% |
| | | | 3-ply | 526178 ± 21752 | 51.07 ± 3.87% |
| | | | 5-ply | 574245 ± 25727 | 60.60 ± 5.02% |
| | MS-OTD+TC 2-stage 8×6-tuple with tile-downgrading | 268.4M | 1-ply | 412785 ± 2208 | 30.16 ± 0.38% |
| | | | 2-ply | 518027 ± 5622 | 49.84 ± 1.07% |
| | | | 3-ply | 563316 ± 5650 | 57.90 ± 1.63% |
| | | | 4-ply | 571386 ± 27809 | 60.02 ± 4.09% |
| | | | 5-ply | 608679 ± 42177 | 67.40 ± 12.21% |
| | | | 6-ply | **625377 ± 40936** | **72.00 ± 12.00%** |

Furthermore, for the 2-stage 8×6-tuple network, we tested an additional 6-ply search with 100 episodes. The average search speed was about 2.5 moves/s by using a single thread of an Intel Core i9-7960X. The 2-stage OTD+TC 8×6-tuple network with 6-ply tile-downgrading expectimax search achieved an average score of 625377 and a rate of 72% to reach 32768-tiles, which is also superior to those with the 1000ms search in [5]. To our knowledge, the results are SOTA among all learning methods in terms of average scores and the reaching rates of large tiles. Although we used three more 6-tuples, we only required 2-stage compared to their work with 16-stage. Since each extra stage requires one more set of tuples, we use far less memory in total, namely 16 (2×8) 6-tuples versus 80 (16×5) 6-tuples in [5].

However, compared with previous methods, the proposed OI methods require significantly more computing resources to explore large state spaces during training. The previous SOTA result [5] was trained with $4 \times 10^{10}$ actions, corresponding to approximately 4M episodes or less. In contrast, our SOTA result was trained with 200M episodes. Nevertheless, such an amount of training is still affordable. It took us 66 hours to complete using our C++ implementation[2] on an Intel E5-2698 v4 processor.

### 3.5.2 Discussion

In this section, we first discuss the improvements of each technique used in the SOTA result. We then discuss other techniques related to 2048 not covered in this work. Finally, we discuss the application of $\epsilon$-greedy and softmax on the game of 2048.

Table 11. Improvement of each technique to the best result.

| Method | Average Score | 32768 [%] |
| --- | --- | --- |
| V0 = 1-stage 8×6-tuple TD ($V_{init}$=0) | 309208 ± 5788 | 0.00 ± 0.00% |
| V1 = V0 + OI ($V_{init}$=320k) | 361471 ± 5473 | 21.75 ± 0.74% |
| V2 = V1 + TC fine-tuning ($P_{TC}$=10%) | 370194 ± 4366 | 22.18 ± 0.92% |
| V3 = V2 + multistage (2-stage) | 404288 ± 2583 | 30.17 ± 0.48% |
| V4 = V3 + expectimax search (6-ply) | 586583 ± 17043 | 65.40 ± 2.28% |
| V5 = V4 + tile-downgrading | **625377 ± 40936** | **72.00 ± 12.00%** |
| X1 = V5 − OI | 592390 ± 39073 | 63.60 ± 10.35% |
| X2 = V5 − TC fine-tuning | 574779 ± 22280 | 59.80 ± 7.40% |
| X3 = V5 − multistage | 574150 ± 29315 | 59.80 ± 10.14% |
| X4 = V5 − expectimax search | 412785 ± 2208 | 30.16 ± 0.38% |

V0–V5 present the incremental improvements to the best design (V5);
X1–X4 present the ablation studies of the best design.

Table 11 shows the improvements of each technique from the non-optimistic method (V0) to the best design (V5) in the upper six methods and ablation studies (X1 to X4) in the lower four methods. V0–V3 present results trained by incrementally improved learning algorithms: V0 was trained by the standard TD learning; V1 was trained by OTD learning with a manually reduced learning rate; V2 was trained by OTD+TC learning where TC adjusts the learning rate; V3 was trained by 2-stage OTD+TC learning. V3–V5 present the performance of the same model evaluated by different settings: V3 was evaluated without additional search; V4 applied an expectimax search; V5 further applied tile-downgrading to the search. To summarize, the best design, V5, adopted a 2-stage 8×6-tuple network trained by OTD+TC and evaluated by tile-downgrading expectimax search. X1–X4 present the ablation studies of each technique: X1 was trained by TD+TC without OI; X2 was trained by OTD with a manually reduced learning

---
[2] Available at https://github.com/moporgic/TDL2048/

rate; X3 used only a 1-stage network; X4 was evaluated with only 1-ply tile-downgrading search.

This work does not cover several techniques proposed by the previous SOTA, including *weight promotion* [5], [27], *redundant encoding* [5], and *carousel shaping* [5]. Weight promotion speeds up multistage training by allowing the weights of $n$-tuple networks to be copied to the corresponding weights in the next stage upon its first access. This method is similar to OI in the sense that many weights are initialized higher than zeros. However, this method only copies weights with small tiles, and those weights with large tiles, say 16384-tile in the next stage, are still initialized to zeros. In OI, these weights are large numbers leading to more exploration. Redundant encoding improves the $n$-tuple network by adding sub-tuples, which may be applied together with OI using an initialization scheme different from that in (32). Finally, carousel shaping ensures that later game stages receive more training, which will also work with OI to improve the performance of playing with large tiles. Since this chapter focuses on the impact of OI, further research on incorporating these methods into OI is left open.

$\epsilon$-greedy and softmax are both commonly applied exploration techniques in reinforcement learning. Before this work, researchers had already noticed the potential issue of insufficient exploration; therefore, they tried these techniques. According to their reports, these exploration techniques significantly inhibited the learning performance on $n$-tuple networks [2], [5]. However, since complex $n$-tuple network designs were not introduced at the time of their work, it is likely that [2] and [5] focused on only smaller networks such as the 5×6-tuple. Therefore, we also tested $\epsilon$-greedy and softmax on larger networks. The comparison of applying OI, $\epsilon$-greedy, and softmax on the 8×6-tuple are provided in Appendix C. To summarize in short, we confirmed that both $\epsilon$-greedy and softmax are still inappropriate on the 8×6-tuple network.

### 3.5.3 Conclusion

For the issue of exploration on 2048, we propose optimistic TD learning to improve the performance. Our approach significantly improves the learning quality of $n$-tuple networks. The significance of this work is summarized as follows.

First, we improve TD methods with OI by using a hyperparameter $V_{init}$ to initialize the network weights, and demonstrate that the proposed OTD and OTC learning with $V_{init}$ = 320k significantly improves the performance, especially the chance of obtaining 32768-tiles, as shown in Section 3.3.1. Second, we propose OTD+TC learning, a hybrid learning paradigm that combines the advantages of both OTD and OTC to encourage exploration and exploitation, respectively. A hyperparameter $P_{TC}$ is used to control the proportion of TC fine-tuning phase. We observe that OTD+TC with $V_{init}$ = 320k and $P_{TC}$ = 10% outperforms both OTD and OTC learning for larger networks, as described in Section 3.3.2. Third, we show that we do not need as many stages as in [5], thereby significantly reducing the required network weights, since OI effectively improves the learning quality, as shown in Sections 3.4.1 and 3.4.2.

Furthermore, we use tile-downgrading to improve the search quality, as shown in Section 3.4.3. Finally, the design in Section 3.5.1 is shown to outperform those in the previous SOTA [5] in terms of average score and 32768-tile reaching rate. Namely, the design with a 2-stage 8×6-tuple network trained by OTD+TC with $V_{init}$ = 320k and $P_{TC}$ = 10% achieved an average score of 625377 and a rate of 72% reaching 32768-tiles through 6-ply tile-downgrading expectimax search. In addition, 65536-tiles were reached at a rate of 0.02%.

We conclude that learning with explicit exploration for 2048 and similar games is useful even if the environment seems stochastic enough. This chapter demonstrates that the optimistic method is promising for stochastic games.

# Chapter 4   Strength Enhancements and Adjustments

This chapter includes preliminary improvements and attempts for 2048, which cover enhancing the performance by *ensemble learning*, evaluating *Monte Carlo tree search* with $n$-tuple networks, and applying *deep reinforcement learning* algorithms. These approaches have shown promise in preliminary experiments and may become an essential technique for the next state-of-the-art of 2048.

## 4.1  Motivation

Many topics still require further investigation for the game of 2048. First, multiple network snapshots taken during the long training may have better use rather than being discarded directly. Second, the expectimax search is saturated at around 5-ply and 6-ply, while a further deeper expectimax search is infeasible due to the massive amount of search tree nodes. Third, the $n$-tuple network relies on the feature designed by human experts, and it does not capture the whole puzzle. These open topics motivate us to improve our designs for even better performance.

The rest of this chapter is organized as follows. Section 4.2 proposes the use of ensemble learning to further improve the performance of $n$-tuple networks. Section 4.3 presents the results of applying Monte Carlo tree search with $n$-tuple networks to 2048. Section 4.4 shows our attempts at applying deep reinforcement learning to 2048. Section 4.5 summarizes these improvements and attempts.

## 4.2  N-Tuple Network Ensemble

*Ensemble learning* is a classical method to improve performance by multiple predictors. In this section, we examine the ensemble of the $n$-tuple network weights by *stochastic weight averaging (SWA)*. SWA improves the performance by averaging multiple networks into a single network, which has neither additional training cost nor runtime cost. Our preliminary experiments show that SWA works well when applied to existing network snapshots taken during a single training trial. Furthermore, by integrating SWA into the training procedure, the performance of the ensemble network is significantly improved. For example, a 1-stage 8×6-tuple network improved by SWA matches the performance of the current state-of-the-art 2-stage 8×6-tuple network.

### 4.2.1 Motivation

Many network snapshots are available during the long learning process. These networks perform similarly in terms of the average score; however, they may have different blind spots related to different states. This assumption motivates us to use ensemble learning [62], [63] to utilize these network snapshots. A commonly used ensemble learning method, *bagging* [62], [63], may achieve this. The bagging method uses majority voting of multiple network outputs to determine the final output. However, it increases the runtime cost to $k$ times for generating the output using $k$ networks, which is not desired. Therefore, we focus on another method, stochastic weight averaging (SWA) [64], to integrate all networks to improve performance.

### 4.2.2 Stochastic Weight Averaging

Stochastic weight averaging (SWA) is a simple but effective method for training neural networks, proposed by Izmailov *et al.* in 2018 [64]. SWA averages multiple neural networks into one network to improve the generalization, which makes it extremely easy to implement and has no computational overhead. SWA has been shown to achieve notable improvement over many network types and has been applied to game-playing programs such as Go [65]. Two principles are essential for improving performance using SWA.

For generalization, the multiple networks used for averaging should be different, i.e., not stuck in the same local optimum. A solution is to increase and decrease the learning rate regularly, which is known as the *cyclical learning rate* [64], [66], [67]. The cyclical learning rate works well with SWA since it generates a lot of saturated network snapshots with different local optimums by taking snapshots at the time when the learning rate is the lowest. For 2048 and $n$-tuple networks, TC learning has already been demonstrated to be an effective method to automatically decrease the learning rate. To apply cyclical learning rate to TC learning, we propose to periodically reset the coherence, i.e., reset both the $E$ and $A$ introduced in Section 2.3. Resetting the coherence immediately resets the learning rate to 1.0. Therefore, it allows the network to escape from a local optimum, thereby improving the diversity of the network snapshots.

On the other hand, it is necessary to train the network with all the required data. For example, if all networks have never seen 32768-tiles during training, their ensemble can still not play with 32768-tiles. Several proposed methods for 2048 are available, e.g., *carousel* *shaping* [5], [55], *restart strategy* [23], and *jump-start strategy* [9]. These methods are designed from the same observation that only a few games can reach large tiles, e.g., 32768-tiles. Therefore, to balance the training data, the solution is to explicitly start new episodes from the states with large tiles, where the states with large tiles are either collected in the previous training or generated by the heuristic. Using these methods ensures that the network adequately learns the states that only appear in the later game phases, even if the learning rate is kept high.

When applying this technique to $n$-tuple networks, each corresponding feature weight is averaged. For $k$ lookup tables $\text{LUT}_1, \text{LUT}_2, \ldots, \text{LUT}_k$ with the same $n$-tuple definitions, an ensemble network is generated by

$$\text{LUT}_\Sigma[i] \leftarrow \frac{1}{k} \sum_{j=1}^{k} \text{LUT}_j[i] \text{ for all } i, \tag{34}$$

where $\text{LUT}_j[i]$ accesses the feature weight in the $j$th lookup table with an index $i$. Note that the feature weights are merged element by element, so the number of features in the ensemble network is the same as in the original networks.

In the rest of this section, two sets of preliminary experiments are performed and analyzed. First, Section 4.2.3 applies SWA to existing network snapshots. Then, Section 4.2.4 proposes a new training procedure with SWA according to the critical principles mentioned above.

### 4.2.3 SWA for Existing Networks

To analyze the effectiveness of SWA, we take three collections of network snapshots from the experiments of optimistic methods as source networks. Each collection of network snapshots was trained using different settings. Finally, we evaluate the performance of all ensembles averaged from the last $k$ snapshots, where $k$ = 1, 2, 3, …, to the total number of snapshots in the target collection. Note that snapshots were ordered according to the generation time; therefore, later snapshots generally performed better. Table 12 lists the training settings of source networks and the ensemble performance.

SWA improves the performance in all tested cases, regardless of whether the learning rate is decayed. Figure 16, Figure 17, and Figure 18 plot the detailed ensemble results of the three tested cases, respectively. In these figures, a blue dot at the index $i$ represents the original performance of the $i$th source network snapshots; an orange dot at the index $i$ represents the performance of the ensemble network averaging from the $i$th to the last snapshots, e.g., the orange dot at index 80 is the ensemble performance of 21 snapshots from the 80th to the 100th.

Table 12. Performance of ensemble learning for existing network snapshots.

| Network | Training Method | Learning Rate | Original | Ensemble | Best $k$ |
|---|---|---|---|---|---|
| 8×6-tuple | OTD | 0.1 (fixed) | 283734 | 361075 | 44 |
| | OTD+TC | 0.1 + TC | 374116 | 380025 | 10 |
| 4×6-tuple | OTC | 1.0 + TC | 282296 | 283632 | 7 |

Networks were trained with the optimistic methods introduced in Chapter 3, where OTD adopted a fixed learning rate; OTD+TC and OTC adopted adaptive learning rates.

For a collection of network snapshots trained with a fixed high learning rate, SWA significantly improves the ensemble performance, as shown in Figure 16. Due to the high learning rate, the source networks are less likely to be stuck in the same local optimum. That is, they have high diversity, so averaging them improves significantly. The best ensemble network is the one that averages the last 44 snapshots, i.e., all snapshots from the 57th to the 100th. For more than 44 snapshots, as the snapshots from the early stage do not perform well, adding them decreases the performance. On the other hand, while the best ensemble network has a significantly higher average score, it still performs poorly for states with 32768-tile since the source networks were not fully trained for these states.

For snapshots taken from a previous training where the learning rate was reduced, SWA still improves the performance. Figure 17 shows the detailed ensemble results of the 8×6-tuple OTD network with decayed learning rates. Note that the learning rate was kept high before the timestep started to decay; only snapshots after reducing the learning rate were taken for the ensemble. For the source network, reducing the learning rate improved the performance but reduced the diversity. Therefore, the improvement in applying SWA is not as significant as the previous one. The best ensemble network is generated by averaging all 10 snapshots. Although the original performance seems saturated, averaging them still helps improve the generalization.

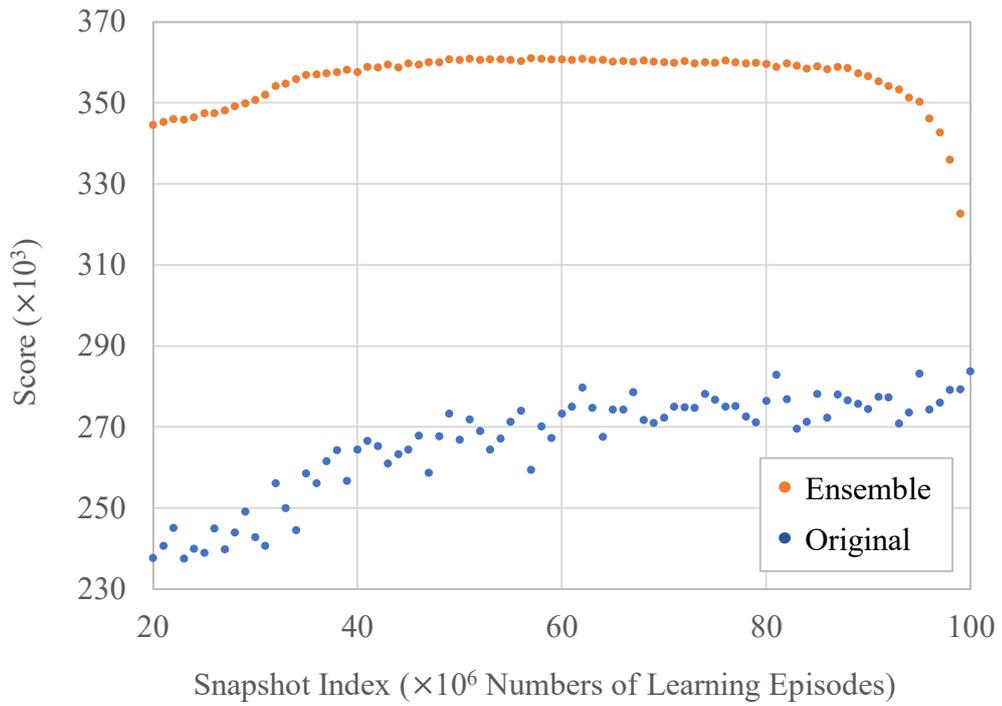

Figure 16. Performance of SWA for OTD learning in the 8×6-tuple network. Note that the learning rate for OTD was not manually reduced in this experiment.

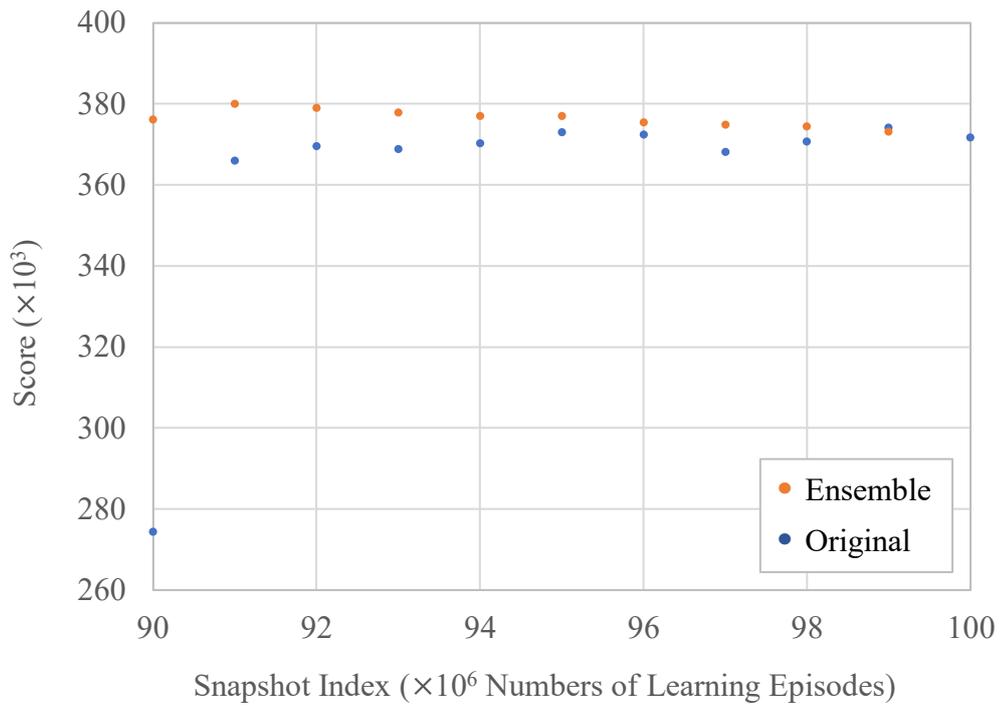

Figure 17. Performance of SWA for OTD+TC learning in the 8×6-tuple network. Note that the TC phase took over the training at index 91.

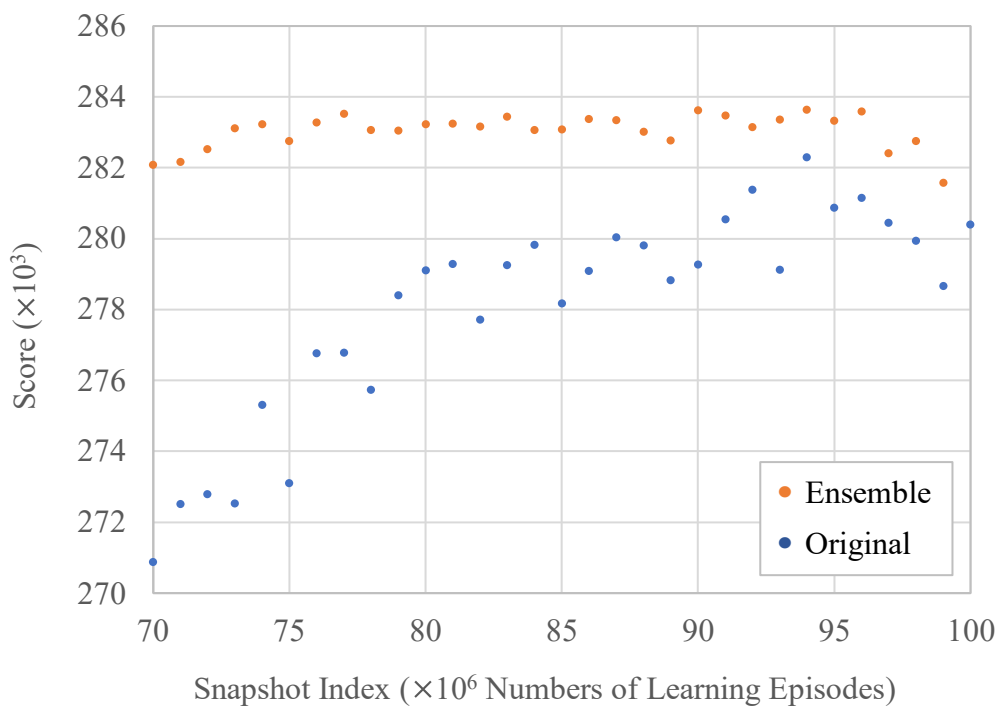

Figure 18. Performance of SWA for OTC learning in the 4×6-tuple network.

As shown in Figure 18, SWA also slightly improves the performance of a collection of snapshots whose learning rate was automatically reduced by TC learning. The best ensemble network is from the last 7 snapshots, while the performance is quite similar from averaging the last 5 to averaging the last 25 snapshots. The improvement is the least significant among the three snapshot collections. We assume that the learning rate was too low, reducing the diversity.

### 4.2.4 SWA for Training Networks

Based on the previous experiments, we modify the training procedure of the optimistic methods to take advantage of SWA. The modified OTC learning is as follows. First, we apply the cyclical learning rate to TC learning by coherence resetting. The learning rate is initially 1.0 and is modulated by TC learning. For every 1M training episodes, a snapshot is taken, and then the coherence is reset. Second, we apply a variant of carousel shaping [55] to ensure sufficient training states with 32768-tiles. This modified procedure allows the network to be thoroughly trained and prevents it from being stuck into the same local optimum. Thus, the snapshots taken should be optimal regarding diversity and strength.

The results of a 1-stage 8×6-tuple network trained with this procedure with a total of 400M episodes are listed in Table 13. With the improved training procedure, a 1-stage network outperforms the previous 1-stage network. Even more, it matches the performance of the 2-stage network currently in the SOTA.

Table 13. Comparison of OTD+TC learning without and with SWA in the 8×6-tuple network.

| Network | Search | Average Score | 8192 [%] | 16384 [%] | 32768 [%] |
|---|---|---|---|---|---|
| 2-stage 8×6-tuple w/o SWA | 1-ply | 412785 ± 2208 | 97.24 ± 0.12% | 85.39 ± 0.35% | 30.16 ± 0.38% |
| | 2-ply | 518027 ± 5622 | 99.24 ± 0.18% | 94.69 ± 0.36% | 49.84 ± 1.07% |
| | 3-ply | 563316 ± 5650 | 99.63 ± 0.19% | 96.88 ± 0.12% | 57.90 ± 1.63% |
| | 4-ply | 571386 ± 27809 | 99.68 ± 0.26% | 97.66 ± 1.32% | 60.02 ± 4.09% |
| | 5-ply | 608679 ± 42177 | **99.80 ± 0.89%** | 97.80 ± 1.67% | 67.40 ± 12.21% |
| | 6-ply | **625377 ± 40936** | **99.80 ± 0.89%** | **98.80 ± 0.89%** | **72.00 ± 12.00%** |
| 1-stage 8×6-tuple w/ SWA | 1-ply | 415945 | 96.67% | 85.01% | 30.42% |
| | 2-ply | 506938 | 98.37% | 93.17% | 47.96% |
| | 3-ply | 558938 | 99.54% | 96.40% | 56.91% |
| | 4-ply | 579921 | 99.58% | 97.54% | 60.64% |
| | 5-ply | 600376 | 99.70% | 97.10% | 66.00% |
| | 6-ply | 607403 | 99.60% | 98.30% | 67.50% |

The 2-stage 8×6-tuple network trained without SWA is the current SOTA in Table 10.

### 4.2.5 Discussion

In this section, we demonstrate that SWA is promising for $n$-tuple networks. And shows that this method is promising for $n$-tuple network for 2048. Our preliminary experiments show that a 1-stage 8×6-tuple network has comparable performance to the current SOTA 2-stage 8×6-tuple network. However, more analysis is necessary before the wide use of this method. For example, a more effective training paradigm based on SWA is also possible. Furthermore, averaging snapshots from multiple individually trained networks is also worth trying since it may allow the ensemble network to generalize better. With more fine-tuning, it is promising that SWA will become an essential technique for training the next generation of the SOTA network.

## 4.3 Monte Carlo Tree Search

*Monte Carlo tree search (MCTS)* is an important search method widely applied to many applications [44], [45]. This section integrates MCTS with the $n$-tuple network and the TD-afterstate learning framework for 2048. First, we use an existing $n$-tuple network as the value function of MCTS. The results are comparable to the current expectimax search. Furthermore, we use MCTS to train new networks, which confirms that training with MCTS is promising. Surprisingly, the results show that the performance generally decreases as the training simulation count increases, which may be caused by the stochasticity of 2048; however, it requires more analysis to be confirmed.

### 4.3.1 Motivation

Recently, the SOTA of 2048 is achieved with a 6-ply expectimax search. However, since expectimax search is not a best-first search algorithm, it expands all the subtrees to the search limit to obtain a reliable result, which results in a massive number of nodes for deep searches. As shown in Table 14, there are 4.4 million nodes on average for a 5-ply expectimax search. Searching for many nodes is inefficient because many originate from low-quality actions and should be pruned. Even more, low-quality actions are less trained, and their corresponding afterstate values are relatively inaccurate, which may interfere with the search result for deep searches. It is observed that the expectimax search is converged at around 5-ply or 6-ply. A deeper search, i.e., using iterative deepening, does not improve the performance very well.

Monte Carlo tree search (MCTS) is a commonly used best-first search algorithm [44], [45], in which the *upper confidence bounds (UCB)* function balances exploration and exploitation, leading the search to explore promising moves more frequently. Therefore, it is promising to replace expectimax search with MCTS to improve the performance further.

Table 14. Average numbers of nodes in the expectimax search tree with different depths.

| Search Depth | Average Number of Nodes | Average Maximum Number of Nodes |
| --- | --- | --- |
| 1-ply | 4.56 ± 1.19 | 4.79 ± 0.90 |
| 2-ply | 139.04 ± 97.86 | 511.18 ± 123.43 |
| 3-ply | 4383.83 ± 5169.29 | 51942.03 ± 13114.39 |
| 4-ply | 139966.40 ± 235342.91 | 5051955.46 ± 1302818.05 |
| 5-ply | 4419503.73 ± 9903876.80 | 117733700.06 ± 109407686.68 |

From 1000 episodes played by a well-trained 8×6-tuple network.

### 4.3.2 MCTS for 2048

Monte Carlo tree search uses several iterations to determine the best action to play, where each iteration consists of four phases: selection, expansion, simulation, and backpropagation [44], [45], which has been introduced in Section 2.6. However, to use MCTS with the $n$-tuple network afterstate value function for 2048, modifications are required as proposed below.

In the selection phase, the max and chance nodes require different selection policies. At the max node, i.e., a state $s$, the child node is selected by the upper confidence bounds (UCB) function as in (23). At the chance node, i.e., an afterstate $s'$, the child node is selected randomly according to the probabilities. In the expansion phase, all child nodes of the selected node are generated. Generating all at once is fine since the average branching factor is not very large. Note that in Stochastic MuZero [11], a variant of the PUCT function [50] is used for 2048, which is not applicable to our work since we do not have a policy function.

In the simulation phase, instead of simulating the game to a terminal state, the normalized value $(r_\Sigma + V(s'))/V_{norm}$ is directly evaluated, where $r_\Sigma$ is the cumulative reward during the selection; $V(s')$ is the value from a trained $n$-tuple network; and $V_{norm}$ is the *normalization constant* that scales the value for MCTS. Note that there is no need to limit the maximum normalized value $(r_\Sigma + V(s'))/V_{norm}$ to 1. However, the $n$-tuple network provides only afterstate values $V(s')$. Thus, to evaluate the value $V(s)$ for a newly expanded state $s$, an additional layer of afterstates needs to be further expanded to calculate $V(s) = \max_a(r + V(s'))$. Therefore, the selected node is evaluated instead of the newly expanded node. With this modification, afterstates that just have been expanded are available for evaluating the selected state. In the backpropagation phase, all selected nodes are updated with the normalized value of the evaluated node.

Figure 19 illustrates the MCTS with the afterstate value function for 2048. First, as shown in (a), MCTS traverses a path from the root node $s_{(0)}$ to the leaf node $s_{(1)}$ using the described selection mechanism. In the case that the selected leaf node $s_{(1)}$ is a state as in (b), all afterstates of $s_{(1)}$ are expanded and evaluated by the value function. Then, the statistics of nodes on the selection path are updated using a normalized value $(r_{(0)} + V(s_{(1)}))/V_{norm}$, where $r_{(0)}$ is the only reward that appears in the selection path, and $V(s_{(1)}) = \max_{a_{(1)}}(r_{(1)} + V(s'_{(1)}))$ is the value of the leaf node. On the other hand, at the next iteration, as shown in (c), MCTS selects an afterstate leaf node $s'_{(0)}$. In this case, all possible subsequent states of $s'_{(0)}$ are also expanded. However, since the afterstate value $V(s'_{(0)})$ must have already been evaluated when $s'_{(0)}$ is expanded by its parent node, the statistics are directly updated using $(r_{(0)} + V(s'_{(0)}))/V_{norm}$. To summarize, MCTS always updates the tree with the value of the selected leaf node, while the value may be either from a newly expanded afterstate or from the selected node itself.

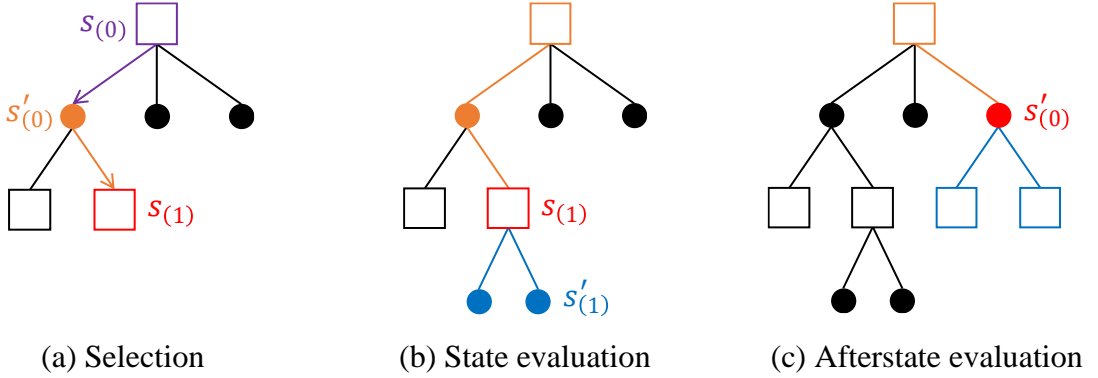

(a) Selection (b) State evaluation (c) Afterstate evaluation

Figure 19. MCTS selection and leaf node evaluation for states and afterstates in 2048. (a) MCTS traverses a path from the root node $s_{(0)}$ to the leaf node $s_{(1)}$. (b) Since the selected leaf node is a state, MCTS expands all its afterstates and evaluates its win rate from the best afterstate value: $V(s_{(1)}) = \max(r_{(1)} + V(s'_{(1)}))$. (c) In the next iteration, MCTS selects a chance node $s'_{(0)}$ as the leaf node, and expands all possible subsequent states. Then, MCTS directly adopts the afterstate value $V(s'_{(0)})$ as the win rate.

### 4.3.3 MCTS for Existing Networks

In this experiment, we use a well-trained 8×6-tuple network as the value function for MCTS and experiment on the performance with different numbers of simulation $N \in \{1, 10, 20, 50, 140, 500, 1000, 2000, 4400, 10000, 20000, 50000\}$ and different exploration constants $c \in \{0.000, 0.001, 0.002, 0.003, 0.004, 0.005, 0.006, 0.007, 0.008, 0.009, 0.010, 0.020, 0.030, 0.040, 0.050, 0.060, 0.070, 0.080, 0.090, 0.100\}$. The normalization constant $V_{norm} = 400000$ is set according to the average performance of the value function. A grid search was performed, in which each combination of the hyperparameters was evaluated for 1000 games.

The results are summarized in Table 15. Without exploration, i.e., $c = 0$, the search performs well when the number of simulation counts $N$ is low, e.g., $10 < N < 40$; however, when $N$ is higher, a non-zero small $c$ performs the best. We assume this highly correlates with the accuracy of the value function. Since the value function we used has a similar performance to the SOTA result (the network trained in Section 4.2.4), if the simulation count is limited, exploration does not help and will even decrease the performance. Also, note that when the simulation counts $N$ increases, the optimal corresponding exploration constant $c$ also increases. For the value function itself, it achieves 415945, 506938, and 558938 points when using a 1-ply, 2-ply, and 3-ply expectimax search, in which the nodes of the search tree are approximately 1, 140, and 4400, respectively. For $N = 4400$, the MCTS is able to achieve about 555000 points, which is comparable to the result of using the expectimax search.

Table 15. Performance of MCTS in the 8×6-tuple network trained by OTD+TC learning.

| $c$ | \multicolumn{12}{c}{$N$} | | | | | | | | | | | |
|---|---|---|---|---|---|---|---|---|---|---|---|---|
| | 1 | 10 | 20 | 50 | 140 | 500 | 1000 | 2000 | 4400 | 10000 | 20000 | 50000 |
| 0.000 | 41.80 | 43.57 | 44.30 | 46.23 | 48.25 | 49.74 | 50.37 | 51.52 | 50.24 | 51.13 | 51.81 | 51.19 |
| 0.001 | 42.43 | 43.30 | 44.03 | 46.98 | 48.33 | 51.41 | 52.59 | 53.41 | 52.47 | 53.58 | 53.76 | 52.83 |
| 0.002 | 39.87 | 42.00 | 44.57 | 45.28 | 48.73 | 50.75 | 52.88 | 53.86 | 53.06 | 54.15 | 54.72 | 55.52 |
| 0.003 | 41.33 | 40.85 | 43.97 | 45.34 | 48.41 | 51.92 | 52.62 | 53.06 | 53.55 | 55.69 | 54.85 | 55.66 |
| 0.004 | 41.03 | 41.07 | 42.95 | 46.39 | 48.97 | 51.60 | 53.33 | 53.21 | 54.50 | 54.88 | 55.00 | 56.88 |
| 0.005 | 42.31 | 40.39 | 42.42 | 46.40 | 48.80 | 51.96 | 53.90 | 55.30 | 55.35 | 55.30 | 56.10 | 56.11 |
| 0.006 | 40.49 | 40.73 | 41.58 | 45.56 | 49.23 | 52.79 | 52.59 | 54.95 | 55.73 | 55.77 | 56.07 | 57.39 |
| 0.007 | 40.42 | 40.15 | 42.10 | 45.55 | 49.27 | 51.73 | 54.03 | 54.44 | 54.92 | 55.49 | 54.55 | 56.27 |
| 0.008 | 41.62 | 40.13 | 41.42 | 45.29 | 49.07 | 52.19 | 54.03 | 54.97 | 55.72 | 56.38 | 55.80 | 57.04 |
| 0.009 | 41.77 | 40.83 | 41.11 | 45.96 | 49.90 | 51.52 | 53.41 | 54.52 | 54.73 | 54.29 | 55.72 | 56.74 |
| 0.010 | 41.43 | 41.16 | 41.29 | 44.06 | 48.99 | 52.28 | 53.15 | 54.50 | 55.98 | 56.16 | 55.13 | 55.99 |
| 0.020 | 40.85 | 39.55 | 37.93 | 43.06 | 46.24 | 51.01 | 52.62 | 54.12 | 54.08 | 55.17 | 56.88 | 56.54 |
| 0.030 | 40.90 | 37.96 | 37.08 | 42.13 | 44.96 | 48.58 | 51.99 | 52.44 | 53.44 | 56.09 | 55.56 | 57.10 |
| 0.040 | 40.16 | 37.40 | 36.49 | 39.73 | 44.83 | 48.82 | 50.33 | 51.47 | 52.50 | 53.91 | 54.38 | 55.60 |
| 0.050 | 41.86 | 35.77 | 35.77 | 38.12 | 43.44 | 48.02 | 49.85 | 50.42 | 51.32 | 52.62 | 53.78 | 56.06 |
| 0.060 | 41.33 | 35.78 | 34.36 | 38.18 | 42.75 | 45.99 | 47.99 | 48.75 | 51.88 | 53.64 | 54.51 | 54.63 |
| 0.070 | 41.76 | 35.67 | 34.81 | 36.29 | 42.14 | 45.41 | 47.78 | 47.10 | 50.37 | 51.63 | 52.40 | 53.22 |
| 0.080 | 41.66 | 36.35 | 34.34 | 36.14 | 40.89 | 44.22 | 44.88 | 46.55 | 49.50 | 51.33 | 50.91 | 52.51 |
| 0.090 | 40.99 | 34.92 | 33.26 | 35.30 | 40.79 | 42.24 | 43.31 | 45.06 | 47.47 | 49.44 | 50.68 | 50.33 |
| 0.100 | 42.15 | 35.38 | 31.67 | 35.40 | 39.63 | 42.33 | 42.23 | 43.74 | 46.12 | 48.88 | 49.55 | 50.37 |

$N$ and $c$ denote the simulation count and the exploration constant, respectively. The values express the performance in average scores ($\times 10^4$); the colors reflect the ranking across all tested settings. Note that the reference baseline performance is 41.59, 50.69, 55.89 for expectimax 1-ply, 2-ply, 3-ply; corresponding to $N \approx 1$, 140, 4400, respectively.

### 4.3.4 MCTS for Training Networks

The following experiments are training new networks with MCTS. The experiments were performed in the 4×6-tuple network; each network was trained with TC learning using 1M training episodes. The training algorithm follows the TD-afterstate learning framework, in which the TD target adopts the MCTS value $Q(s_{t+1}, a_{t+1})$ instead of the original value $r_{t+1} + V(s'_{t+1})$. Specifically, the TD error $\delta_t$ of the afterstate $V(s'_t)$ becomes

$$\delta_t = Q(s_{t+1}, a_{t+1})V_{norm} - V(s'_t), \tag{35}$$

where $Q(s_{t+1}, a_{t+1})V_{norm} \approx r_{t+1} + V(s'_{t+1})$ is the TD target; $a_{t+1} = \text{argmax}_a N(s_{t+1}, a)$ is the best action found by MCTS. The proposed algorithm keeps the afterstate values $V(s')$ unnormalized. Therefore, the afterstate values $V(s')$ need to be normalized using $V_{norm}$ during MCTS; and the MCTS values $Q(s, a)$ also need to be unnormalized using the same $V_{norm}$. The normalization constant can be set using the average score of the applied $n$-tuple network. Nevertheless, for the cases where longer training is required or where an average score is unavailable, $V_{norm}$ can be set dynamically for each $Q(s, a)$ by using

$$V_{norm} = \max\left\{\max_a(r + V(s')), 1\right\}, \tag{36}$$

where $\max_a(r + V(s'))$ is the best 1-step return of state $s$ estimates by the $n$-tuple network value function. This approach changes the $V_{norm}$ for each MCTS; however, the value function $V(s')$ remains the same since the TD targets are unnormalized using the same $V_{norm}$.

First, we trained multiple networks by using different simulation counts $\tilde{N} \in \{1, 20, 50, 100, 200, 400, 800\}$ with different exploration constants $\tilde{c} \in \{0.000, 0.001, 0.002, 0.005, 0.010, 0.050, 0.100, 0.500, 1.000, 1.500\}$. Note that the network trained with $\tilde{N} = 1$ is the same as training by the original TC algorithm. For comparing the results trained by different sets of hyperparameters, we set $V_{norm}$ as 250000 based on the average performance of the 4×6-tuple network. Then, we evaluated these networks with MCTS and expectimax search. MCTS evaluations were performed using simulation counts $N \in \{50, 140, 4400\}$ with exploration constants $c \in \{0.000, 0.005, 1.000, 1.500\}$. Expectimax search evaluations were performed using 1-ply, 2-ply, and 3-ply. Note that the values of $N$ are so chosen as the 2-ply and 3-ply expectimax searches have approximately 140 and 4400 nodes, respectively.

The results of performing a grid search on hyperparameters $\tilde{N}, \tilde{c}, N$, and $c$ are listed in Table 16. We found that the networks trained by MCTS may have better performance when evaluated by also MCTS, e.g., trained with $\tilde{N} = 400$ and $\tilde{c} = 1.500$ then evaluated with $N = 4400$ and $c = 1.500$ in Table 16.

Table 16. Performance of MCTS in the 4×6-tuple networks trained by TC learning that adopts MCTS as training target.

| Network | | Evaluation Result | | | | | | | | | | | | | |
|---|---|---|---|---|---|---|---|---|---|---|---|---|---|---|---|
| | | $N=50$, $c=$ (below) | | | | $N=140$, $c=$ (below) | | | | $N=4400$, $c=$ (below) | | | | Expectimax | | |
| $\tilde{N}$ | $\tilde{c}$ | 0.000 | 0.005 | 1.000 | 1.500 | 0.000 | 0.005 | 1.000 | 1.500 | 0.000 | 0.005 | 1.000 | 1.500 | 1-ply | 2-ply | 3-ply |
| 1 | N/A | 22.38 | 19.88 | 1.19 | 0.93 | 24.73 | 23.77 | 5.11 | 3.72 | 29.21 | 31.52 | 15.29 | 12.86 | 20.84 | 28.64 | 31.44 |
| 20 | 0.000 | 21.19 | 18.49 | 0.91 | 0.81 | 23.19 | 22.75 | 3.64 | 2.75 | 26.50 | 29.74 | 13.11 | 11.72 | 17.23 | 26.36 | 30.07 |
| 20 | 0.001 | 25.92 | 22.58 | 1.04 | 0.87 | 28.59 | 26.91 | 4.36 | 3.32 | 32.30 | 34.38 | 14.27 | 12.34 | 20.47 | 31.02 | 34.36 |
| 20 | 0.002 | 21.69 | 19.21 | 0.91 | 0.80 | 23.94 | 23.62 | 3.59 | 2.75 | 27.29 | 30.82 | 13.30 | 11.67 | 16.78 | 27.37 | 31.15 |
| 20 | 0.005 | 24.23 | 21.33 | 1.02 | 0.86 | 26.97 | 25.74 | 4.15 | 3.19 | 30.50 | 33.05 | 14.75 | 12.54 | 18.83 | 29.77 | 33.13 |
| 20 | 0.010 | 21.99 | 20.52 | 1.08 | 0.90 | 24.21 | 24.18 | 4.48 | 3.43 | 27.74 | 31.56 | 15.02 | 12.72 | 17.10 | 28.27 | 32.11 |
| 20 | 0.050 | 14.74 | 14.79 | 1.50 | 1.16 | 16.32 | 16.63 | 5.18 | 4.35 | 18.97 | 20.24 | 17.21 | 15.18 | 9.04 | 16.05 | 18.90 |
| 20 | 0.100 | 14.16 | 14.34 | 1.73 | 1.43 | 15.73 | 16.07 | 5.39 | 4.69 | 18.11 | 18.84 | 18.65 | 16.93 | 8.08 | 15.56 | 18.55 |
| 20 | 0.500 | 13.02 | 13.25 | 2.24 | 2.03 | 14.89 | 15.11 | 7.26 | 6.51 | 17.42 | 17.74 | 18.92 | 18.75 | 6.85 | 14.87 | 17.32 |
| 20 | 1.000 | 13.09 | 13.11 | 3.21 | 2.52 | 14.75 | 14.98 | 8.72 | 7.98 | 17.20 | 17.48 | 18.66 | 18.76 | 6.82 | 14.68 | 17.11 |
| 20 | 1.500 | 12.01 | 12.18 | 5.77 | 3.14 | 14.00 | 14.18 | 9.19 | 8.42 | 16.75 | 16.92 | 18.46 | 18.39 | 6.05 | 13.84 | 16.64 |
| 50 | 0.000 | 26.75 | 26.03 | 2.59 | 1.97 | 29.75 | 29.74 | 9.24 | 7.62 | 33.88 | 35.39 | 24.11 | 22.00 | 11.98 | 31.46 | 35.22 |
| 50 | 0.001 | 26.48 | 26.00 | 2.93 | 2.28 | 29.47 | 29.71 | 10.42 | 8.62 | 33.86 | 35.15 | 25.20 | 22.70 | 11.77 | 31.32 | 35.11 |
| 50 | 0.002 | 26.58 | 26.21 | 3.22 | 2.50 | 29.75 | 29.77 | 10.99 | 9.32 | 34.02 | 34.90 | 27.36 | 23.83 | 11.82 | 31.44 | 35.18 |
| 50 | 0.005 | 26.40 | 26.68 | 4.08 | 3.35 | 29.79 | 29.93 | 13.34 | 11.46 | 33.95 | 34.86 | 31.56 | 27.48 | 11.54 | 31.29 | 35.07 |
| 50 | 0.010 | 26.42 | 26.53 | 4.62 | 4.02 | 29.68 | 29.95 | 15.41 | 13.43 | 34.02 | 34.64 | 33.58 | 31.70 | 11.42 | 31.33 | 35.01 |
| 50 | 0.050 | 26.17 | 26.53 | 12.42 | 8.72 | 29.75 | 29.88 | 22.45 | 20.25 | 33.88 | 34.24 | 35.57 | 35.12 | 11.41 | 31.24 | 35.00 |
| 50 | 0.100 | 26.36 | 26.65 | 16.64 | 14.32 | 29.84 | 29.89 | 26.36 | 23.97 | 34.07 | 34.28 | 36.12 | 35.79 | 11.55 | 31.46 | 35.28 |
| 50 | 0.500 | 26.16 | 26.33 | 24.98 | 23.66 | 29.35 | 29.60 | 29.02 | 28.35 | 33.94 | 33.96 | 35.07 | 35.48 | 11.67 | 31.03 | 34.79 |
| 50 | 1.000 | 26.17 | 26.39 | 25.95 | 25.35 | 29.45 | 29.62 | 29.36 | 29.19 | 33.80 | 33.91 | 34.71 | 34.84 | 11.79 | 31.16 | 34.78 |
| 50 | 1.500 | 26.20 | 26.39 | 26.32 | 26.01 | 29.43 | 29.79 | 29.82 | 29.75 | 33.86 | 34.05 | 34.83 | 35.02 | 11.66 | 31.19 | 34.78 |
| 100 | 0.000 | 26.10 | 25.65 | 3.00 | 2.28 | 29.99 | 29.58 | 10.51 | 8.44 | 34.42 | 35.44 | 26.04 | 22.80 | 8.16 | 31.83 | 35.27 |
| 100 | 0.001 | 26.52 | 25.80 | 3.15 | 2.46 | 30.17 | 30.10 | 10.88 | 9.09 | 34.43 | 35.28 | 26.72 | 23.49 | 8.03 | 31.51 | 35.02 |
| 100 | 0.002 | 26.72 | 26.73 | 3.86 | 3.06 | 30.38 | 30.31 | 12.71 | 10.63 | 34.45 | 35.35 | 30.22 | 25.73 | 8.20 | 31.88 | 35.27 |
| 100 | 0.005 | 26.50 | 26.40 | 4.52 | 3.78 | 30.24 | 30.28 | 14.33 | 12.47 | 34.59 | 35.02 | 32.74 | 29.49 | 8.10 | 31.37 | 35.08 |
| 100 | 0.010 | 26.22 | 26.53 | 6.07 | 5.17 | 30.27 | 30.59 | 17.81 | 15.91 | 34.57 | 34.98 | 34.71 | 33.87 | 7.53 | 31.64 | 35.23 |
| 100 | 0.050 | 26.52 | 26.77 | 14.49 | 10.93 | 30.54 | 30.64 | 23.95 | 22.31 | 34.65 | 34.98 | 35.93 | 35.53 | 7.80 | 31.96 | 35.51 |
| 100 | 0.100 | 26.35 | 26.31 | 18.39 | 16.11 | 30.24 | 30.56 | 27.10 | 25.26 | 34.49 | 34.74 | 35.94 | 35.79 | 7.79 | 31.78 | 35.25 |
| 100 | 0.500 | 26.26 | 26.38 | 25.52 | 24.99 | 30.11 | 30.31 | 29.90 | 29.34 | 34.45 | 34.69 | 35.60 | 35.82 | 7.66 | 31.61 | 35.42 |
| 100 | 1.000 | 26.14 | 26.33 | 26.11 | 25.61 | 30.27 | 30.32 | 30.09 | 29.91 | 34.51 | 34.64 | 35.29 | 35.38 | 8.55 | 31.44 | 34.87 |
| 100 | 1.500 | 26.31 | 26.45 | 26.37 | 26.15 | 30.20 | 30.52 | 30.49 | 30.19 | 34.54 | 34.84 | 35.20 | 35.46 | 7.22 | 31.54 | 35.43 |
| 200 | 0.000 | 23.97 | 21.94 | 2.08 | 1.52 | 29.20 | 27.80 | 7.65 | 6.13 | 34.36 | 35.35 | 21.26 | 18.74 | 5.63 | 30.15 | 34.33 |
| 200 | 0.001 | 24.45 | 23.27 | 2.53 | 1.88 | 29.57 | 28.80 | 9.01 | 7.24 | 34.22 | 34.96 | 23.39 | 21.19 | 5.88 | 29.80 | 33.66 |
| 200 | 0.002 | 24.74 | 23.92 | 2.84 | 2.24 | 29.68 | 29.15 | 9.99 | 8.34 | 34.29 | 35.04 | 25.92 | 22.69 | 6.00 | 30.36 | 34.11 |
| 200 | 0.005 | 24.77 | 24.35 | 3.82 | 3.09 | 29.91 | 29.81 | 12.54 | 10.73 | 34.53 | 35.37 | 31.38 | 27.73 | 5.09 | 30.47 | 34.50 |
| 200 | 0.010 | 24.55 | 24.49 | 4.90 | 4.15 | 29.62 | 30.11 | 15.62 | 13.50 | 34.64 | 35.15 | 33.58 | 32.22 | 4.92 | 30.52 | 34.73 |
| 200 | 0.050 | 24.50 | 24.76 | 9.93 | 8.32 | 29.64 | 29.93 | 22.20 | 20.76 | 34.67 | 35.03 | 34.84 | 34.75 | 5.28 | 30.32 | 35.01 |
| 200 | 0.100 | 24.50 | 24.47 | 13.37 | 10.47 | 29.88 | 29.99 | 23.67 | 22.40 | 34.84 | 35.00 | 35.18 | 34.93 | 5.39 | 30.82 | 34.91 |
| 200 | 0.500 | 24.45 | 24.59 | 23.27 | 22.06 | 29.61 | 29.89 | 28.99 | 28.15 | 34.61 | 34.97 | 35.74 | 35.94 | 5.22 | 31.00 | 35.04 |
| 200 | 1.000 | 24.34 | 24.50 | 23.65 | 23.36 | 29.71 | 29.80 | 29.55 | 28.95 | 34.74 | 34.84 | 35.63 | 35.78 | 5.02 | 30.34 | 34.98 |
| 200 | 1.500 | 24.55 | 24.80 | 24.64 | 24.26 | 29.65 | 29.99 | 30.11 | 29.75 | 34.75 | 34.97 | 35.57 | 35.67 | 4.97 | 31.04 | 35.01 |
| 400 | 0.000 | 21.84 | 15.45 | 1.57 | 1.15 | 28.59 | 23.68 | 6.20 | 4.90 | 34.06 | 34.25 | 19.48 | 15.90 | 4.44 | 28.96 | 33.39 |
| 400 | 0.001 | 21.43 | 18.14 | 1.87 | 1.42 | 28.54 | 25.68 | 7.05 | 5.55 | 34.90 | 35.30 | 21.21 | 18.66 | 4.01 | 29.25 | 33.99 |
| 400 | 0.002 | 20.87 | 19.30 | 2.23 | 1.75 | 28.35 | 26.42 | 7.97 | 6.54 | 34.76 | 35.55 | 23.34 | 20.90 | 3.46 | 29.20 | 34.37 |
| 400 | 0.005 | 20.52 | 19.75 | 2.89 | 2.35 | 28.09 | 27.72 | 9.96 | 8.51 | 34.20 | 35.01 | 27.81 | 23.52 | 2.94 | 28.51 | 33.50 |
| 400 | 0.010 | 21.23 | 21.31 | 4.01 | 3.36 | 28.53 | 28.48 | 13.43 | 11.46 | 34.48 | 35.05 | 32.51 | 29.95 | 3.69 | 29.47 | 34.33 |
| 400 | 0.050 | 20.63 | 20.54 | 7.23 | 6.21 | 27.93 | 28.39 | 19.91 | 18.30 | 34.91 | 35.21 | 34.56 | 34.29 | 2.66 | 28.79 | 34.13 |
| 400 | 0.100 | 21.09 | 21.23 | 10.32 | 8.81 | 28.40 | 28.58 | 22.54 | 21.43 | 35.25 | 35.42 | 35.61 | 35.36 | 2.98 | 28.89 | 34.85 |
| 400 | 0.500 | 20.27 | 20.58 | 16.76 | 14.62 | 27.84 | 28.31 | 25.38 | 24.69 | 34.76 | 35.04 | 35.94 | 35.72 | 2.65 | 29.02 | 34.33 |
| 400 | 1.000 | 20.40 | 20.63 | 19.65 | 18.31 | 28.31 | 28.56 | 27.53 | 26.60 | 34.98 | 35.31 | 36.18 | 36.20 | 2.95 | 29.43 | 34.59 |
| 400 | 1.500 | 20.02 | 20.42 | 19.82 | 19.05 | 27.93 | 28.17 | 27.78 | 26.96 | 34.64 | 34.87 | 35.72 | 35.76 | 2.55 | 28.69 | 34.20 |
| 800 | 0.000 | 18.64 | 12.21 | 1.26 | 0.92 | 27.24 | 22.67 | 4.98 | 3.87 | 34.97 | 35.55 | 17.25 | 13.91 | 2.86 | 26.25 | 33.90 |
| 800 | 0.001 | 17.06 | 11.99 | 1.48 | 1.11 | 26.18 | 22.64 | 5.45 | 4.31 | 33.79 | 34.61 | 18.44 | 15.41 | 2.54 | 26.08 | 32.80 |
| 800 | 0.002 | 16.96 | 13.45 | 1.81 | 1.38 | 26.49 | 25.63 | 6.80 | 5.36 | 34.91 | 36.15 | 21.16 | 18.87 | 1.98 | 26.52 | 33.86 |
| 800 | 0.005 | 17.97 | 16.11 | 2.44 | 1.84 | 26.51 | 25.36 | 8.99 | 7.28 | 34.35 | 35.29 | 25.29 | 22.32 | 1.95 | 27.34 | 33.76 |
| 800 | 0.010 | 17.43 | 16.96 | 3.40 | 2.81 | 26.60 | 26.60 | 11.62 | 10.03 | 35.11 | 36.44 | 32.06 | 27.94 | 2.00 | 28.03 | 35.32 |
| 800 | 0.050 | 5.38 | 5.02 | 0.70 | 0.63 | 6.63 | 7.31 | 1.32 | 1.23 | 8.19 | 12.92 | 8.35 | 7.65 | 2.36 | 6.68 | 8.23 |
| 800 | 0.100 | 5.08 | 4.89 | 0.75 | 0.67 | 6.33 | 6.93 | 1.39 | 1.26 | 7.58 | 10.99 | 7.80 | 7.37 | 2.38 | 6.38 | 7.52 |
| 800 | 0.500 | 4.49 | 4.75 | 0.80 | 0.75 | 5.66 | 6.41 | 1.52 | 1.39 | 7.00 | 8.43 | 7.52 | 7.41 | 2.14 | 5.57 | 6.71 |
| 800 | 1.000 | 4.41 | 4.71 | 0.87 | 0.80 | 5.56 | 6.21 | 1.70 | 1.51 | 6.87 | 8.00 | 7.50 | 7.38 | 2.12 | 5.38 | 6.49 |
| 800 | 1.500 | 4.39 | 4.61 | 0.87 | 0.85 | 5.47 | 6.10 | 1.88 | 1.65 | 6.81 | 7.73 | 7.43 | 7.41 | 2.11 | 5.31 | 6.36 |

$\tilde{N}$ and $\tilde{c}$ denote the simulation count and the exploration constant for training;
$N$ and $c$ denote the simulation count and the exploration constant for evaluation.
The values express the performance in average scores ($\times 10^4$); the colors reflect the ranking across all tested settings. The results of the expectimax search are provided as baselines.

For moderate training simulations, e.g., $20 < \widetilde{N} < 800$, each block (40 records) in Table 16 with the same training simulation $\widetilde{N}$ and the same evaluation simulate $N$ has a similar performance tending: high $\tilde{c}$ + high $c$ > high $\tilde{c}$ + low $c$ > low $\tilde{c}$ + low $c$ > low $\tilde{c}$ + high $c$. The observation that high $\tilde{c}$ + high $c$ > low $\tilde{c}$ + high $c$ is reasonable since if the network did not be trained with a high exploration constant $\tilde{c}$, evaluate it with a high exploration constant $c$ causing too many untrained states to affect the search result, thereby reducing the performance. The observation that high $\tilde{c}$ + low $c$ > low $\tilde{c}$ + low $c$ indicates that 2048 requires some form of exploration, the same as our conclusion in the previous chapter that applies optimistic methods.

Results from Table 16 are further extracted into Table 17 and Table 18 to clearly illustrate the difference between networks trained with low exploration and high exploration. For simplicity, denote $\{\tilde{c}\}_L = \{0.000, 0.001, 0.002, 0.005, 0.010\}$ to be low exploration; and $\{\tilde{c}\}_H \in \{0.050, 0.100, 0.500, 1.000, 1.500\}$ to be high exploration. Table 17 further averages the evaluation results for the low exploration networks. i.e., the value "23.00" for $\widetilde{N} = 20$, $N = 50$, and $c = 0$ is the average of 5 networks with the same $\widetilde{N}$, $N$, $c$, and $\tilde{c} \in \{\tilde{c}\}_L$ in Table 16. Table 18 is also the average of the evaluation results, but it is for high exploration networks.

Table 17. Performance of MCTS in the 4×6-tuple networks trained by TC learning that adopts MCTS with lower exploration as training target.

| Network | Evaluation Result | | | | | | | | | | | | | | |
|---|---|---|---|---|---|---|---|---|---|---|---|---|---|---|---|
| $\widetilde{N}$ | $N = 50$, $c =$ (below) | | | | $N = 140$, $c =$ (below) | | | | $N = 4400$, $c =$ (below) | | | | Expectimax | | |
| | 0.000 | 0.005 | 1.000 | 1.500 | 0.000 | 0.005 | 1.000 | 1.500 | 0.000 | 0.005 | 1.000 | 1.500 | 1-ply | 2-ply | 3-ply |
| 1 | 22.38 | 19.88 | 1.19 | 0.93 | 24.73 | 23.77 | 5.11 | 3.72 | 29.21 | 31.52 | 15.29 | 12.86 | 20.84 | 28.64 | 31.44 |
| 20 | 23.00 | 20.43 | 0.99 | 0.85 | 25.38 | 24.64 | 4.04 | 3.09 | 28.87 | 31.91 | 14.09 | 12.20 | 18.08 | 28.56 | 32.16 |
| 50 | 26.52 | 26.29 | 3.49 | 2.82 | 29.69 | 29.82 | 11.88 | 10.09 | 33.94 | 34.99 | 28.36 | 25.54 | 11.70 | 31.37 | 35.12 |
| 100 | 26.41 | 26.22 | 4.12 | 3.35 | 30.21 | 30.17 | 13.25 | 11.31 | 34.49 | 35.21 | 30.09 | 27.08 | 8.00 | 31.65 | 35.17 |
| 200 | 24.49 | 23.59 | 3.23 | 2.58 | 29.60 | 29.13 | 10.96 | 9.19 | 34.41 | 35.18 | 27.11 | 24.51 | 5.50 | 30.26 | 34.26 |
| 400 | 21.18 | 18.79 | 2.51 | 2.01 | 28.42 | 26.39 | 8.92 | 7.39 | 34.48 | 35.03 | 24.87 | 21.78 | 3.71 | 29.08 | 33.91 |
| 800 | 17.61 | 14.14 | 2.08 | 1.61 | 26.60 | 24.18 | 7.57 | 6.17 | 34.63 | 35.61 | 22.84 | 19.69 | 2.26 | 26.84 | 33.93 |

The values are the average of the corresponding $\widetilde{N}$ with $\tilde{c} \in \{\tilde{c}\}_L$ in Table 16; the colors reflect the ranking across all tested settings in this table.

In both Table 17 and Table 18, we observe that the evaluation performance of MCTS reaches the max at around $\widetilde{N} = 50$ or 100, then the performance starts to decrease as the training simulation $\widetilde{N}$ increases. Only the networks in Table 18 that trained high exploration $c_\tau$ and evaluated with high simulation $N$ are relatively not affected. This observation indicates that training with a high simulation count is not optimal for 2048 in general. However, the exact reason for this issue is still open.

Table 18. Performance of MCTS in the 4×6-tuple networks trained by TC learning that adopts MCTS with higher exploration as training target.

| Network $\tilde{N}$ | Evaluation Result | | | | | | | | | | | | | | |
|---|---|---|---|---|---|---|---|---|---|---|---|---|---|---|---|
| | $N = 50$, $c =$ (below) | | | | $N = 140$, $c =$ (below) | | | | $N = 4400$, $c =$ (below) | | | | Expectimax | | |
| | 0.000 | 0.005 | 1.000 | 1.500 | 0.000 | 0.005 | 1.000 | 1.500 | 0.000 | 0.005 | 1.000 | 1.500 | 1-ply | 2-ply | 3-ply |
| 20  | 13.40 | 13.53 | 2.89  | 2.06  | 15.14 | 15.40 | 7.15  | 6.39  | 17.69 | 18.24 | 18.38 | 17.60 | 7.37  | 15.00 | 17.70 |
| 50  | 26.21 | 26.46 | 21.26 | 19.61 | 29.56 | 29.76 | 27.40 | 26.30 | 33.91 | 34.09 | 35.26 | 35.25 | 11.62 | 31.21 | 34.93 |
| 100 | 26.32 | 26.45 | 22.18 | 20.76 | 30.27 | 30.47 | 28.31 | 27.40 | 34.53 | 34.78 | 35.59 | 35.60 | 7.80  | 31.66 | 35.30 |
| 200 | 24.47 | 24.63 | 18.97 | 17.69 | 29.70 | 29.92 | 26.90 | 26.01 | 34.72 | 34.96 | 35.39 | 35.41 | 5.18  | 30.70 | 34.99 |
| 400 | 20.48 | 20.68 | 14.76 | 13.40 | 28.08 | 28.40 | 24.63 | 23.60 | 34.91 | 35.17 | 35.60 | 35.47 | 2.76  | 28.97 | 34.42 |
| 800 | 4.75  | 4.80  | 0.80  | 0.74  | 5.93  | 6.59  | 1.56  | 1.41  | 7.29  | 9.61  | 7.72  | 7.44  | 2.22  | 5.86  | 7.06  |

The values are the average of the corresponding $\tilde{N}$ with $\tilde{c} \in \{\tilde{c}\}_H$ in Table 16; the colors reflect the ranking across all tested settings in this table.

Furthermore, the baseline performance of expectimax decreases as the training simulation $\tilde{N}$ increases. This observation indicates that as the training simulation increases, the afterstate values trained by MCTS become inaccurate for direct comparison, as in the 1-ply search. However, there has been no convincing explanation so far, as these networks still perform well when evaluated with more simulations.

On the other hand, when training simulation is low or high, e.g., $\tilde{N} = 20$ or $\tilde{N} = 800$, high exploration significantly inhibits the training, as in Table 18. For the low simulation $\tilde{N} = 20$, there are only a few nodes inside the search tree. Therefore, it makes sense that using high exploration seriously decreases performance. However, for the high simulation $\tilde{N} = 800$, using high exploration also corrupts the network. This may be caused by the stochastic characteristic of 2048; however, the exact reason for this issue is currently open.

### 4.3.5 Strength Adjustment

This section presents an approach to adjusting the program strength for the MCTS during the evaluation. The strength adjustment method discussed in this section has been published in the 33rd AAAI Conference on Artificial Intelligence (AAAI-19) [68], the IEEE Computational Intelligence Magazine [69], and two patents [70], [71]. For adjusting MCTS strength, a straightforward method to adjust the strength is changing the simulation count. However, the strength and the simulation count do not form a linear relationship. Even more, a smaller search tree could make the search result vulnerable to tactical traps. To address this issue, we propose to use a *softmax policy* with a *strength index z* to choose moves as

$$\pi(s, a) = \frac{N(s, a)^z}{\sum_b N(s, b)^z}, \tag{37}$$

where $a$ is an action at the root node; and $N(s, a)$ is the simulation count of action $a$. By this approach, if $z$ approaches infinity, the best action with the highest simulation count is selected, which is the same as the original MCTS; if $z$ is zero, the action is selected with equal probability; if $z$ approaches negative infinity, the worst action with the lowest quality is selected.

However, actions with a few simulation counts are visited by MCTS exploration; some of these low-quality actions should not be considered even when adjusting the strength. To generate the selected action exceeding a lower bound of quality, we filter these low-quality actions by excluding actions that have a lower simulation count than a pre-defined *threshold ratio* $R_{th}$ of the maximum simulation count as

$$a_{LQ} \in \{a \mid N(s, a) < R_{th} \times N(s, a_{best})\}, \tag{38}$$

where $a_{LQ}$ are actions to exclude; $N(s, a_{best})$ is the maximum simulation count of the best action of state $s$. The experiment results of the Go program ELF OpenGo [72] show that $z$ is highly correlated to the empirical strength when $-2 \leq z \leq 2$. To our knowledge, this result is state-of-the-art in terms of the range of strengths while maintaining a controllable relationship between the strength and the strength index.

To investigate the strength adjustment for the game of 2048, we used a well-trained 8×6-tuple network as the value function. According to previous experiments in Table 15 in Section 4.3.3, the exploration constant $c$ is set as 0.005, where MCTS performed 423140, 488028, 538961, and 553499 for $N = 1$, 140, 1000, and 4400, respectively. We evaluate the settings with $z \in \{0.0, 0.5, 1.0, 1.5, 2.0, 2.5, 5.0\}$ and $R_{th} \in \{0.25, 0.50, 0.75\}$. Each evaluation was tested for 1000 games. Table 19 summarizes the results. When using $R_{th} = 0.25$, the program covers a wide range of strength from about 320000 points to 550000 points. Note that the average score of the base network is already 423140 points with $N = 1$, i.e., a 1-ply search. This empirical experiment shows that the strength adjustment works not only for two-player deterministic games like Go, but also for single-player stochastic games like 2048.

Table 19. Performance of MCTS strength adjustment in the 8×6-tuple network trained by OTD+TC learning.

| N | z | $R_{th}$ | | | | | |
|---|---|---|---|---|---|---|---|
| | | 0.25 | 0.50 | 0.75 | 0.25 | 0.50 | 0.75 |
| 140 | 0.0 | 325432 | 430734 | 457867 | 17.00% | 36.00% | 37.30% |
| | 0.5 | 347651 | 442059 | 468457 | 20.40% | 36.50% | 38.90% |
| | 1.0 | 383458 | 442281 | 481774 | 26.50% | 35.30% | 41.90% |
| | 1.5 | 407406 | 443795 | 475201 | 29.90% | 36.00% | 40.60% |
| | 2.0 | 423477 | 444896 | 475208 | 33.80% | 37.00% | 42.00% |
| | 2.5 | 440157 | 467237 | 488522 | 35.60% | 40.10% | 43.70% |
| | 5.0 | 471969 | 466996 | 478539 | 41.10% | 38.90% | 41.00% |
| 1000 | 0.0 | 491162 | 525047 | 535922 | 46.10% | 51.90% | 51.70% |
| | 0.5 | 496499 | 522629 | 533247 | 46.60% | 49.70% | 50.90% |
| | 1.0 | 510089 | 532106 | 537490 | 48.30% | 51.30% | 52.90% |
| | 1.5 | 508603 | 522553 | 544108 | 46.20% | 49.90% | 54.50% |
| | 2.0 | 520074 | 520491 | 537651 | 50.40% | 49.30% | 54.20% |
| | 2.5 | 540526 | 538222 | 529196 | 53.10% | 53.50% | 51.20% |
| | 5.0 | 526608 | 524378 | 540858 | 49.20% | 48.70% | 54.00% |
| 4400 | 0.0 | 538237 | 547118 | 554767 | 53.50% | 54.30% | 55.80% |
| | 0.5 | 544015 | 544903 | 543239 | 54.20% | 53.90% | 54.60% |
| | 1.0 | 543803 | 548802 | 546717 | 54.60% | 54.70% | 53.60% |
| | 1.5 | 554422 | 553106 | 561091 | 56.10% | 55.40% | 57.60% |
| | 2.0 | 548600 | 540559 | 548705 | 53.80% | 52.00% | 53.70% |
| | 2.5 | 541920 | 549934 | 552893 | 53.60% | 55.00% | 55.90% |
| | 5.0 | 548652 | 556618 | 556267 | 55.30% | 56.50% | 56.10% |

$N$, $z$, and $R_{th}$ denote the simulation count, the strength index, and the quality threshold. The left side shows the average scores; the right side shows the 32768-tile reaching rates. The colors reflect the ranking across all tested settings. Note that the baseline performance is 488028, 538961, and 553499 for $N$ = 140, 1000, and 4400, respectively.

### 4.3.6 Discussion

The experiment result shows that for evaluation, 2048 requires the exploration constant $c$ to be a non-zero small number, and the optimal $c$ gradually increases when the available number of simulations increases. We assume that constant $c$ is highly correlated to the accuracy of the value network, and may be affected by other factors. For example, if an optimistically initialized network is not fully trained, the remained high feature weights will cause the search to focus on low-quality actions since their feature weights are high. However, this preliminary experiment shows that it is possible to replace the expectimax search with MCTS.

For the experiments that train networks with MCTS, we found that the networks may perform better with MCTS evaluation instead of expectimax evaluation. This show that MCTS may be used for training, while more analysis is required. On the other hand, we also observe that the performance generally decreases when the training simulation increases. This phenomenon is interesting. However, there is currently no way to explain its cause. These unexplained phenomena show that 2048 is still an exciting testbed for future research.

## 4.4 Deep Reinforcement Learning

*Deep reinforcement learning (DRL)* has achieved remarkable success for a wide range of applications. This section presents the attempts to apply DRL to the game of 2048. First, an early attempt follows the *deep Q-learning* proposed by Mnih *et al.* [25] is presented. The experiment results show that the training has some effect wherein it outperforms two baselines but has significant room for improvement. Nevertheless, this preliminary work surpasses all publicly available results related to DRL and 2048 in 2016, when it was published [34].

Second, another attempt based on *MuZero* proposed by Schrittwieser *et al.* [10] is presented, which combines two MuZero extensions, *Gumbel MuZero* [73] and *Stochastic MuZero* [11]. The experiments demonstrate that the combined algorithm achieved comparable performance to other programs, namely an average score of 394645. Furthermore, compared with other DRL-related works of 2048 [11], this work reduced the computational resources significantly, namely only three simulations during training.

### 4.4.1 Motivation

Deep reinforcement learning has been demonstrated to perform well in game-playing programs. Mnih *et al.* first presented a learning algorithm based on deep Q-learning with *convolutional neural networks (CNN)*, which was then applied to seven different Atari 2600 games, and outperformed humans in many games [25]. Later, many practical DRL algorithms, including the well-known AlphaGo [47], AlphaGo Zero [48], AlphaZero [49], and MuZero [10], were proposed. In their works, the input of the application consists of graphical information, namely, the pixels on the screen, which is similar to what a human player can see when playing these games. Unlike $n$-tuple networks use a predefined feature extractor, these DRL approaches learn the required features during by themselves. Therefore, we aim to investigate the effectiveness of DRL approaches for the game of 2048. The first approach is to replace the network from $n$-tuple networks to more complex deep convolutional neural networks (DCNN). The second approach is to apply the well-known DRL algorithms, Gumbel MuZero and Stochastic MuZero, to the game of 2048.

### 4.4.2 Early Attempt

Before this work, there have been several attempts to adapt neural networks to 2048 [74], [75]; however, they could not achieve the 2048-tile reliably. Therefore, we would like to explore the possibility of applying deep learning to 2048. More specifically, we used Q-learning and TD learning with deep convolutional neural networks that followed the architecture proposed by Mnih *et al.* [25]. In the rest of this section, we will present our network architecture, methods, and experiment results, with a brief discussion and remarks at the end.

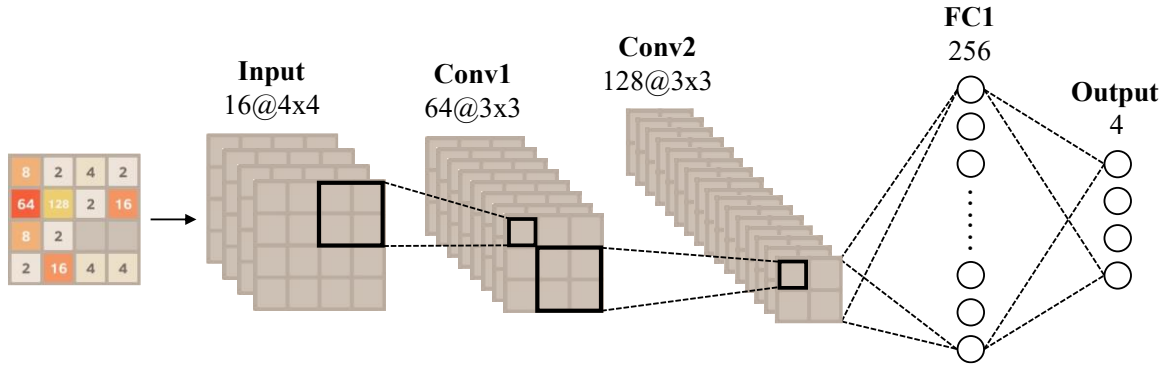

Figure 20. The Deep Q-Network architecture with 2×2 convolutional filters.

For network architectures, we follow the architecture proposed by Mnih *et al.* [25], with two variations depending on the filter size (2×2 and 3×3). The network with 2×2 filters is shown in Figure 20. It has two hidden layers for convolution, each followed by a *rectified linear unit (ReLU)*. After the convolution layers, there is a fully connected hidden layer and its ReLU. Finally, the fully connected output layer is designed with four possible move directions as the output. For the network that convolves with 3×3 filters, an additional convolutional hidden layer is added, i.e., for a total of three hidden convolutional layers.

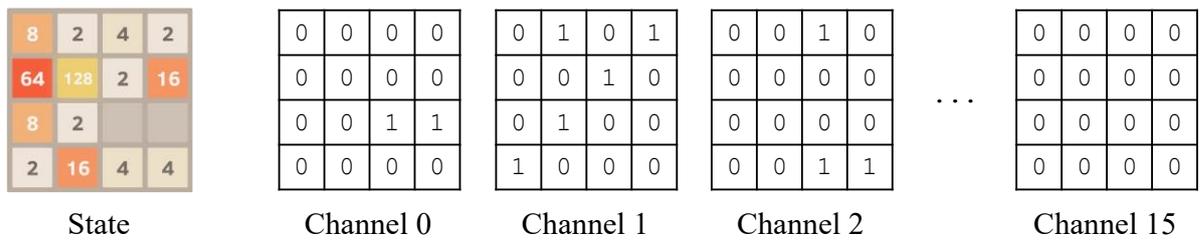

Figure 21. The input encoding for the DCNN model.
The left-most puzzle in the figure represents the original state.

The input state is encoded into 16 channels, where each channel is a 4×4 binary image; the $i$th channel marks each cell of the game position that contains the $i$th tile as 1, and 0 otherwise.

Note that $i$ = 0, 1, 2, 3, ..., 15 represent empty cells, 2-tiles, 4-tiles, 8-tiles, …, 32768-tiles, respectively. The state encoding is illustrated in Figure 21. For each hidden layer, we convolve either with 2×2 or 3×3 filters, as illustrated in Figure 22. The 2×2 case is convolved as is without padding. In the 3×3 case, the input image is zero-padded so that the convolution output, i.e., input for the next hidden layer, is a 4×4 image. In either case, the stride is 1.

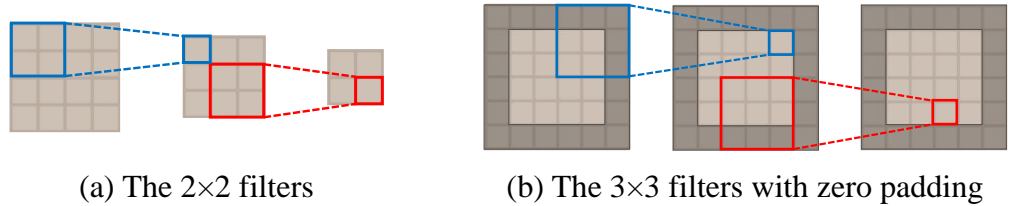

(a) The 2×2 filters     (b) The 3×3 filters with zero padding

Figure 22. The adopted convolutional filters for the 4×4 input encoding.

We used both Q-learning and TD-afterstate as the reinforcement learning component of our network updates. Although Q-learning does not perform as well as TD learning when using $n$-tuple networks [2], it is still worth trying Q-learning with deep neural networks. The algorithms for updating networks are the same as those introduced in Section 2.2. For the network gradient descent algorithm, we used the *AdaDelta* method [76] with a minibatch size of 32, implemented using the Caffe library [77]. For both Q-learning and TD-afterstate, we tested two network architectures with filters of different sizes. Note that the architecture is also changed according to the learning algorithm, e.g., only one output node for TD-afterstate.

Figure 23 and Figure 24 show the learning processes of the deep reinforcement learning methods in the deep convolutional neural networks with the 2×2 and the 3×3 filters, respectively. The real lines represent the average scores; the dotted lines represent the maximum scores. The networks with the 2×2 filter were trained with 100000 games; however, the networks with the 3×3 filter were trained with 10000 games as the 3×3 filter requires significantly more time than the 2×2 filter. We observed that the performance for TD(0) exceeds that for Q-learning in both kinds of networks, which corroborates the findings in [2], [28]. The 2×2 filter network performs better than the 3×3 network, but additional training for the latter needs to be conducted for a more conclusive result.

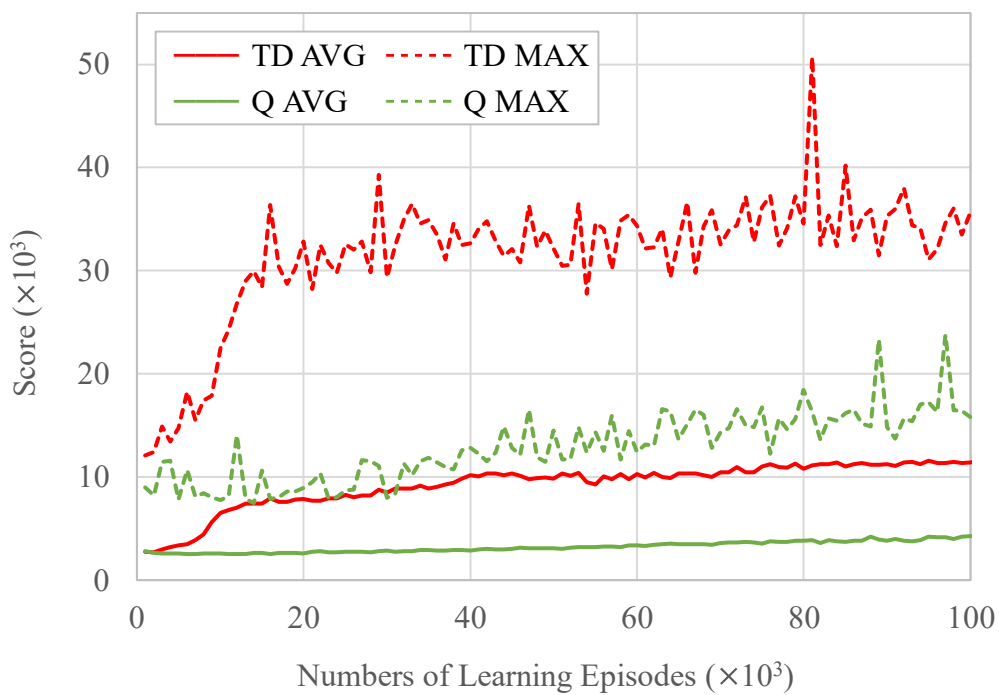

Figure 23. Average scores and maximum scores of Q-learning and TD learning in the DCNN with 2×2 convolutional filters.

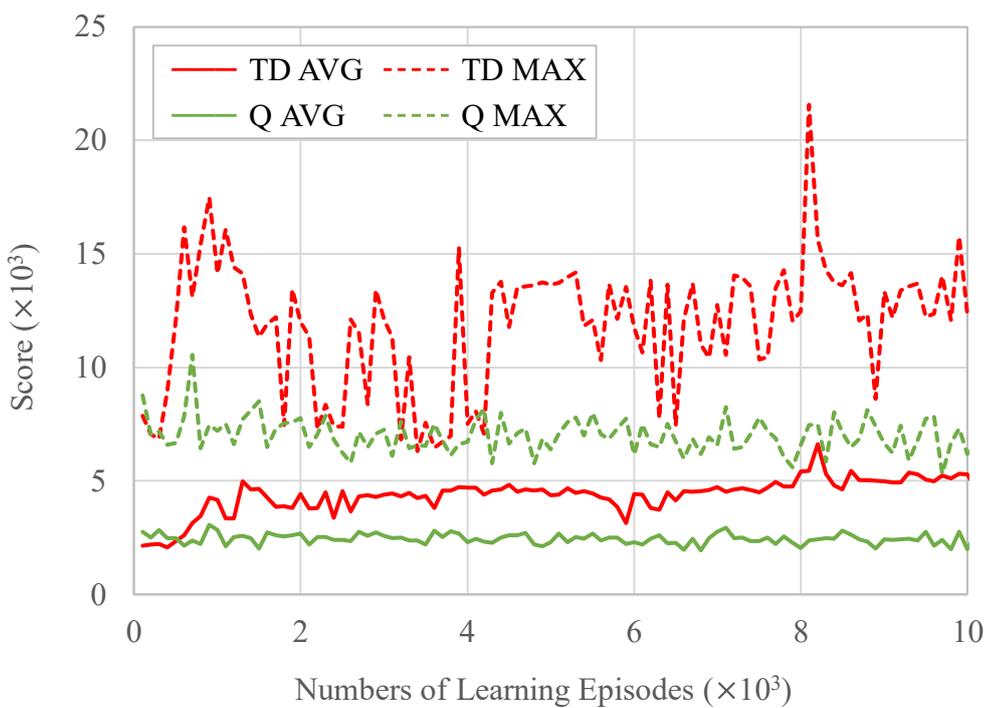

Figure 24. Average scores and maximum scores of Q-learning and TD learning in the DCNN with 3×3 convolutional filters.

Figure 25 compares the occurrences of maximum tiles between two baselines and two best-performing networks trained by Q-learning and TD learning. We used the random policy and the greedy policy as baselines. The random policy always selects a legal action randomly; the greedy policy always selects the action with the maximum immediate reward. Note that for 2048, the random policy cannot achieve 1024-tiles, and the greedy policy cannot achieve 2048-tiles. The experiment results show that the training has some effect, as we can confirm that at the very least the neural networks perform better than the two baselines. A single game was able to achieve a score above 50000, which corresponds to the single case in Figure 25 where the game ended with a maximum tile of 4096-tile. Again, we can see that TD(0) works better than Q-learning. However, both methods perform worse than the results of $n$-tuple networks as discussed in above chapters.

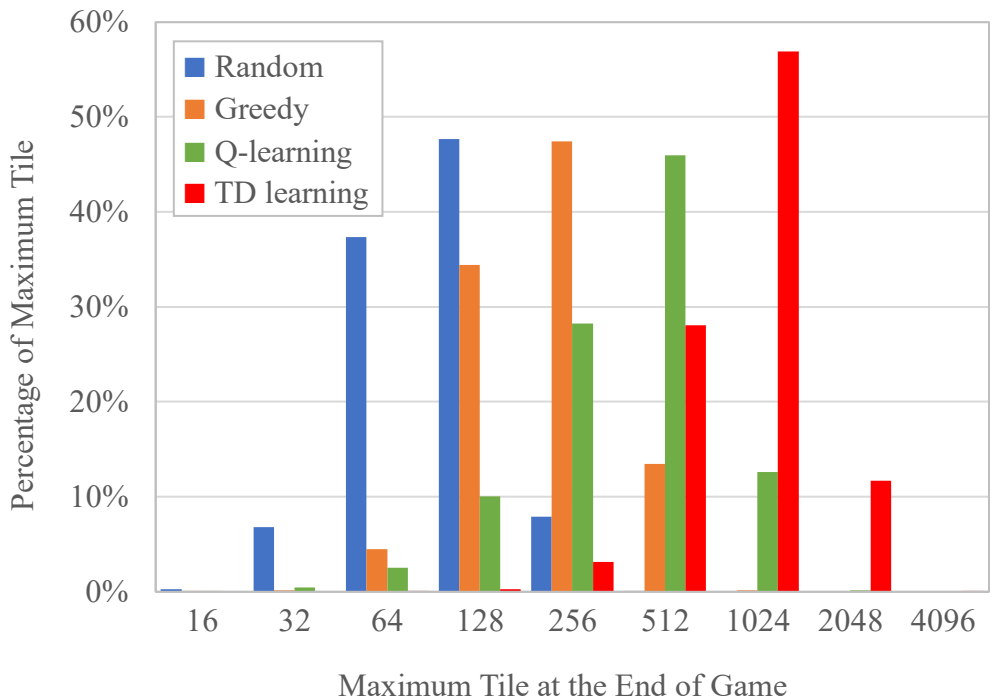

Figure 25. Maximum tiles of Q-learning and TD learning in the DCNN with 2×2 filters. Random policy and greedy policy are baselines.

### 4.4.3 MuZero

MuZero [10] is a well-known DRL algorithm that further generalizes AlphaZero [49], its ancestor, by learning the dynamics of the environment. This improvement allows the agent to plan without a perfect simulator. However, MuZero was initially designed for deterministic applications, such as Atari, Go, chess and shogi; and it requires hundreds of simulations for each move to generate high-quality training data. To address these limitations, DeepMind proposed Stochastic MuZero [11] and Gumbel MuZero [73]. Stochastic MuZero extends MuZero to stochastic environments by introducing the concepts of afterstates and chance events. Gumbel MuZero extends MuZero by guaranteeing *policy improvement* using *Gumbel top-k trick* [78] for selection, allowing the agent to learn with only a small number of simulations.

Stochastic MuZero has been successfully applied to 2048 and achieved an average score of approximately 510000 points [11]. However, for generating the training data, each action requires 100 simulations, which is a large number of computational resources. Gumbel MuZero appears to be a solution to provide reliable learning with a reduced number of simulations for collecting data; nevertheless, it has not been applied to stochastic environments [73]. Therefore, this section proposes to combine Gumbel MuZero and Stochastic MuZero, to take advantage of guaranteed policy improvement in stochastic environments, then empirically evaluate the combined algorithm on the game of 2048. The experiments demonstrate that the combined algorithm performs well, achieving an average score of 394645 points with only 3 simulations for each action during training. The empirical result was published at the 27th International Conference on Technologies and Applications of Artificial Intelligence (TAAI 2022) [79].

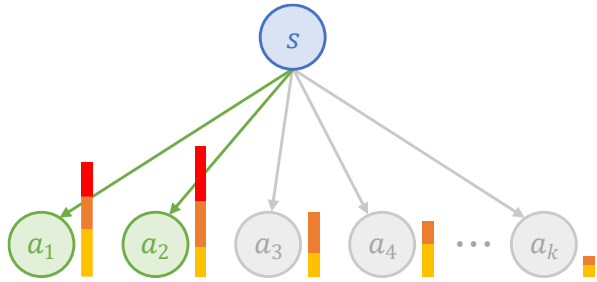

Figure 26. Selection mechanism in Gumbel MuZero.
In this example, the top $n = 2$ actions are sampled and evaluated.
Then, action $a_2$ is considered the best.

The procedure of Gumbel MuZero with $n$ simulations works as follows. First, at the root state, the top $n$ actions with the highest $g(a) + \text{logits}(a)$ are sampled, where $g(a)$ is a *Gumbel variable* sampled for action $a$, and $\text{logit}(a)$ is the logit of the action $a$ from a categorical distribution $\pi$. Then, the values $q(a)$ of the top $n$ actions are evaluated. Finally, the best action is considered to be the action with the highest $g(a) + \text{logits}(a) + \varsigma(q(a))$ selected from the top $n$ actions, where $\varsigma$ is a monotonically increasing transformation. Figure 26 shows an example that selects the top 2 actions in Gumbel MuZero.

If the number of available simulations $n$ is more than the desired number of sampled actions $m$, Gumbel MuZero uses *Sequential Halving* [80] to allocate the budget to the top $m$ actions, in which $n$ simulateions are divided into several phases. On the other hand, as a fixed budget allocation is applied, the search does not construct a search policy $\pi$. Therefore, Gumbel MuZero proposes the *completed Q-values* for training. This method computes the training target for policy by $\pi' = \text{softmax}(\text{logits} + \varsigma(\text{completedQ}))$, where $\text{completedQ}(a)$ equals to $q(a)$ when the action $a$ has been visited during the search; otherwise, it takes the value of the root node as an approximation.

The approach to combining Gumbel MuZero and Stochastic MuZero is as follows. For planning with few simulations $n = m$, the root node follows the mechanism proposed by Gumbel MuZero to select actions for evaluation. For planning with more simulations $n > m$, the root node allocates the budget with Sequential Halving as Gumbel MuZero does; however, the non-root nodes adopt the mechanisms in Stochastic MuZero to handle afterstates and chance events. On the other hand, the model consists of five functions, following that in Stochastic MuZero. For training the model, the policy $p$ is updated with the completed Q-values method in Gumbel MuZero, the value $v$ is updated with the best child value similar to that discussed in (35) in Section 4.3.4, and other functions are updated with the methods in Stochastic MuZero.

The effectiveness of the combined algorithm was evaluated using the game of 2048. The algorithm was implemented using a MuZero framework [81], [82]. The network model architecture was designed using *SE-ResNet* with 3 blocks × 256 filters; the input encoding followed that described in Section 4.4.2. For 2048, the $n$-step and the discount factor were both 1. For testing, 100 games with 50 simulations for each action were performed.

First, the effectiveness of the combined algorithm was compared by using Stochastic MuZero as a baseline. Figure 27 shows the learning performance with $n$ = 2, 3, and 4; and Figure 28 shows the learning performance with $n$ = 16 and 50. Note that for $n$ = 2, 3, 4, the combined algorithm used $m = n$ for training; and for larger $n$, $m = 3$. Since Gumbel MuZero guarantees a policy improvement, the combined algorithm learned well with only $n$ = 2. At the same condition, Stochastic MuZero apparently cannot learn with low simulations. For $n$ = 16, Stochastic MuZero started to learn, however, its performance was still not as good as the combined algorithm. Finally, for $n$ = 50, Stochastic MuZero matched the perfromace with the combined algorithm with sufficient simulations.

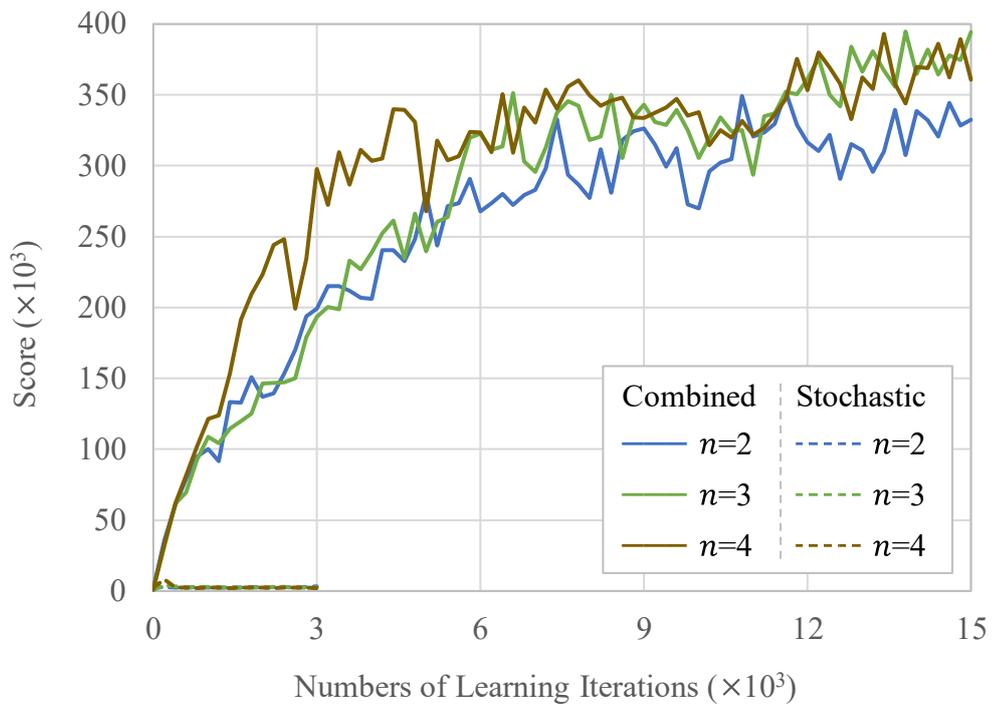

Figure 27. Average scores of MuZero-based algorithms with $n$ = 2, 3, and 4.

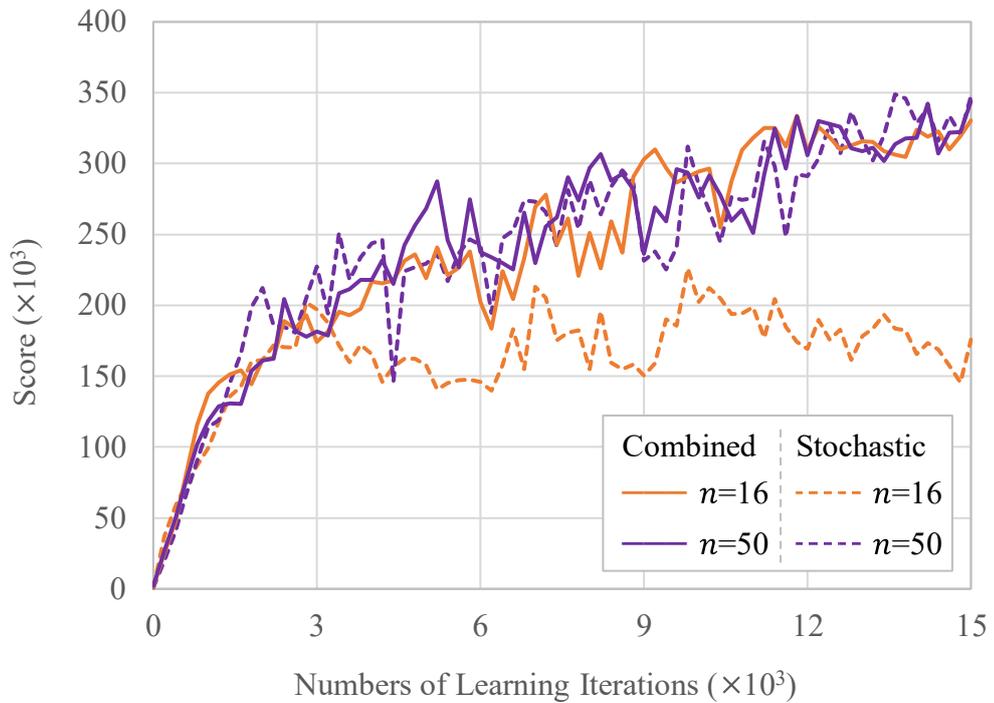

Figure 28. Average scores of MuZero-based algorithms with $n$ = 16 and 50.

Furthermore, the performance of using the different number of simulations for training was analyzed; more specifically, this experiment analyzed the learning performance using the different number of sampled actions $m \in \{2, 3, 4\}$ with the different number of simulations $n \in \{m, 16, 50\}$. Figure 29, Figure 30, and Figure 31 depict the average scores of $m = 2$, 3, and 4, respectively. Surprisingly, training with $n = 2$, 3, and 4 performed the best in each setting; while training with more simulations $n = 16$ or $50$ performed even worser than training with $n = m$. This is unexpectedly since usually a higher simulations results in a better training result. However, when the simple MCTS was applied in Section 4.3.4, a similar phenomenon has also been observed. The stochastic characteristic of 2048 may cause this phenomenon; however, more analysis is required to address it.

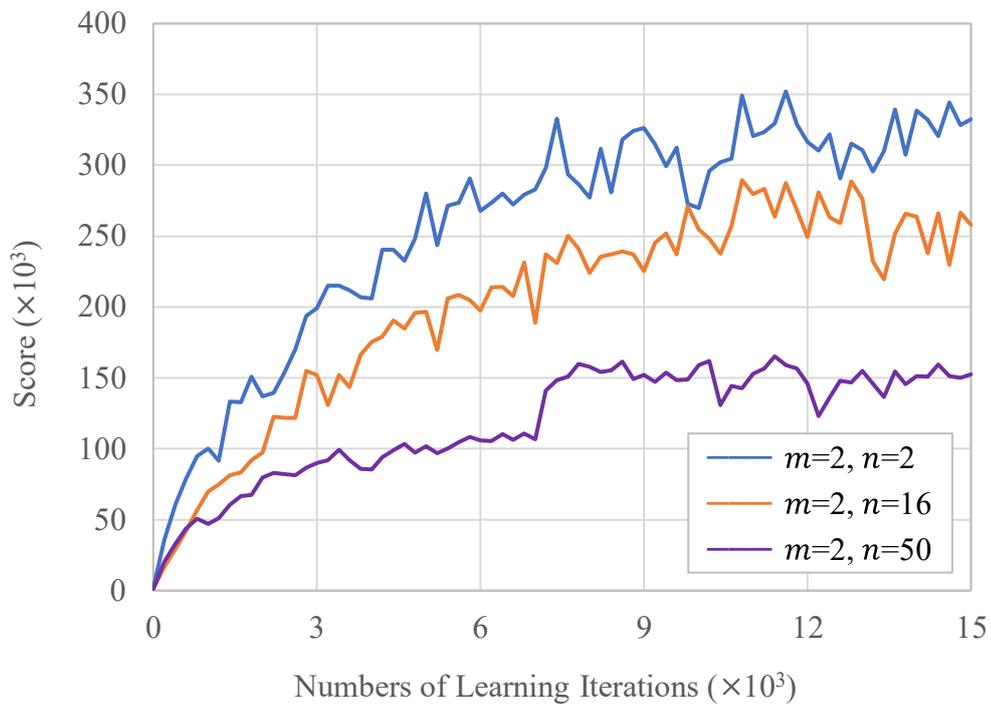

Figure 29. Average scores of the combined algorithm with $m = 2$ and different $n$.

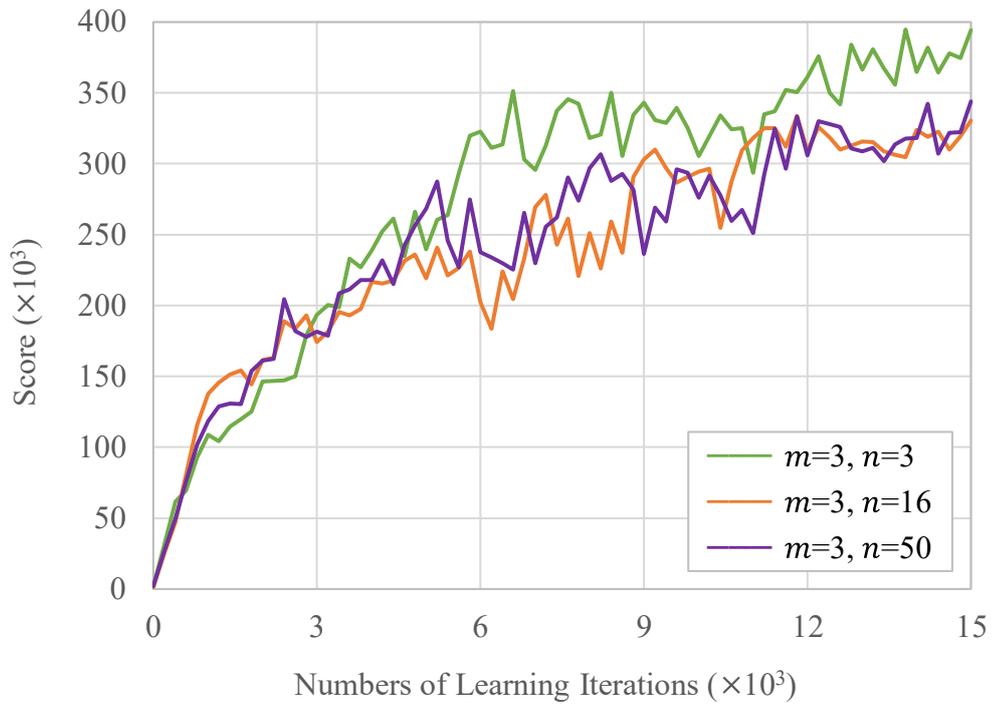

Figure 30. Average scores of the combined algorithm with $m = 3$ and different $n$.

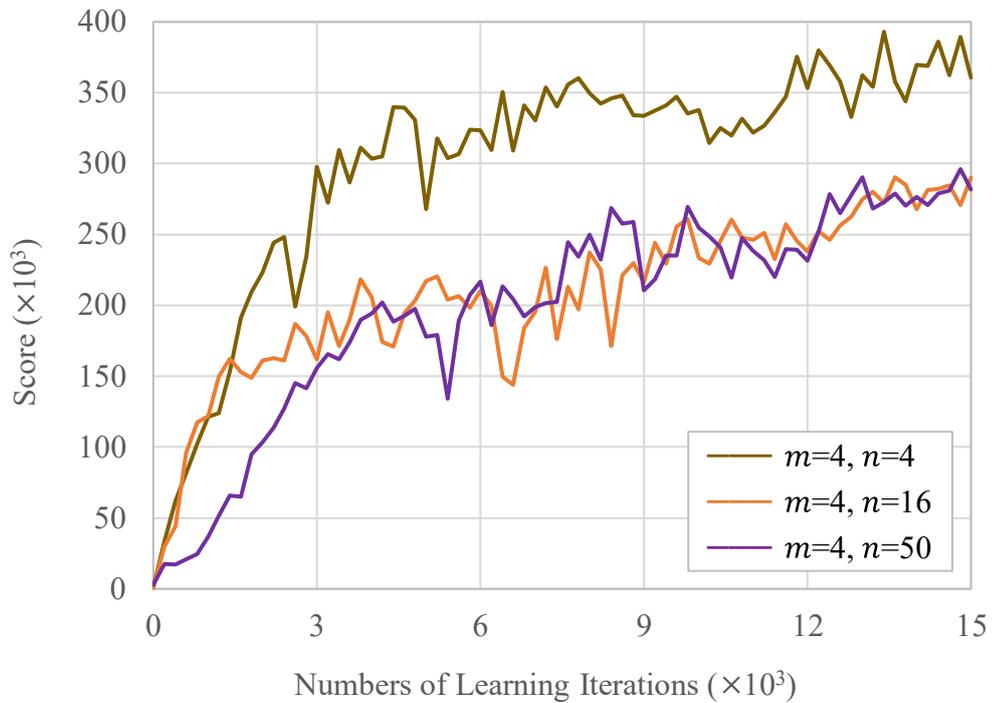

Figure 31. Average scores of the combined algorithm with $m = 4$ and different $n$.

With the combination of Gumbel MuZero and Stochastic MuZero, this work successfully applied Gumbel MuZero to 2048, a stochastic environment. The comparison of this work and other high-performance DRL-related algorithms is summarized in Table 20, which shows that the combined algorithm performs comparably with other DRL-related algorithms. Furthermore, the empirical experiments demonstrate that the combined algorithm performs well, achieving an average score of 394645 points with only $n = 3$ simulations for each action during training.

Table 20. Comparison of the DRL-based learning methods for 2048.

| Method | Training | | Testing | |
|---|---|---|---|---|
| | $n$ | # States | $n$ | Score |
| TD with DCNN [9] | 1-ply | $7 \times 10^8$ | 1-ply | 228100 |
| | | | 2-ply | 367024 |
| Stochastic MuZero [11] | 100 | $1 \times 10^{10}$ | $\approx 100$ | $\approx 250000$ |
| | | $1 \times 10^{11}$ | $\approx 100$ | $\approx 510000$ |
| Gumbel MuZero + Stochastic MuZero | 3 | $1 \times 10^{10}$ | 3 | 359721 |
| | | | 50 | 394645 |

Experiments with 1-ply and 2-ply adopted expectimax search instead of MCTS.

### 4.4.4 Discussion

This section presents two attempts at applying DRL to the game of 2048. The first attempt is an early attempt to adapt deep neural networks to the game of 2048, with methods based on TD learning and deep Q-learning. Although the result is much worse than the methods that relied solely on $n$-tuple networks, it was the best implementation among all publicly available cases that used DRL for 2048 at its publish time [34]. After this work, several studies on applying DRL to 2048 have been published, e.g., [35]–[38], [83]–[86]. In addition, the input encoding proposed in this work, as shown in Figure 21, has been used by many subsequent studies [9], [35]–[37]. Matsuzaki used this encoding with a much more complex DCNN and achieved an average score of 406927 with a 3-ply search, which is better than $n$-tuple networks when the number of network weights is the same [9]. These results imply the reason for the unsuccessful early attempt was that the network architecture was too simple and the number of training states was insufficient.

The second attempt is to combine Gumbel MuZero and Stochastic MuZero and evaluate the combined algorithm on the game of 2048. This work demonstrates that the combined algorithm achieved comparable performance to other DRL algorithms with few simulations during training. Furthermore, with the guarantee of policy improvement provided by Gumbel MuZero, the program can be trained using fewer computational costs, which can be a significant breakthrough for the research of DRL on the game of 2048. However, the performance of the MuZero algorithm did not achieve that of the optimistic learning algorithm with $n$-tuple network, which is surprising as MuZero outperforms many other learning algorithms in many other games. Based on the experiment results, we conjecture that it is mainly caused by (i) the difference in the number of training states; and (ii) the difference in the exploration techniques.

For the number of training states, since $n$-tuple network applies an efficient feature extraction method, it may efficiently learn more than $10^{12}$ states during training. However, for the deep neural networks used by the MuZero-based algorithm, we could only provide $10^{10}$ states for training due to computational costs; even a large company like DeepMind only provided $10^{11}$ actions using Stochastic MuZero [11]. It is important to emphasize that deep neural networks capture more information about each learned state. Consequently, with a limited number of learning states, the performance of the deep neural networks trained by MuZero-based algorithms may be much better than that of the $n$-tuple networks trained by the TD-based algorithms. For example, with only $10^{10}$ actions (approximately $10^6$ episodes) for learning, the combined algorithm achieved 359721 points, as in Table 20; but the $n$-tuple networks could not even achieved 240000 points, as in Figure 9 and Figure 11. Based on this observation, once the deep neural networks are trained with more states, they may perform better than the $n$-tuple networks. With the proposed combination of Gumbel MuZero and Stochastic MuZero, such a large training amount becomes feasible.

For the exploration techniques, MuZero applies standard techniques such as softmax and *Dirichlet noise*. In Chapter 3, we show that optimistic initialization (OI) works well for 2048. It is promising that applying OI with MuZero can improve learning performance. However, it is challenging to directly apply OI to deep neural networks in MuZero as it requires significant computing resources for training and a different method for initializing deep neural networks. Another potential direction is to adapt the MuZero algorithm to use $n$-tuple networks as the value function, which allows OI to be efficiently applied, as introduced in Chapter 3. Regardless, the combination of MuZero and OI is left open.

In the attempt to apply MuZero, we observed that increasing the number of simulations during the training decreases the learning performance. Since such a similar phenomenon was also observed when using vanilla MCTS for training, we conjecture that the stochastic environment of 2048 is the cause; however, more analysis is required to answer this issue. Finally, as there are so many active works on DRL for 2048, these works indicate that 2048 is a suitable testbed for stochastic games with DRL and confirm our research direction.

## 4.5 Chapter Conclusion

This chapter presents several attempts to enhance the performance of 2048.

Section 4.2 demonstrates that ensemble learning by stochastic weight averaging is promising for 2048. It works well with current optimistic methods and provides more possible training paradigms. As a 1-stage 8×6-tuple network already has comparable performance to the existing 2-stage SOTA 8×6-tuple network, with further fine-tuning, SWA may be the key to achieving the next SOTA for 2048.

Section 4.3 discusses the use of MCTS for 2048. However, many unexplained issues remain, e.g., increasing the number of simulations for training surprisingly decreases the performance. We assume that these interesting issues are caused by the stochasticity of 2048. However, 2048 may not be the only stochastic game that suffers from these issues. Addressing these unexplained issues may significantly improve reinforcement learning of stochastic games.

Section 4.4 presents the studies on applying deep reinforcement learning to 2048. We found that combining Gumbel MuZero and Stochastic MuZero significantly reduces the computational cost while maintaining comparable performance. In addition, as more and more researchers have joined this field, we believe that 2048 is still an exciting testbed where many open topics motivate us to improve our designs for better performance.

# Chapter 5    Pedagogical Applications

*2048-like games* are a family of single-player stochastic puzzle games, which consist of sliding numbered tiles that combine to form tiles with larger numbers. Notable examples of games in this family include *2048*, *2584*, and *Threes!*. 2048-like games are highly suitable for pedagogical purposes due to their simplicity and popularity. This chapter proposes guidelines for using 2048-like games to teach reinforcement learning and computer game algorithms, while also summarizing our experience using 2048-like games as pedagogical tools. Two types of course designs are proposed: *the lightweight course* and *the comprehensive course*. The lightweight course adopts 2048 as material, intending to familiarize students with temporal difference learning. The comprehensive course adopts 2584 and Threes! as materials, intending to guide students to develop their strong AI program using temporal difference learning and related computer game algorithms. The proposed course designs have been published in ICGA Journal [6] and two international conferences [24], [87]. More importantly, the courses have been successfully applied to graduate-level students and were well received by students.

## 5.1  Motivation

2048 is easy to learn and to play reasonably well, yet mastering the game is far from trivial. As teaching tools, 2048-like games are good candidates for beginners to learn both machine learning and computer games for their simplicity and popularity. Due to their simplicity, machine learning methods can be easily applied to these games without requiring complex implementations or expensive equipment. The present machine learning methods for 2048, e.g., [2], [3], [5], [8], [9], [23], [27], [28], [35]–[38], [46], [11], [56], [79], also provide a well-established basis for education. Due to their popularity, students have a high interest in and personal experience with the games, which may inspire student engagement. Because of these advantages, we have applied 2048-like games to graduate-level courses for teaching reinforcement learning and computer games for several years. To promote the pedagogical use of 2048-like games, we propose our course guidelines and summarize our experiences using 2048-like games as pedagogical tools.

This chapter is organized as follows. First, Section 5.2 introduces other 2048-like games and their common properties. Next, Sections 5.3 and 5.4 describe how teaching materials were designed using 2048-like games and summarize student results in our courses. Finally, Section 5.5 covers relevant discussion and makes concluding remarks.

## 5.2 Types of 2048-Like Games

The original 2048-like games are single-player stochastic, as introduced in Section 5.2.1. For the pedagogical applications, it is possible to extend the original single-player games to two-player paradigms further, as discussed in Section 5.2.2.

### 5.2.1 Single-Player 2048-Like Games

This subsection first introduces other single-player 2048-like games, including 2584 and Threes!, then summarizes the similarities and the differences between 2048, 2584, and Threes!.

2584[3] is a 2048-like game, differing from 2048 only in the tile values and the merging rules. As in Figure 32, the tiles are labeled by the numbers in the Fibonacci series. The player wins the game when reaching the 2584-tile in the puzzle. Like 2048, the environment generates 1st and 2nd tiles (1-tiles and 2-tiles) with probabilities of 0.9 and 0.1. However, the major difference is that the two tiles whose values are adjacent numbers in the series can be merged into the next larger number. For example, 1-tile + 1-tile = 2-tile; 1-tile + 2-tile = 3-tile; 2-tile + 3-tile = 5-tile; 3-tile + 5-tile = 8-tile.

Figure 32. Example of the puzzle of 2584.
In this example, all adjacent tile pairs are not mergeable.
This figure is obtained from an online site[3] for playing 2584.

Threes![4], developed by Vollmer and Wohlwend, precedes 2048 and 2584 and is often attributed as the first game in the 2048 family of games. The illustration of Threes! is shown in

---

[3] Available at https://davidagross.github.io/2048/
[4] Available at http://asherv.com/threes/ and http://threesjs.com/

Figure 33. The game is played on a 4×4 puzzle, starting with nine initial tiles that consist of 1-tile, 2-tile, or 3-tiles. The sliding distance in Threes! is at most 1, which is different from 2048. The merging rule for tiles also has many differences. First, a $2v$-tile is merged from two $v$-tiles for all $3 \leq v < 6144$; and a 3-tile is merged from a 1-tile and a 2-tile. Second, 6144-tiles are the maximum allowed in Threes! and can no longer be merged any further. Third, tiles can only be merged when the corresponding direction on the puzzle is filled. Fourth, if there is more than one mergeable tile pair in the same row, only the first pair can be merged. The game of Threes! ends when the player can no longer take legal action. The final score is the sum of $3^{\log_2(v/3)+1}$ of all $v$-tiles with $v \geq 3$. However, Threes! does not define a win as reaching the 2048-tile in 2048; its objective is simply to achieve the maximum possible score.

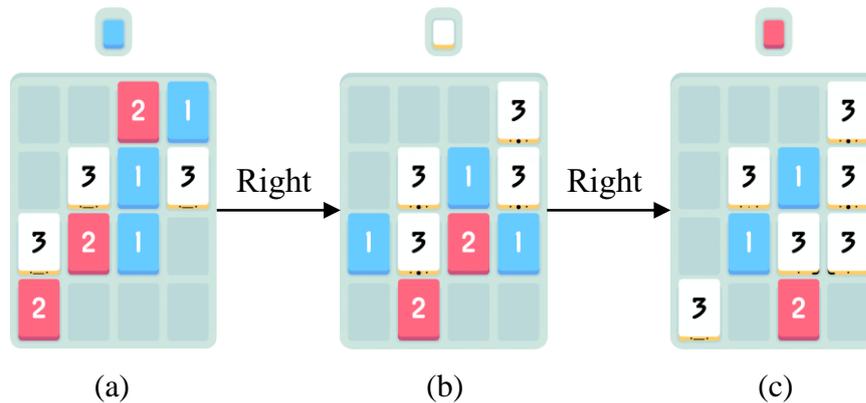

(a)          (b)          (c)

Figure 33. Example of the gameplay of Threes!.
The game starts with an initial state, as in (a). The player slides the puzzle right, and the environment generates a 1-tile, changing the puzzle from (a) to (b). Then, the player slides the puzzle right again, and the environment generates a 3-tile, changing the puzzle from (b) to (c). The figures of puzzles are taken from the official Threes! app[4].

The environment of Threes! also generates a new tile after every action. However, the rules for generating new tiles in Threes! are much more complicated than those for 2048. First, the player can peek at the type of next generated tile before taking action, the so-called *hint*. The hint appears at the top of the puzzle, as in Figure 33. There are four kinds of hints: 1-hint, 2-hint, 3-hint, and +-hint. The $v$-hint for $v \in \{1, 2, 3\}$ indicates that the next tile will be a $v$-tile; while the +-hint indicates a $v_+$-tile where $v_+ > 3$, which is so-called the *bonus tile*.

Second, the environment of Threes! generates *normal tiles* and *bonus tiles*. The normal tiles include 1-tiles, 2-tiles, and 3-tiles; the bonus tiles include $v_+$-tile for $v_+ > 3$. Consider that there is a bag of 12 tiles composed of equal amounts of 1-tiles, 2-tiles, and 3-tiles. After each action, the environment randomly draws a tile from the bag; then refills the bag with equal amounts of the normal tiles when it becomes empty. When the current largest tile, say the $v_{\text{max}}$-tile, is at least a 48-tile, there is a probability of 1/21 to generate a $v_+$-tile and a probability of 20/21 to generate a normal tile from the bag. The number $v_+$ of a bonus tile is uniformly selected from $6 \leq v_+ \leq 1/8 \times v_{\text{max}}$, e.g., $v_+ \in \{6, 12, 24, 48\}$ if $v_{\text{max}} = 384$.

Third, the environment randomly places the generated tile only at an empty cell such that (i) it appears on the opposite side of the last sliding direction; and (ii) it appears on a line that has just been changed. For example, from Figure 33 (b) to (c), only locations 8 and 12 (see Figure 3 (c) for cell locations) are possible for placing the new tile, which is determined as follows. (i) The last sliding direction, right, limits the possible locations to the left-most empty cells, i.e., 0, 4, 8, and 12. (ii) The first and second rows, (0, 0, 0, 3) and (0, 3, 1, 3), are not changed by sliding, which further limits the possible locations to cells 8 and 12.

The objective of 2048, 2584, and Threes! is to merge tiles and maximize the score. In 2048, merging occurs only for two tiles of the same value, where the final score is the sum of rewards, i.e., the sum of merged tile values. In 2584, merging occurs for tiles with adjacent values in the Fibonacci series, where the score is calculated the same way as in 2048. In Threes!, 1-tiles and 2-tiles can be merged into 3-tiles, and $v$-tiles for $v \geq 3$ can also be merged. However, the final score is more complicated, calculated by summing $3^{\log_2(v/3)+1}$ for all $v$-tiles where $v \geq 3$.

While the reward of actions is straightforward in 2048 and 2584 (the sum of all merged tiles from the action), the same cannot be said for Threes!, where the reward of actions is not apparent at first. More specifically, the score is calculated for the whole puzzle by tallying all the tiles when the game is over, so rewards for intermediate actions are not obtainable from the environment. To obtain rewards for intermediate actions, we can calculate the difference between the value of the puzzle before and after the action.

Generating a new tile is quite simple in 2048 and 2584, where the 1st and the 2nd tiles correspond to probabilities of 0.9 and 0.1, respectively. However, generating a new tile in Threes! is much more complicated, which depends on the player's actions and current tiles on the puzzle. When the largest tile of a Threes! puzzle is less than the 48-tile, the environment generates 1-tiles, 2-tiles, and 3-tiles. Otherwise, the environment may generate a bonus tile with a probability of 1/21, as described above.

The theoretical maximum tile generally depends on the board size, the merging rule, and the possible generated tiles. All three games are played on a 4×4 puzzle. In 2048, the maximum tile is theoretically 131072-tile, the 17th tile. However, the chances of seeing 65536-tiles and 131072-tiles are extremely low. In 2584, since adjacent-valued tiles are mergeable, tile indices can easily grow to more than 20, which is significantly larger than 2048 or Threes!. Therefore, the theoretical maximum tile is the 3524578-tile, the 32nd tile. However, as the case in 2048, large tiles like these are rare. In Threes!, although there is enough space for large value tiles, the maximum tile is limited by the game rules to be the 6144-tile, the 14th tile.

### 5.2.2 Two-Player 2048-Like Games

To accommodate for a competition at the end of the semester and to allow students to have hands-on experience with the minimax paradigm and adversarial techniques, we proposed changes to the original games so that they may be treated as two-player games as follows.

In our course design, we define the *player* as the role that slides the puzzle and plays the game; and define the *adversary* as the role of an antagonistic environment that tries to make the game more difficult for the player. In other words, the player maximizes the score while the adversary minimizes the score. Thus, the modified two-player game begins with the adversarial side. First, the adversary places some tiles on an empty puzzle. Then, the player and the adversary take turns sliding the puzzle or placing a new tile. Finally, the game ends when the player is unable to move.

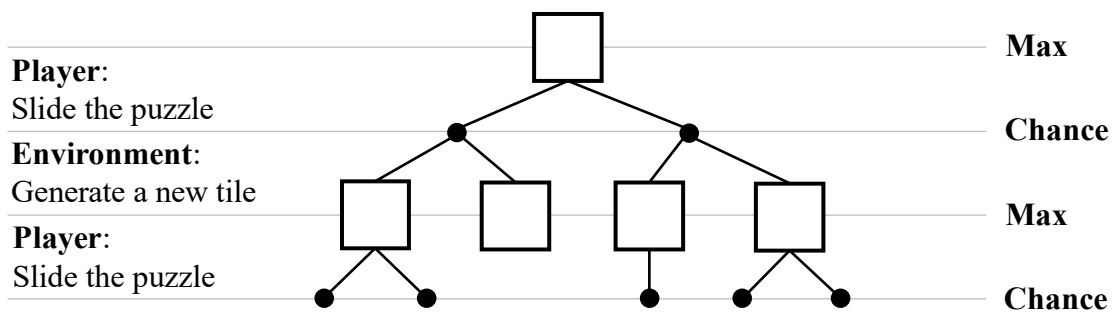

Figure 34. The expectimax paradigm of the original single-player 2048-like games.

Note that the original single-player 2048-like games conform to the *expectimax paradigm*, i.e., the player tries to maximize the score, and the type and the position of the tile added by the environment are selected randomly, as shown in Figure 34. However, in the two-player game where the adversary can determine both the type and the position of new tiles, a minimax search should be performed instead of an expectimax search. The minimax search tree comprises max and min nodes, corresponding to states and afterstates, as shown in Figure 35. The behavior of max nodes is similar to that in the expectimax paradigm, while min nodes minimize the player's score. In this case, the game conforms to the *minimax paradigm*. Conversely, if the environment still randomly selects the tile type and the adversary can only determine the tile position, the game conforms to the *expectiminimax paradigm*, as shown in Figure 36.

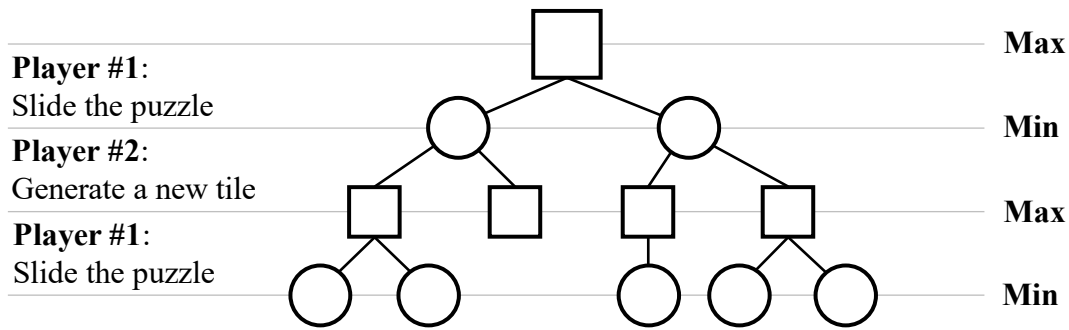

Figure 35. The minimax paradigm of the modified two-player 2048-like games. The second player determines both the tile type and the tile location.

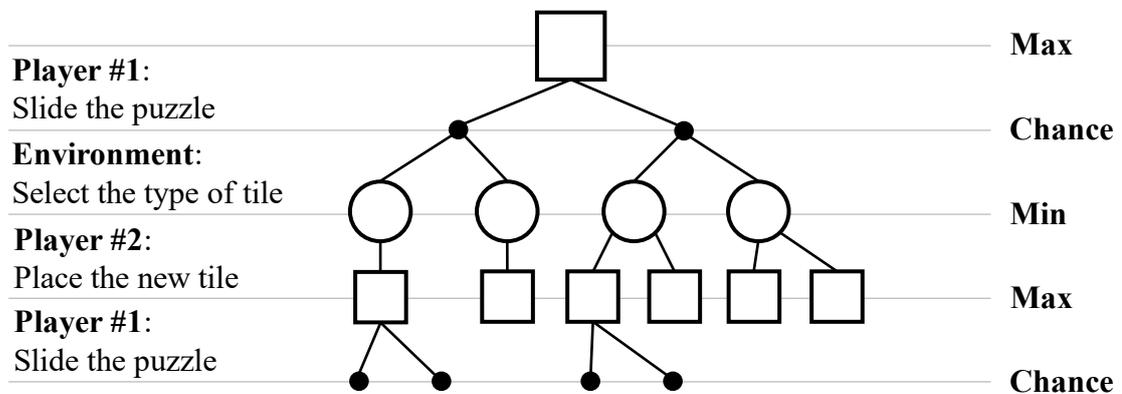

Figure 36. The expectiminimax paradigm of the modified two-player 2048-like games. The second player determines the location to place the generated tile, while the type of the generated tile is still determined randomly by the environment.

## 5.3 Design of Lightweight Course

The lightweight course familiarizes students with temporal difference (TD) learning, a basic reinforcement learning method. This subsection demonstrates the lightweight course design that adopts 2048 as pedagogical material for *Deep Learning and Practice*[5] in 2018.

---

[5] Deep Learning and Practice is a graduate-level course taught by Professor Wen-Hsiao Peng, I-Chen Wu, and Wen-Hsiao Peng at National Chiao Tung University in 2018.

In the lightweight course, students are not required to prepare the whole implementation. Instead, they are only required to implement the TD-afterstate method in a provided framework. The provided framework includes all other components required by a simple 2048 program, including the game engine and the $n$-tuple network. More specifically, the provided framework is modified from a complete framework[6] by removing lines related to TD learning.

With a correct TD-afterstate implementation, the performance of the 2048 program grows very fast. The first 2048-tile appears after several thousands of training episodes, which happened within minutes. Note that 2048 has the following characteristics: 1024-tile is unreachable if the agent randomly plays, and 2048-tile is unreachable if the agent always selects the direction with the max rewards. Therefore, by using the win rate of 2048-tile, it is easy to judge the correctness of the implementation. The student results are summarized in Table 21. The average win rate is only 85.9% because the pre-defined $n$-tuple network is relatively weak and the number of total training episodes is limited.

Table 21. Student results in the lightweight course.

| Game | Submission | Win Rate |
|---|---|---|
| 2048 | 73 submitted<br>10 were late | 85.9% ± 15.8%<br>8 reached 95%; 37 reached 90%; 25 did not reach 80% |

Besides the implementation, we also prepare several questions for students to answer. These questions help students clarify concepts about TD learning for 2048. Selected questions are listed as follows.

- Compare $V(s_t)$, $V(s'_t)$, and $Q(s_t, a_t)$.
- Describe how to calculate $V(s'_t)$ by using a trained $V(s)$ model.
- Describe how to calculate $V(s_t)$ by using a trained $V(s')$ model.
- Compare the following training methods.
  i) $\delta_t = r_{t+1} + V(s'_{t+1}) - V(s'_t)$
  ii) $\delta_t = r_{t+1} + r_{t+2} + V(s'_{t+2}) - V(s'_t)$

---

[6] Available at https://github.com/moporgic/TDL2048-Demo/

The proposed lightweight course does not require expensive hardware and can even be completed in a few hours. These essential properties make it relatively easy to be attached to comprehensive training programs about artificial intelligence or reinforcement learning. Furthermore, it is also possible to extend the proposed course. For example, after completing the TD-afterstate algorithm, implement the TD-state algorithm and compare the difference between TD-afterstate and TD-state.

## 5.4 Design of Comprehensive Courses

In the comprehensive course, students are guided to develop their game-playing program for a 2048-like game step by step, using a series of projects. In addition to reinforcement learning, the comprehensive course includes some related computer game algorithms, such as search methods. Students are required to improve their programs while learning new knowledge progressively. The requirements are not complicated; students can even try their ideas on their projects. Therefore, the courses not only provide beginners with an intuitive way to learn reinforcement learning and computer games effectively but also motivate them to challenge higher performances. The overall program is broken down into six projects as follows.

- Project 1: Get familiar with the game
- Project 2: Develop a player with high win rates
- Project 3: Solve a simplified game by search
- Project 4: Improve the performance of the player
- Project 5: Design an adversarial environment
- Project 6: Participate in the final tournament

In the subsequent subsections, we will exemplify how to use two 2048-like games, 2584 and Threes!, as pedagogical materials in the comprehensive courses designed for *Theory of Computer Games*[7] in 2017 and 2018, respectively.

---

[7] Theory of Computer Games is a graduate-level course taught by Professor I-Chen Wu at National Chiao Tung University in 2017 and 2018.

### 5.4.1 Project 1: Get Familiar With the Game

First, students must implement the framework and the environment for all the following projects in two weeks. We provide a framework[8] as the sample code in both C++ and Python. The framework is a demo program for 2048, which contains standardized IO functions, such as the format of the recording played episodes. We expect students to start their projects using the sample code, modifying it from the original 2048 rules to the target game.

The target environment needs to be simplified accordingly. We scale the difficulty of the target game such that the environment allows a win rate of about 85%–95% after simple TD training, which we discuss in the next subsection. Table 22 lists modifications for both 2584 and Threes!. In order to make sure that students are familiar with the rules and the framework, they also need to design a simple strategy to play the game. For example, a trivial strategy involves selecting the action that yields the maximum reward.

Table 22. Rule modifications in Project 1.

| Game | Rule Modifications |
| --- | --- |
| 2584 | Nothing is changed. |
| Threes! | Bag size is set to 3 (the original size is 12); no bonus tiles. New tiles can be randomly placed at any empty position on the opposite side of the last sliding direction. |

We do not expect students to use machine learning or sophisticated search techniques in this first project. Strict time requirements (1 CPU core; 1GB memory; 100000 moves/s for C++ and 1000 moves/s for Python) are enforced. Students receive 85% of full credit for this project if they implement the correct environment. The remaining 15% is dependent on their program performance. The detailed grading criteria are provided in Table 23, and the student results are summarized in Table 24. Most of the students simply chose to play according to the maximum reward. In some cases, the agents were designed to always select left and down if possible. Some of them performed a simple 2-ply search. Only a few students designed complex heuristic functions using patterns or custom rules.

---

[8] Project frameworks and specifications are available at following GitHub repositories.
  (C++) https://github.com/moporgic/2048-Framework/branches/
  (Python) https://github.com/moporgic/2048-Framework-Python/branches/

Table 23. Grading criteria in Project 1.

| Game | Environment | Average Score | Maximum Tile |
|---|---|---|---|
| 2584 | 85%, By correctness | 15% By $\min(\lfloor \text{Score}_{\text{AVG}} \div 1000 \rfloor, 15)$ | Bonus By $\max(\text{MaxTile} - 13, 0)$ |
| Threes! | 85% By correctness | 10% By $\min(\log_2(\text{Score}_{\text{AVG}} \div 3) + 1, 10)$ | 5% By $\max(\text{MaxTile} - 9, 0)$ |

Table 24. Student results in Project 1.

| Game | Submission | Average Score | Maximum Tile |
|---|---|---|---|
| 2584 | 39 submitted 4 were late | 10.64 ± 4.11 15 reached full credit (15 points) | 16.64th ± 1.29 11 reached the 18th tile |
| Threes! | 44 submitted 3 were late | 8.41 ± 0.73 3 reached full credit (10 points) | 9.36th ± 0.86 11 reached the 10th tile |

### 5.4.2 Project 2: Develop a Player With High Win Rates

Students need to train a stronger player using temporal difference learning (TD) with $n$-tuple networks in Project 2. The primary purpose is to ensure that students understand the mechanism of TD and $n$-tuple networks. They are not required to apply advanced TD techniques such as TD($\lambda$) or MS-TD in this project. This project should be completed within the one-month deadline. The environment of this project follows the one shown in Table 22.

We recommend the TD-afterstate learning framework for students as it is easier to implement in taking action and training. In most training cases, there are only minor differences between forward and backward implementations. Students can choose whichever they prefer, but we recommend backward training since it is easier to understand and debug.

A clear reward definition is necessary, which ensures that the agents are rewarded for surviving as long as possible. We suggest students follow the definition of the original games, but they are free to define their reward functions, such as giving a constant reward for each action taken. Note that Threes! does not define rewards in terms of actions but instead provides state value according to the game state. Practically, a simple reward can be calculated by taking the difference between the value function outputs of the state before and after an action.

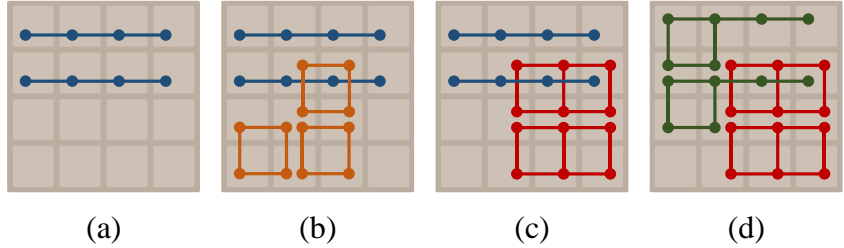

(a) (b) (c) (d)

Figure 37. Simple $n$-tuple network configurations.
(a) Lines; (b) Lines+Boxes; (c) Lines+Rects; (d) Axes+Rects.

Students can either survey papers for suitable $n$-tuple network configurations or design the patterns independently. However, we recommend that students start their project with the simplest configuration, "Lines," as shown in Figure 37 (a), and apply feature extraction with only rotations and without sharing lookup tables. This implementation comprises eight 4-tuple patterns (four rows and four columns) corresponding to eight standalone sub-lookup tables. Since this is relatively simple, it is a good starting point for students who are unfamiliar with $n$-tuple networks.

In summary, the baseline setting for students adopts (i) the standard TD(0) afterstate learning with backward training implementation, (ii) the $n$-tuple network "Lines" in Figure 37 (a) with only rotations, and (iii) a base learning rate $\alpha = 0.1$ without learning rate decay. When implemented correctly, the win rate grows rapidly in the early stages of training, and students should see their first win in the first 10k training episodes; then, students could reach a win rate of about 85% after 100k training episodes, which should only take at most several hours. Note that for this baseline setting to learn appropriately, we simplify the environment accordingly, as mentioned in Table 22.

Table 25. Grading criteria in Project 2.

| Game | Win Rate | Maximum Tile |
|---|---|---|
| 2584 | 100% <br> By [WinRate$_{384}$] | Bonus <br> By max(MaxTile $-$ 17, 0) |
| Threes! | 100% <br> By [WinRate$_{2584}$] | Bonus <br> By max(MaxTile $-$ 9, 0) |

WinRate$_v$ and MaxTile denote the win rate of $v$-tile and the index of the maximum tile in 1000 testing episodes, respectively. The credit is not counted without a correct environment.

Similar to Project 1, restrictions on memory usage and program speed are given (1 CPU core; 2GB memory; 50000 moves/s for C++ and 500 moves/s for Python). We use the win rate of 1000 games as the grade in Project 2, as shown in Table 25. For Threes!, since the 6144-tile is challenging to obtain, we use the win rate of 384-tile instead.

From the result listed in Table 26, most students achieved a win rate of 90% with TD for both 2584 and Threes!. However, the $n$-tuple network configurations were different, as in Table 27. In 2584, one-third of the students submitted their work with the most straightforward "Lines" configuration since reaching the 2584-tile is relatively easy, even for simple 4-tuple networks. However, in Threes!, the most straightforward configuration could only achieve a win rate of about 85%–90%. Therefore, many students applied the efficient 6-tuple "Axes+Rects" configuration for better performance.

Table 26. Student results in Project 2.

| Game | Submission | Win Rate | Maximum Tile |
| --- | --- | --- | --- |
| 2584 | 34 submitted<br>2 were late | 96.1% ± 4.2%<br>10 reached 100%; 3 did not reach 90% | 21.47th ± 1.24<br>15 reached the 22nd |
| Threes! | 43 submitted<br>8 were late | 93.5% ± 11.2%<br>5 reached 100%; 9 did not reach 90% | 12.91th ± 1.07<br>14 reached the 14th |

Table 27. Methods used by students in Project 1.

| Game | TD Method | Forward/Backward | N-Tuple Size |
| --- | --- | --- | --- |
| 2584 | TD(0): 32<br>TC(0): 2 | Forward: 3<br>Backward: 31 | 4.6 ± 0.8<br># of {4,5,6}-tuple: {20,7,7} |
| Threes! | TD(0): 41<br>TC(0): 1<br>MS-TD(0): 1 | Forward: 7<br>Backward: 36 | 5.6 ± 0.8<br># of {4,5,6}-tuple: {9,1,33} |

Another reason that 6-tuples were not widely used for 2584 is their high memory usage. As merging the tiles in both ways is possible in 2584, tile indices can easily grow to more than 20, which is significantly larger than Threes!. Therefore, students need to design compression techniques for their lookup tables if they want to use more extensive networks. As a workaround, some students trained their agents with custom 4-tuple or 5-tuple configurations, as shown in Figure 38. A student even tuned an efficient 5-tuple configuration, illustrated in Figure 38 (c), with comparable performance to the 6-tuple "Axes+Rects," winning 99% of its games in 2584.

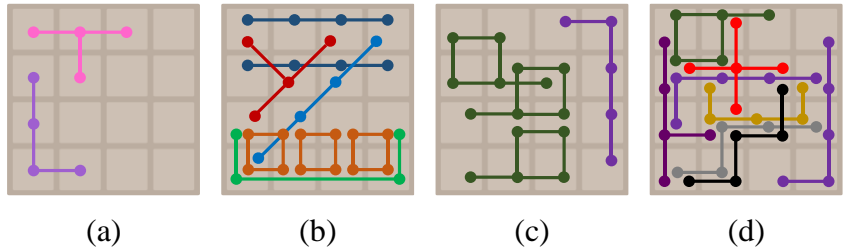

(a)   (b)   (c)   (d)

Figure 38. Custom $n$-tuple configurations designed by students.

### 5.4.3 Project 3: Solve a Simplified Game by Search

This project aims to familiarize students with the expectimax search, which is necessary for the following projects. Students should use the expectimax algorithm to solve a simplified game with a board size of $2 \times 3$. Even with the reduced board, brute-force expansion of the game tree is not feasible, so students are also required to implement a simple transposition table. The time allocated for this project can range between 2 weeks to 1 month, depending on the complexity of implementation between 2584 and Threes!.

The environment follows the previous one. However, since the board size is limited, we need to make minor changes for Threes!: the initial state only contains one initial tile here, while the previous environment contains nine tiles. In Project 3, we define three types of puzzles: *states*, *afterstates*, and *unreachable states*. The solver program is expected to be able to distinguish them. States and afterstates are described in detail above; unreachable states are puzzles that are unreachable from the initial states under legal gameplay. For example, a state filled with six 1-tiles is obviously unreachable for 2584 since it is impossible to accumulate that many 1-tiles as they will definitely merge with each other. The simplest way to distinguish unreachable states is to start expanding the game tree from the root. After the entire tree is expanded, all unvisited puzzles are unreachable.

The expected value can be calculated by performing max and weighted sum operations at max and chance nodes, respectively. Besides the expected value of the state, it is also possible to calculate *the best value* and *the worst value*. The best value is the total return when the environment coincidentally generates the best tile, which leads the player to the highest score under oracle play. The worst value is exactly the opposite. Some examples of Threes! can be found in Table 28.

In this project, we require student programs to solve the simplified game within a time limit of one minute, i.e., expand the entire game tree and store the results in a transposition table. Then, their solver programs are expected to answer problems. In typical cases, with proper implementation, calculating the value of the whole 2 × 3 game tree with a transposition table costs less than one second. Although some redundant recalculations may increase the total time to several seconds, one minute is generally safe for student programs. However, without the transposition table, the search tree of the simplified game cannot be expanded within the time limit. The expected value of states and afterstates are not necessarily the same; therefore, the transposition table should store their values independently. Note that the solver programs are not allowed to use any pre-calculated external database.

Table 28. Examples of state values for Threes!.

| State/Afterstate | | | | Value | | |
|---|---|---|---|---|---|---|
| Puzzle | Type | Hint | Last Action | Expected | Best | Worst |
| 0,1,0 / 0,0,2 | State | 3 | N/A | 455.97 | 1092 | 258 |
| 3,2,1 / 48,6,0 | Afterstate | 2 | Up | 198.94 | 837 | 75 |
| 1,12,3 / 1,2,6 | State | 2 | N/A | 326.66 | 1053 | 81 |
| 0,0,3 / 3,6,12 | Afterstate | 1 | Down | 310.07 | 1050 | 0 |
| 48,3,2 / 96,12,6 | State | 1 | N/A | 0.00 | 0 | 0 |
| 96,12,6 / 0,6,2 | Afterstate | 3 | Right | 3.00 | 3 | 3 |
| 0,1,0 / 0,0,3 | State | 1 | N/A | (unreachable) | | |
| 96,192,12 / 0,1,3 | Afterstate | 3 | N/A | (unreachable) | | |

Calculating afterstates requires the last actions to predict the location of the generated tile.

Project 3 is graded by the percentage of problems solved. Some problems and their corresponding answers are provided as sample inputs and outputs. The sample outputs we provided covered all possible boundary cases, so students had the opportunity to correct any problems accordingly before submission. Table 29 lists the student results for this project, where the result for Threes! is worse than those for 2584. There are two potential reasons for this. First, we observed that the students encountered difficulties implementing the expectimax search with hints, which was unique to Threes!. Since the framework we provided was designed for 2048, students had to program hint processing on their own. Second, we introduced some changes to Threes! in 2018 to increase the project difficulty so that we can better discriminate student performance. For example, although the definitions for the best and worst values are not difficult to understand, more specifications lead to more error avenues for students.

Table 29. Student results and the requirements in Project 3.

| Game | Submission | Requirement | Average Problems Solved [%] |
|---|---|---|---|
| 2584 | 34 submitted<br>3 were late | Expected value | 99.9% ± 0.5%<br>31 reached 100%; 3 did not reach 100% |
| Threes! | 35 submitted<br>11 were late | Expected value<br>Best value<br>Worst value | 92.5% ± 15.5%<br>24 reached 100%; 6 did not reach 90% |

## 5.4.4 Project 4: Improve the Performance of the Player

Project 4 tries to encourage students to improve their players with optional techniques. It is similar to Project 2 but with more stringent environmental conditions, which are listed in Table 30. Take 2584 as an example. The environment for Project 4 drops 1-tiles and 3-tiles instead of 1-tiles and 2-tiles. Since 3-tiles are not mergeable with both 1-tiles and other 3-tiles, it is more challenging to get a high score. A case in point is that the simple "Lines" 4-tuple design shown in Figure 37 can only reach a win rate of about 40% under these new conditions.

Table 30. Rule modifications in Project 4.

| Game | Rule Modifications |
|---|---|
| 2584 | *1-tiles and 3-tiles* are generated with probabilities of *0.75 and 0.25*, respectively. |
| Threes! | *Bag size and bonus tiles follow the original.*<br>New tiles can be randomly placed at any empty position on the opposite side of the last sliding direction. |

Differences between Project 4 and Project 2 are in italic.

To maintain good performance under more difficult environments, students must improve their programs using additional techniques that were covered in class. Also, they are encouraged to read recent papers for 2048 to find ideas to improve their program. Students are given one month to complete Project 4.

The four additional techniques we covered are listed as follows. The first suggestion is to adopt the expectimax search, which students used in Project 3. With an additional 1-ply search added, without retraining the network, the win rate of the "Lines" configuration can reach a 70% win rate. Since expectimax search is costly in computation time, bitboard and lookup tables are also optional improvements to consider. The second technique is to use more complex network structures. For example, replace 4-tuple networks with 6-tuple networks. To avoid high memory usage and to speed up the training, students should also implement isomorphism of features or include some form of compression. The third is to lower the learning rate, which will likely push the performance further. The last suggestion is to try an assortment of advanced TD methods, e.g., TC, MS-TD.

We use the same scoring criteria as Project 2, i.e., the final grades are calculated by win rate, where the winning tile is either the 384-tile or the 2584-tile for Threes! or 2584, respectively. However, we relax the computing resource restrictions (1 CPU core; 4GB memory; 10000 moves/s for C++ and 100 moves/s for Python) since the more sophisticated agent designs require more resources. The student results are summarized in Table 31 and Table 32.

Table 31. Student results in Project 4.

| Game | Submission | Win Rate | Maximum Tile |
|---|---|---|---|
| 2584 | 31 submitted<br>0 were late | 82.0% ± 11.3%<br>12 reached 90%; 12 did not reach 80% | 19.52th ± 0.71<br>16 reached the 20th |
| Threes! | 40 submitted<br>13 were late | 88.8% ± 15.1%<br>26 reached 90%; 4 did not reach 80% | 12.25th ± 0.66<br>13 reached the 13th |

Students were also encouraged to use advanced training techniques such as TC, TD($\lambda$), or MS-TD. However, only a few students adopted these techniques in Project 4. Most students tried to improve their player by using a larger network and incorporating expectimax search. Contrary to previous results in Project 2, students got better performances in Threes! in Project 4. One possible reason is that the rule changes for 2584 are much more complex, as 3-tiles are unmergeable with both 1-tiles and other 3-tiles; and are generated with a high probability of 0.25. Another possible reason for this result is the usage of larger networks. Many students encoded the hint into their network, which greatly improved performance. As the maximum tile for Threes! is limited to the 6144-tile (the 14th tile), students could easily enlarge the network.

Table 32. Methods used by students in Project 4.

| Game | TD Method | Forward/Backward | Depth | N-Tuple Size |
|---|---|---|---|---|
| 2584 | TD(0): 29<br>TC(0): 1<br>TC($\lambda$): 1 | Forward: 3<br>Backward: 28 | $2.6 \pm 0.8$<br>25 applied | $5.5 \pm 0.7$<br># of {4,5,6}-tuple: {4,8,19} |
| Threes! | TD(0): 35<br>TC(0): 4<br>MS-TD(0): 1<br>MS-TD($\lambda$): 1 | Forward: 7<br>Backward: 33 | $1.5 \pm 0.9$<br>10 applied | $6.3 \pm 0.7$<br># of {4,5,6,7}-tuple: {2,0,21,17} |

"Depth" summarizes the average search used by students; having no additional lookahead corresponds to a depth of 1; an additional 1-ply search corresponds to 3. Extra encoding of the network, such as hints, is counted as one extra tuple when counting the average $n$-tuple size.

Due to the computation time limit, adding an extra 2-ply expectimax search was nearly impossible with the original framework. Also, since Project 3 for Threes! was relatively difficult, fewer students added the expectimax search. Therefore, the average win rate in Threes! might have been higher if expectimax was applied more often.

### 5.4.5 Project 5: Design an Adversarial Environment

The objective of Project 5 is to build an adversary, which is a clear departure from previous projects where students work solely from the perspective of the player. The new rules for the adversary are listed in Table 33. We try different search paradigms in our course designs. The adversary in two-player Threes! controls both the type and position of a new tile, which conforms to the minimax paradigm. In contrast, the adversary in 2584 controls the position, while the type is randomly generated, which conforms to an expectiminimax paradigm.

Table 33. Rule modifications for the adversary in Project 5.

| Game | Modifications to Tile Generation Rules | | Paradigm |
|---|---|---|---|
| | Type of New Tile | Position of New Tile | |
| 2584 | Generated randomly | Determined by the adversary | Expectiminimax |
| Threes! | Determined by the adversary | Determined by the adversary | Minimax |

One month is given for Project 5. The best practice is to retrain the networks for the player and adversary under the new paradigm. However, it is still possible to reuse the weight table trained for Project 4 since the environments are the same. Previously, the player estimated afterstate values by looking up the corresponding weight tables and selecting the action based on the maximum value. For the adversary, one can oppositely adopt the same weight table, i.e., by choosing the minimum value. However, the weight table for Project 4 is designed for a player under the expectimax paradigm; the performance may drop slightly due to paradigm mismatch. Also, since the previous implementation usually takes an afterstate as input, it is not ideal for an adversary to access such a weight table since an additional 1-layer search is needed.

Students are encouraged to try a simple minimax search before adding *alpha-beta pruning*. However, in 2017 and 2018, we did not recommend Monte Carlo tree search (MCTS) to students since its performance tends to be weaker than minimax search together with TD value function from our experience[9]. Note that with the expectiminimax paradigm, the adversary cannot determine the type of dropped tiles. Therefore, programs need to handle chance nodes when performing the search. However, students may incorrectly bypass chance nodes by pre-generating the sequence of new tiles with the correct probabilities, which is inappropriate as the agent should never know the future tiles. The lecturer should emphasize the correct design of the expectiminimax search to avoid this situation.

Table 34. Grading criteria in Project 5.

| Game | Loss Rate of the Player | Maximum Tile |
|---|---|---|
| 2584 | 100% <br> By $[100\% - \text{WinRate}_{384}]$ | Bonus <br> By $\max(17 - \text{MaxTile}, 0)$ |
| Threes! | 100% <br> By $[100\% - \text{WinRate}_{2584}]$ | Bonus <br> By $\max(14 - \text{MaxTile}, 0)$ |

$\text{WinRate}_v$ and MaxTile denote the win rate of $v$-tile and the index of the maximum tile in 1000 testing episodes, respectively. The credit is not counted without a correct environment.

Players and adversaries communicate over a network connection. Due to the network communication cost, the time restriction is relaxed to 10 moves/s; the hardware resource restriction remains the same as in Project 4. We grade the adversary by its win rate, defined as the rate at which the player loses. Table 34 lists the grading details, and Table 35 summarizes five standard players with different configurations designed for grading. Five standard players share two trained networks; players No.1 and No.2 use one, and players No.3, No.4, and No.5 use another. Note that players No.2 and No.4 apply expectimax search intentionally since the mismatched paradigm causes them to play weaker, thereby reducing the difficulty.

---

[9] The vanilla MCTS that adopts random playout for simulations performs weaker than traditional search that adopts TD value function. However, with adopting TD value function as win rate estimator, MCTS performs not worser than or even better than traditional search (as mentioned in Section 4.3), which is a possible improvement to the course design.

Table 35. Standard players for grading in Project 5.

| N-Tuple Network | Training Setting | Player | Extra Search |
| --- | --- | --- | --- |
| Figure 37 (a) Lines With 4 isomorphisms, without shared LUTs | 1M episodes With a learning rate of 0.1, no learning rate decay | No.1 | N/A |
| | | No.2 | Expectimax 1-ply |
| Figure 37 (d) Axes+Rects With 8 isomorphisms, with shared LUTs | 1M episodes With a learning rate of 0.1, no learning rate decay | No.3 | N/A |
| | | No.4 | Expectimax 1-ply |
| | | No.5 | Expectiminimax 1-ply for 2584 Minimax 1-ply for Threes! |

The student results are summarized in Table 36 and Table 37. The variance of adversary performance in Threes! is quite large; half of the students who performed better had an average win rate of 84.3% ± 15.9%. This result was unexpected since the minimax setting for Threes! should have been easier to handle than the expectiminimax setting for 2584. The possible reasons for this unexpected observation were that hint processing is much more complex, mistakes during network reuse, or incorrectly implemented minimax search.

Table 36. Student results in Project 5.

| Game | Submission | Loss Rate of the Standard Players |
| --- | --- | --- |
| 2584 | 31 submitted 4 were late | 87.5% ± 14.7% 18 reached 90%; 4 did not reach 80% |
| Threes! | 37 submitted 1 was late | 52.7% ± 35.6% 9 reached 90%; 26 did not reach 80% |

The loss rate of the standard players is calculated by 100% − WinRate.

Students were told that they could retrain the weight table for an adversary that takes the states as the input, but only a few followed the suggestion. In 2017, most students completed the search, but only half followed up with alpha-beta pruning. Since the time restriction did not allow deep search even with alpha-beta pruning, the advantage of using alpha-beta search was insignificant. However, in 2018, not so many students adopted the search. Also, it was likely that many students implemented the incorrect minimax search for Threes! as the adversary performance is quite low.

Table 37. Methods used by students in Project 4.

| Game | TD Method | Forward/Backward | Depth | N-Tuple Size |
|---|---|---|---|---|
| 2584 | TD(0): 29<br>TC(0): 1<br>TC($\lambda$): 1 | Forward: 3<br>Backward: 28 | $2.9 \pm 0.5$<br>29 applied | $5.5 \pm 0.7$<br># of {4,5,6}-tuple: {4,7,20} |
| Threes! | TD(0): 31<br>TC(0): 4<br>MS-TD(0): 1<br>MS-TD($\lambda$): 1 | Forward: 7<br>Backward: 30 | $2.4 \pm 0.9$<br>25 applied | $6.4 \pm 0.7$<br># of {4,5,6,7}-tuple: {2,0,17,18} |

"Depth" summarizes the average search used by students, in which the additional 1-layer search of reusing the network is not included. Extra encoding of the network, such as hints, is counted as one extra tuple when counting the average $n$-tuple size

### 5.4.6 Project 6: Participate in the Final Tournament

In the final project, all students are required to participate in the tournament and compete with other classmates. The environment of our final tournament is the same as that of Projects 4 and 5. Students do not need to retrain or redesign their player and adversary, but some fine-tuning may help. In 2017 and 2018, we adopted round-robin scheduling since the number of students was manageable. However, a Swiss-system tournament would be better if the number of students increases for future semesters. Since the final tournament is the most sophisticated, we do not impose hardware limitations, and the time restriction is still 10 moves per second over the internet.

As only the player receives a score, two games are necessary to determine the result of a student pair. Each student acts as the player and the adversary exactly once, where the student whose player gets the higher score wins. We applied a simple strategy to produce the final ranking in 2017 and 2018: the winner of each student pair receives one point, and the total number of points determines the ranking. It is also possible to apply other rating systems, such as the Elo rating, but more games are required if the time limit is not an issue.

The whole tournament took about five hours to run. 30 students participated in the 2584 tournament in 2017; and 40 in the Threes! tournament in 2018. The top-ranked student got 28 points in 29 matches in 2017; the winner for 2018 got 109 points in 117 matches for Threes!. It is worth mentioning that the top-ranked programs adopted different strategies. The 1st place 2584 program adopted a simple 4-tuple network, as in Figure 38 (a), while the 2nd place program adopted a custom 5-tuple network, as in Figure 38 (c), with a relatively long training time. However, both the 1st and 2nd place Threes! programs applied complex 7-tuple networks (a 6-tuple in Figure 37 (d) with extra encodings). In addition, students performed inconsistently across projects, as in Figure 39. For example, the student who was awarded 1st place in the tournament did not perform very well in previous projects; on the contrary, some students who finished not very well in the tournament surprisingly performed well in early projects.

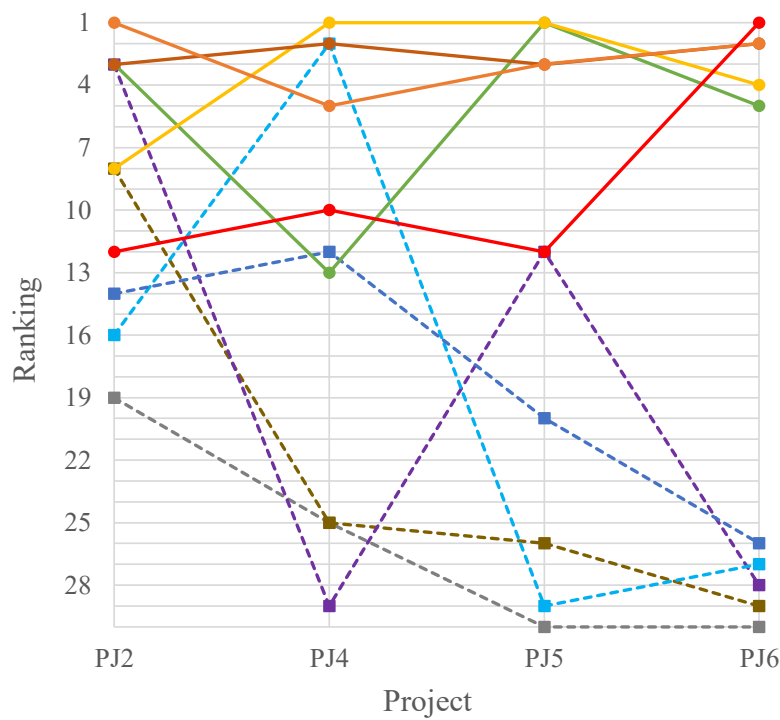

Figure 39. Student ranking trends across main projects in the course of 2584.
Projects 1 and 3 are not included since most students received full credits.
Only includes the ranking trends of the best 5 and the worst 5 students in the tournament.

In the 2584 tournament, only a few students did not use search techniques. Based on our recommendation, MCTS was not applied. Some students did not handle chance nodes under the expectiminimax paradigm correctly. Their programs simply generated a new tile with its corresponding probabilities when performing the search, thereby bypassing chance nodes altogether. This trick tends to result in a minor loss of strength. However, in the Threes! tournament, due to the complexity of search implementation, many students ended up with incorrect search algorithms that had a detrimental effect on performance. As a result, only a portion of students applied search methods.

Another difference between the above two tournaments was the usage of advanced TD methods in 2018. Only a few students adopted TC in 2017. However, many advanced TD methods such as $n$-step TD, TC, TC($\lambda$), or MS-TD appeared in 2018. The application of advanced methods increased the strength of the highest-performing programs and intensified the competition between students.

## 5.5 Chapter Conclusion

In this section, we will first discuss the course designs from several perspectives, including the grading of assignments, the common mistakes by students, the possible improvements, and the alternative course designs. We will finally make a summary of this chapter.

### 5.5.1 Grading Assignments

Students were expected to submit their source code and the trained network before the due date of each project; they were also required to answer questions about their implementation and present their program in person.

Table 38. Given time for completing each project.

| Game | Project 1 | Project 2 | Project 3 | Project 4 | Project 5 | Project 6 |
|---|---|---|---|---|---|---|
| 2584 | 1 week | 4 weeks | 2 weeks | 4 weeks | 4 weeks | 2 weeks |
| Threes! | 2 week | 4 weeks | 4 weeks | 4 weeks | 4 weeks | 2 weeks |

For the lightweight course, we allocated a week for students, although it can be completed in a few hours. For the comprehensive course, we allocated time according to the difficulty of each project, which is summarized in Table 38. We overlapped the ongoing projects to allow students to implement new techniques while running the TD learning algorithm, e.g., Project 3 was announced before Project 2 was due. Also, note that we attached the projects as part of the main courses (*Deep Learning and Practice* and *Theory of Computer Games*), so we reserved some time for students to complete other work unrelated to the projects.

For grading, as the program completely handles the environment, it is essential to ensure that the program does not "cheat" by simplifying the environment to obtain a higher score. One solution is to prepare the environment for students and prohibit them from modifying it. Specifically, students develop the project based on a modularized framework. Before grading, the source files related to the environment are replaced by the official files. We adopted this scheme from 2014 to 2016. However, since 2017, students were expected to implement the environment and modify the game rules as the semester progressed, so we changed to a different grading method using an online judging system.

We designed an online judging system to judge and grade student programs automatically. The student programs must output game records using a recording format designed for 2048-like games. This format represents each move with 2 characters and uses parentheses for storing additional information during gameplay. For example, the first 10 moves in an episode of 2048 can be

```
11D1(1)#L(2)61(3)#D[4](3)11(2)#D(3)B1(3)#L[4](3)61
```

For an environment move, the first character is the position in the hexadecimal number, and the second character is the value of the new tile, e.g., `11` represents place 1-tile at board location 1. For a player move, the first character is always `#`, and the second character is one of the sliding directions in `U`, `R`, `D`, `L`, e.g., `#L` represents slide to the left. The time usage and the reward are provided using parentheses after the move when available, e.g., `D1(1)` means `D1` takes 1 ms to complete; `#D[4]` means `#D` receives a reward of 4 points.

From the output, we can then judge the correctness of the environment. This online judging system also benefited students since they could debug programs or evaluate their performance with the online judge. We built the tournament server based on the online judging system and made the server online before the official tournament. As a result, students could play against one another freely over the internet once they connected their programs to the server. We believe this motivated students to try new techniques during play with classmates.

### 5.5.2 Student Feedback and Possible Improvements

From student feedback and our observations, there are several common issues students tend to run into. First, some students were not familiar with the Linux shell and programming. They were not experienced in debugging their programs when encountering fatal errors, e.g., segmentation faults. A crash course on basic Linux commands and debugging with GDB could be offered to address this. Second, students tended to share the same inquiries. Though we hosted an online forum for questions, there were many frequently asked questions by email or in person. Several workshops or an online FAQ may be necessary. Third, while we generally provided enough time for students, procrastination was still an issue. Since the projects were designed sequentially, delays in previous projects tend to propagate, which might cause students to drop the course altogether.

There are still several ways to improve the course design. First, only the basic TD learning algorithm is covered. The TD-afterstate algorithm is the only required algorithm in our current projects since it is the most effective method for 2048-like games. A possible further extension may require students to implement not only TD-afterstate but also TD-state and Q-learning. By comparing these training implementations, students will learn more about the mechanisms of these well-known reinforcement learning algorithms. In addition, only *model-free* training is covered now, while an extension to *model-based* training is also possible. We may be able to achieve this by designing a complex black box environment that changes, say, the distribution of generated tiles and the rewards based on time step or current state, then providing students with APIs to interact with the environment. In such a case, the grade can be determined by using Grade $\propto$ WinRate/EnvSampleCount, where the students are rewarded for better environment sampling efficiency. Another possible extension is to employ the $2 \times 3$ simplified puzzle to experiment with *policy iteration* and *value iteration*.

Second, we would like to address the lack of MCTS and deep learning in the proposed projects. MCTS is a critical algorithm used in contemporary programs and should be included. In the proposed projects, all the mentioned techniques were designed as part of a complete program, resulting in a progressive improvement of the program. Thus, MCTS was not covered in projects since an efficient MCTS approach for 2048 has not been analyzed in 2017 and 2018. However, as demonstrated in Section 4.3, the trained $n$-tuple networks work fine with MCTS, which provides a direction to incorporate MCTS into future lecture designs. On the other hand, while many modern programs use deep neural networks (DNN) as the function approximator, the current state-of-the-art learning algorithm for 2048 still adopts $n$-tuple networks. Following the reasoning behind the omission of MCTS, DNN was also not included in the projects. Recently, Matsuzaki demonstrated that deep convolutional neural networks (DCNN) could replace $n$-tuple network in the TD-afterstate learning [9], which also provides a direction for future lecture designs. However, the resources used for teaching need to be carefully planned, as replacing $n$-tuple networks with DCNNs requires more computing resources.

Third, the final tournament could be improved. Network connections between programs and the tournament server led to time constraints due to network latency. As a result, we could not allow multiple matches between each matchup. For the expectiminimax paradigm, it would be better to have multiple matches due to the randomness. It is possible to host the tournament locally by asking students to submit their programs beforehand. However, the tournament was designed in its current form because we found that students enjoyed the interactivity between their opponents; while also providing a good opportunity for students to exchange ideas.

### 5.5.3 Using Other Games as Materials

Throughout the more than 10 years of experience in applying games as teaching materials, we have considered using other games such as Go, Amazons, Hex, LOA, Chinese dark chess, NoGo, Gomoku, and Connect6. However, after considering and even using some of them, we concluded that 2048-like games are more suitable for teaching temporal difference learning and even arguably for reinforcement learning in general.

First, for the two-player games listed above, the opponent is treated as the RL environment, which introduces additional considerations such as *opponent modeling*. While it is still teachable, it adds an arguably unnecessary layer of complexity for the students. Since the objectives of the course include a significant portion of RL concepts, we want to prioritize teaching a simple, cohesive RL framework to the students before they think about concepts such as self-play in the two-player setting. In 2048-like games, the environment behaves separately from the agent's actions and can truly represent situations where the environment and agents are asymmetric, similar to many real-world problems. This property opens an opportunity to teach students about model-free learning much more intuitively and prepares them to apply their newly acquired knowledge to other fields.

Second, the reward signals are much simpler and more abundant for 2048-like games. For the above two-player games, the natural reward that does not require additional human-defined knowledge is simply the outcome (say, 1 for a win and 0 for a loss). The typical, naïve way of approaching this would be to teach Monte-Carlo learning [13]. While we agree that Monte-Carlo learning is important, we can teach the same concept with 2048, whereas if we use any of the examples above (Go, LOA, Amazons, Hex, etc.), we would also have to teach heuristic evaluations at the same time. In contrast, 2048-like games have rewards built into the game rules, giving students an intuitive feeling for the RL progress. Meanwhile, heuristics can be devised for 2048-like games, so it does not exclude the possibility of teaching hand-tuned evaluation functions.

Third, while the rules for Go are simple, it is a difficult game to play. Even in Asia, where Go is a well-known and popular game, few students can play and interpret game records at an adequate level. For the students who have not played Go before, it is unfair to expect them to understand their program's progress. While being a strong player personally is by no means necessary to create strong programs (that is, after all, the purpose of machine learning), it certainly helps. At the very least, familiarity with the game increases students' enjoyment, which we think is essential if part of our goal is to engage students and attract them to our research community.

In the end, it is all a matter of choice. While the same goals can be reached in many different ways, 2048-like games have the advantage of similarity to real-world scenarios, and simplicity in terms of teachable components, reward-rich environment, and familiarity to students.

### 5.5.4 Comparison with Alternatives

Various courses on reinforcement learning or computer game algorithms are available. However, most of them focus on only the former or the latter. Appendix D provides a more detailed introduction to some popular courses [88]–[93]. In short, these courses focus on their areas of expertise and allow students to learn more about either reinforcement learning or computer game algorithms. In addition, they may provide tutorials for various algorithms, but not as complete assignments as the projects listed in this work.

Again, this work aims to present a complete series of projects covering both reinforcement learning and computer game techniques, with a coherent goal of guiding students to develop their own strong AI programs of 2048-like games. Unlike other alternatives that use simple topics as assignments, we guide students to follow up on the recent research on 2048-like games. Students may therefore try new ideas on 2048-like games. Even more, their program may even be comparable to some of the state-of-the-art programs. For example, a student achieved 3072-tile at 78.3% in Project 4 of Threes!, where the environment was almost equivalent to the original. This result even surpassed the SOTA for Threes! in [3], where the 3072-tile reaching rate was 67.8%. The student results demonstrate that through our lectures, students were capable of not only grasping the key concepts but also improving current methods.

### 5.5.5 Summary

Due to the popularity and simplicity of 2048, 2048-like games have become our staple application since 2014. From positive student feedback (4.21 and 4.35 points on average) and progressive improvement in program strength over years of using 2048-like games as term projects, we consider 2048-like games are highly suitable for educational purposes, especially for beginners to learn the foundation of the temporal difference learning. Simultaneously, the techniques used in student programs may inspire new research. Furthermore, designing a game-playing program can also be an excellent way to recruit and motivate young minds to join our field and community.

# Chapter 6  Conclusion

This dissertation conducts comprehensive research on reinforcement learning and related computer game algorithms for the game 2048. This chapter concludes the proposed methods and discusses the research direction for future works.

First, optimistic temporal difference learning significantly enhances the program strength for 2048 by using optimistic initialization to improve the learning quality. With this approach and additional tunings, we achieved state-of-the-art performance, namely an average score of 625377 and a rate of 72% for reaching 32768-tiles. Table 39 compares the high-performance learning-based programs based on different algorithms for 2048. The future direction of this work is to integrate more techniques for an even better learning algorithm, e.g., TD($\lambda$), SWA, carousel shaping, and also tile-downgrading. The tile-downgrading technique has a lot of room for improvement; e.g., it could be incorporated into the training process to allow the agent learns larger tiles more fluently. We conclude that learning with explicit exploration for 2048 and similar applications is beneficial even if the environment appears stochastic enough. More specifically, for many applications, optimistic initialization is worth trying when standard exploration techniques do not work.

Table 39. Comparison of the high-performance learning-based programs for 2048.

| Author | Method | Search | Average Score | 32768 [%] |
|---|---|---|---|---|
| Matsuzaki [9] | TD-afterstate DNN | 3-ply | 406927 ± 27410 | 26.0% |
| Antonoglou *et al.* [11] | Stochastic MuZero DNN | ≈ 100 (MCTS) | ≈ 510000 | N/A |
| Jaśkowski [5] | MS-TC 16-stage 5×6-tuple | 1000ms | 609104 ± 38433 | 70% |
| This work | MS-OTD+TC 2-stage 8×6-tuple | 6-ply | **625377 ± 40936** | **72.00 ± 12.00%** |

Second, $n$-tuple network ensemble learning is promising. Leveraging stochastic weight averaging (SWA) improves performance significantly. SWA works well with current optimistic temporal difference learning methods and reveals more possible training paradigms worth investigating, e.g., the integration with cyclic learning rate. Furthermore, averaging snapshots from multiple individually trained networks is also worth trying since it may allow the ensemble network to generalize better. In this work, preliminary experiments on SWA improved the learning performance of the 1-stage 8×6-tuple network, even comparable to the current 2-stage

SOTA 8×6-tuple network. Therefore, with further fine-tuning, SWA may be the key to achieving the next SOTA for 2048.

Third, Monte Carlo tree search (MCTS) is promising for replacing the expectimax search for better evaluation performance. The experiments showed that MCTS with an exploration constant slightly larger than zero works well for 2048. However, many unexplained issues remain when attempting to train networks with MCTS, e.g., increasing the number of simulations for training surprisingly decreases the performance. We assume that these interesting issues are caused by the stochasticity of 2048. However, 2048 may not be the only stochastic game that suffers from these issues. Addressing these unexplained issues may benefit reinforcement learning in other stochastic environments.

Fourth, deep reinforcement learning (DRL) may become more critical for 2048 in the future. Previous DRL-related works for 2048 require high computational resources; however, combining Gumbel MuZero and Stochastic MuZero significantly reduces the required number of simulations during the training, thereby revealing further DRL approaches for the game of 2048. As more and more researchers have joined this field, we believe that 2048 is a suitable testbed for DRL and confirm that this research direction is correct.

Fifth, the pedagogical applications for teaching RL using 2048-like games have been successfully applied to graduate-level classes. The proposed courses guide students to develop a strong game-playing program by progressively improving their design while learning new knowledge. Therefore, the courses not only provide beginners with an intuitive way to learn RL and computer game-playing techniques effectively but also motivate them to challenge higher performances. In addition, the current courses do not cover several essential topics, such as MCTS or DRL. However, an improved course design is promising as these topics are all available now.

This dissertation conducts comprehensive research on reinforcement learning and related computer game algorithms for the game 2048. The research results include a state-of-the-art program, three journal publications, five conference publications, two patents, two courses, three open-source programs, and many promising techniques with preliminary experimental results. At the end of this dissertation, we conclude that 2048 is an interesting and challenging platform, not only for experts investigating advanced topics and methods, but also for beginners learning the foundations of reinforcement learning.

# References

[1] G. Cirulli. "2048, success and me." May 9, 2014. Accessed: Jun. 16, 2020. [Online]. Available: https://web.archive.org/web/20140517152011/http://gabrielecirulli.com/articles/2048-success-and-me

[2] M. Szubert and W. Jaśkowski, "Temporal difference learning of N-tuple networks for the game 2048," in *Proc. 2014 IEEE Conf. Comput. Intell. Games (CIG 2014),* Dortmund, Germany, 2014, pp. 1–8, doi: 10.1109/CIG.2014.6932907. [Online]. Available: http://www.cs.put.poznan.pl/mszubert/pub/szubert2014cig.pdf

[3] K.-H. Yeh, I-C. Wu, C.-H. Hsueh, C.-C. Chang, C.-C. Liang, and H. Chiang, "Multistage temporal difference learning for 2048-like games," *IEEE Trans. Comput. Intell. AI Games*, vol. 9, no. 4, pp. 369–380, Dec. 2017, doi: 10.1109/TCIAIG.2016.2593710. [Online]. Available: *arXiv:1606.07374*.

[4] R. Mehta, "2048 is (PSPACE) hard, but sometimes easy," Electron. Colloq. Comput. Complex., Tech. Rep. TR14-116, Sep. 6, 2014. [Online]. Available: https://eccc.weizmann.ac.il/report/2014/116/download

[5] W. Jaśkowski, "Mastering 2048 with delayed temporal coherence learning, multistage weight promotion, redundant encoding and carousel shaping," *IEEE Trans. Games*, vol. 10, no. 1, pp. 3–14, Mar. 2018, doi: 10.1109/TCIAIG.2017.2651887. [Online]. Available: *arXiv:1604.05085*.

[6] H. Guei, T.-H. Wei, and I-C. Wu, "2048-like games for teaching reinforcement learning," *ICGA J.*, vol. 42, no. 1, pp. 14–37, May 28, 2020, doi: 10.3233/ICG-200144.

[7] W. Konen, "General Board Game Playing for Education and Research in Generic AI Game Learning," in *Proc. 2019 IEEE Conf. Games (CoG 2019),* London, UK, 2019, pp. 1–8, doi: 10.1109/CIG.2019.8848070.

[8] K. Matsuzaki, "Systematic selection of N-tuple networks with consideration of interinfluence for game 2048," in *Proc. 21st Int. Conf. Technol. Appl. Artif. Intell. (TAAI 2016),* Hsinchu, Taiwan, 2016, pp. 186–193, doi: 10.1109/TAAI.2016.7880154.

[9] K. Matsuzaki, "Developing Value Networks for Game 2048 with Reinforcement Learning," J. Inf. Process., vol. 29, pp. 336–346, Apr. 2021, doi: 10.2197/ipsjjip.29.336. [Online]. Available: https://www.jstage.jst.go.jp/article/ipsjjip/29/0/29_336/_pdf/-char/en

[10] J. Schrittwieser, I. Antonoglou, T. Hubert, K. Simonyan, L. Sifre, S. Schmitt, A. Guez, E. Lockhart, D. Hassabis, T. Graepel, T. Lillicrap, and D. Silver, "Mastering atari, go, chess and shogi by planning with a learned model," *Nature*, vol. 588, no. 7839, pp. 604–609, Dec. 2020, doi: 10.1038/s41586-020-03051-4.

[11] I. Antonoglou, J. Schrittwieser, S. Ozair, T. K. Hubert, and D. Silver, "Planning in Stochastic Environments with a Learned Model," in *the 10th Int. Conf. Learn. Representations (ICLR 2022)*, Virtual, Apr. 25–29, 2022. [Online]. Available: https://openreview.net/pdf?id=X6D9bAHhBQ1

[12]  T. W. Neller, "Pedagogical possibilities for the 2048 puzzle game," *J. Comput. Sci. Colleges*, vol. 30, no. 3, pp. 38–46, Jan. 1, 2015, doi: 10.5555/2675327.2675335. [Online]. Available: https://cupola.gettysburg.edu/cgi/viewcontent.cgi?article=1025&context=csfac

[13]  R. S. Sutton and A. G. Barto, *Reinforcement Learning: An Introduction*, 1st ed., Cambridge, MA, USA: MIT Press, 1998. [Online]. Available: http://incompleteideas.net/book/first/ebook/the-book.html

[14]  G. Tesauro, "TD-Gammon, a self-teaching backgammon program, achieves master-level play," *Neural Comput.*, vol. 6, no. 2, pp. 215–219, Mar. 1994, doi: 10.1162/neco.1994.6.2.215.

[15]  J. Schaeffer, M. Hlynka, and V. Jussila, "Temporal difference learning applied to a high-performance game-playing program," in *Proc. 17th Int. Joint Conf. Artif. Intell. (IJCAI-01)*, Seattle, WA, USA, 2001, pp. 529–534.

[16]  J. Baxter, A. Tridgell, and L. Weaver, "Learning to play chess using temporal differences," *Mach. Learn.*, vol. 40, no. 3, pp. 243–263, Sep. 2000, doi: 10.1023/A:1007634325138.

[17]  D. F. Beal and M. C. Smith, "First results from using temporal difference learning in Shogi," in *Proc. 1st Int. Conf. Comput. Games (CG 1998)*, Tsukuba, Japan, 1998, pp. 113–125, doi: 10.1007/3-540-48957-6_7.

[18]  D. Silver, "Reinforcement learning and simulation-based search in computer Go," Ph.D. dissertation, Dept. Comput. Sci., Univ. Alberta, Edmonton, AB, Canada, 2009.

[19]  I-C. Wu, H.-T. Tsai, H.-H. Lin, Y.-S. Lin, C.-M. Chang, and P.-H. Lin, "Temporal difference learning for Connect6," in *Proc. 13th Int. Conf. Adv. Comput. Games (ACG 2011)*, Tilburg, The Netherlands, 2011, pp. 121–133, doi: 10.1007/978-3-642-31866-5_11.

[20]  M. Thill, P. Koch, and W. Konen, "Reinforcement learning with n-tuples on the game Connect-4," in *Proc. 12th Int. Conf. Parallel Problem Solving Nature (PPSN 2012)*, Taormina, Italy, 2012, pp. 184–194, doi: 10.1007/978-3-642-32937-1_19.

[21]  S. M. Lucas, "Learning to play Othello with n-tuple systems," *Australian J. Intell. Inf. Process.*, vol. 9, no. 4, pp. 1–20, Feb. 2008. [Online]. Available: http://repository.essex.ac.uk/3820/1/NTupleOthello.pdf

[22]  R. S. Sutton, "Learning to predict by the methods of temporal differences," *Mach. Learn.*, vol. 3, no. 1, pp. 9–44, Aug. 1988, doi: 10.1023/A:1022633531479.

[23]  K. Matsuzaki, "Developing a 2048 player with backward temporal coherence learning and restart," in *Proc. 15th Int. Conf. Adv. Comput. Games (ACG 2017),* Leiden, The Netherlands, 2017, pp. 176–187.

[24]  H. Guei, T.-H. Wei, and I-C. Wu, "Using 2048-like games as a pedagogical tool for reinforcement learning," presented at the *10th Int. Conf. Comput. Games (CG 2018)*, New Taipei, Taiwan, Jul. 7 – Jul. 9, 2018, in *ICGA J.*, vol. 40, no. 3, pp. 281–293, Mar. 5, 2019, doi: 10.3233/ICG-180062.

[25]  V. Mnih, K. Kavukcuoglu, D. Silver, A. Graves, I. Antonoglou, D. Wierstra, and M. Riedmiller, "Playing Atari with Deep Reinforcement Learning," 2013, *arXiv:1312.5602*.

[26] A. G. Barto and S. Mahadevan, "Recent advances in hierarchical reinforcement learning," *Discrete Event Dyn. Syst.*, vol. 13, no. 1–2, pp. 41–77, Oct. 2003, doi: 10.1023/A:1025696116075. [Online]. Available: https://people.cs.umass.edu/~mahadeva/papers/hrl.pdf

[27] K. Matsuzaki, "Evaluation of multi-staging and weight promotion for game 2048," Kochi Univ. Technol., Kami, Japan, Tech. Rep. KUTBTR2017, Oct. 27, 2017. [Online]. Available: http://hdl.handle.net/10173/1564

[28] I-C. Wu, K.-H. Yeh, C.-C. Liang, C.-C. Chang, and H. Chiang, "Multi-stage temporal difference learning for 2048," in *Proc. 19th Int. Conf. Technol. Appl. Artif. Intell. (TAAI 2014)*, Taipei, Taiwan, 2014, pp. 366–378, doi: 10.1007/978-3-319-13987-6_34.

[29] D. F. Beal and M. C. Smith, "Temporal coherence and prediction decay in TD learning," in *Proc. 16th Int. Joint Conf. Artif. Intell. (IJCAI-99)*, vol. 1., San Mateo, CA, USA, 1999, pp. 564–569. [Online]. Available: http://ijcai.org/Proceedings/99-1/Papers/081.pdf

[30] W. W. Bledsoe and I. Browning, "Pattern recognition and reading by machine," in *Proc. East. Joint Comput. Conf. (IRE-AIEE-ACM '59 Eastern)*, Boston, Massachusetts, 1959, pp. 225–232, doi: 10.1145/1460299.1460326. [Online]. Available: https://dl.acm.org/doi/pdf/10.1145/1460299.1460326

[31] K. Oka and K. Matsuzaki, "Systematic selection of N-tuple networks for 2048," in *Proc. 9th Int. Conf. Comput. Games (CG 2016)*, Leiden, The Netherlands, 2016, pp. 81–92.

[32] J. Austin, RAM-Based Neural Networks, 1st ed., River Edge, NJ, USA: World Scientific, 1998.

[33] J. Austin, "A review of RAM based neural networks," in *Proc. 4th Int. Conf. Microelectron. Neural Networks Fuzzy Syst.*, Turin, Italy, 1994, pp. 58–66, doi: 10.1109/ICMNN.1994.593179. [Online]. Available: https://dl.acm.org/doi/pdf/10.1145/1460299.1460326

[34] H. Guei, T.-H. Wei, J.-B. Huang, and I-C. Wu, "An empirical study on applying deep reinforcement learning to the game 2048," presented at *the Workshop of Neural Network at the 9th Int. Conf. Comput. Games (CG 2016)*, Leiden, The Netherlands, Jun. 29 – Jul. 1, 2016. [Online]. Available: https://docplayer.net/20912615-An-early-attempt-at-applying-deep-reinforcement-learning-to-the-game-2048.html

[35] K. Matsuzaki and M. Teramura, "Interpreting neural-network players for game 2048," in *Proc. 23rd Int. Conf. Technol. Appl. Artif. Intell. (TAAI 2018)*, Taichung, Taiwan, 2018, pp. 136–141, doi: 10.1109/TAAI.2018.00038.

[36] N. Kondo and K. Matsuzaki, "Playing game 2048 with deep convolutional neural networks trained by supervised learning," J. Inf. Process., vol. 27, pp. 340–347, Apr. 2019, doi: 10.2197/IPSJJIP.27.340. [Online]. Available: https://www.jstage.jst.go.jp/article/ipsjjip/27/0/27_340/_pdf/-char/en

[37] K. Matsuzaki, "A further investigation of neural network players for game 2048," in *Proc. 16th Int. Conf. Adv. Comput. Games (ACG 2019)*, Macao, China, 2019, pp. 53–65, doi: 10.1007/978-3-030-65883-0_5. [Online]. Available: https://docplayer.net/157398580-A-further-investigation-of-neural-network-players-for-game-2048.html

[38] S. Watanabe and K. Matsuzaki, "Enhancement of CNN-based 2048 Player with Monte-Carlo Tree Search," presented at *the 27th Int. Conf. Technol. Appl. Artif. Intell. (TAAI 2022)*, Tainan, Taiwan, Dec. 1–3, 2022.

[39] B. W. Ballard, "The *-minimax search procedure for trees containing chance nodes," *Artif. Intell.*, vol. 21, no. 3, pp. 327–350, Sep. 1983, doi: 10.1016/S0004-3702(83)80015-0.

[40] E. Melkó and B. Nagy, "Optimal strategy in games with chance nodes," *Acta Cybern.*, vol. 18, no. 2, pp. 171–192, Jan. 2007, doi: 10.5555/1375766.1375768. [Online]. Available: https://www.researchgate.net/publication/220123091_Optimal_strategy_in_games_with_chance_nodes

[41] N. Pezzotti. "An artificial intelligence for the 2048 game." Mar. 26, 2014. Accessed: Jun. 16, 2020. [Online]. Available: https://diaryofatinker.blogspot.com/2014/03/an-artificial-intelligence-for-2048-game.html

[42] A. L. Zobrist, "A new hashing method with application for game playing," *ICGA J.*, vol. 13, no. 2, pp. 69–73, Jun. 1990, doi: 10.3233/ICG-1990-13203.

[43] A. Appleby. "MurmurHash3 64-bit finalizer." Aug. 1, 2016. Accessed: Jun. 16, 2020. [Online]. Available: https://github.com/kcwu/2048-c

[44] R. Coulom, "Efficient Selectivity and Backup Operators in Monte-Carlo Tree Search," in *Comput. Games: 5th Int. Conf. (CG 2006)*, Turin, Italy, May 29–31, 2006, pp. 72–83, doi: 10.1007/978-3-540-75538-8_7. [Online]. Available: https://hal.inria.fr/file/index/docid/116992/filename/CG2006.pdf

[45] C. B. Browne, E. Powley, D. Whitehouse, S. M. Lucas, P. I. Cowling, P. Rohlfshagen, S. Tavener, D. Perez, S. Samothrakis, and S. Colton, "A survey of monte carlo tree search methods," *IEEE Trans. Comput. Intell. AI Games*, vol. 4, no. 1, pp. 1–43, Mar. 2012, doi: 10.1109/TCIAIG.2012.2186810.

[46] P. Rodgers and J. Levine, "An investigation into 2048 AI strategies," in *Proc. 2014 IEEE Conf. Comput. Intell. Games (CIG 2014)*, Dortmund, Germany, 2014, pp. 1–2, doi: 10.1109/CIG.2014.6932920.

[47] D. Silver, A. Huang, C. J. Maddison, A. Guez, L. Sifre, G. van den Driessche, J. Schrittwieser, I. Antonoglou, V. Panneershelvam, M. Lanctot, S. Dieleman, D. Grewe, J. Nham, N. Kalchbrenner, I. Sutskever, T. Lillicrap, M. Leach, K. Kavukcuoglu, T. Graepel, and D. Hassabis, "Mastering the game of Go with deep neural networks and tree search," *Nature*, vol. 529, no. 7587, pp. 484–489, Jan. 2016, doi: 10.1038/nature16961.

[48] D. Silver, J. Schrittwieser, K. Simonyan, I. Antonoglou, A. Huang, A. Guez, T. Hubert, L. Baker, M. Lai, A. Bolton, Y. Chen, T. Lillicrap, F. Hui, L. Sifre, G. van den Driessche, T. Graepel, and D. Hassabis, "Mastering the game of go without human knowledge," *Nature*, vol. 550, no. 7676, pp. 354–359, Oct. 2017, doi: 10.1038/nature24270.

[49] D. Silver, T. Hubert, J. Schrittwieser, I. Antonoglou, M. Lai, A. Guez, M. Lanctot, L. Sifre, D. Kumaran, T. Graepel, T. Lillicrap, K. Simonyan, and D. Hassabis, "A general reinforcement learning algorithm that masters chess, shogi, and Go through self-play," *Science*, vol. 362, no. 6419, pp. 1140–1144, Dec. 2018, doi: 10.1126/science.aar6404.

[50] C. D. Rosin, "Multi-armed bandits with episode context," *Annu. Math. Artif. Intell.*, vol. 61, no. 3, pp. 203–230, doi: 10.1007/s10472-011-9258-6. [Online]. Available: http://gauss.ececs.uc.edu/Conferences/isaim2010/papers/rosin.pdf

[51] A. van den Oord, O. Vinyals, and K. Kavukcuoglu, "Neural discrete representation learning," in *Proc. 31st Int. Conf. Neural Inf. Process. Syst. (NIPS 2017)*, pp. 6309–6318, Long Beach, California, USA, Dec. 4–9, 2017. [Online]. Available: https://proceedings.neurips.cc/paper/2017/file/7a98af17e63a0ac09ce2e96d03992fbc-Paper.pdf

[52] K.-H. Yeh. "2048 AI." Accessed: Jun. 16, 2020. [Online]. Available: https://github.com/tnmichael309/2048AI

[53] K.-C. Wu. "2048-c." Accessed: Jun. 16, 2020. [Online]. Available: https://github.com/kcwu/2048-c

[54] W. Jaśkowski and A. Szczepański. "2048 AI." Accessed: Jun. 16, 2020. [Online]. Available: https://github.com/aszczepanski/2048

[55] H. Guei. "moporgic/TDL2048+." Accessed: Apr. 8, 2021. [Online]. Available: https://github.com/moporgic/TDL2048

[56] H. Guei, L.-P. Chen, and I-C. Wu, "Optimistic Temporal Difference Learning for 2048," *IEEE Trans. Games*, vol. 14, no. 3, pp. 478–487, Sep. 2022, doi: 10.1109/TG.2021.3109887. [Online]. Available: *arXiv:2111.11090*.

[57] M. C. Machado, S. Srinivasan, and M. Bowling, "Domain-independent optimistic initialization for reinforcement learning," in *Workshop of Learn. General Compet. Video Games at the 29th AAAI Conf. Artif. Intell. (AAAI-15)*, Pittsburgh, PA, USA, 2015. [Online]. Available: *arXiv:1410.4604*.

[58] I. Szita and A. Lőrincz, "The many faces of optimism: a unifying approach," in *Proc. 25th Int. Conf. Mach. Learn. (ICML 2008)*, Helsinki, Finland, 2008, pp. 1048–1055, doi: 10.1145/1390156.1390288. [Online]. Available: https://dl.acm.org/doi/pdf/10.1145/1390156.1390288

[59] V. Kuleshov and D. Precup, "Algorithms for multi-armed bandit problems," 2014, *arXiv:1402.6028*.

[60] E. Even-Dar and Y. Mansour, "Convergence of optimistic and incremental Q-learning," in *Proc. 14th Int. Conf. Neural Inf. Process. Syst. (NIPS 2001)*, Vancouver, BC, Canada, 2001, pp. 1499–1506. [Online]. Available: http://papers.nips.cc/paper/1944-convergence-of-optimistic-and-incremental-q-learning.pdf

[61] N. K. Jong, T. Hester, and P. Stone, "The utility of temporal abstraction in reinforcement learning," in *Proc. 7th Int. Joint Conf. Auton. Agents Multiagent Syst. (AAMAS 2008)*, Estoril, Portugal, 2008, pp. 299–306, doi: 10.5555/1402383.1402429. [Online]. Available: https://dl.acm.org/doi/pdf/10.5555/1402383.1402429

[62] S. J. Russell and P. Norvig, *Artificial Intelligence: A Modern Approach*, 3rd ed., Malaysia: Pearson Education Limited, 2010.

[63] O. Sagi and L. Rokach, "Ensemble learning: A survey," Wiley Interdisciplinary Reviews: Data Mining and Knowledge Discovery, vol. 8, no. 4, pp. e1249, Aug. 2018, doi:

10.1002/widm.1249. [Online]. Available: https://wires.onlinelibrary.wiley.com/doi/epdf/10.1002/widm.1249

[64] P. Izmailov, D. Podoprikhin, T. Garipov, D. Vetrov, and A. G. Wilson, "Averaging weights leads to wider optima and better generalization," in *Proc. 34th Conf. Uncertainty Artif. Intell. (UAI 2018)*, California, USA, 2018, pp. 876–885. [Online]. Available: http://auai.org/uai2018/proceedings/papers/313.pdf

[65] D. J. Wu, "Accelerating self-play learning in go," presented at *the Workshop on Reinforcement Learning in Games at the 34th AAAI Conf. Artif. Intell. (AAAI-20)*, New York, USA, Feb. 7–12, 2020. [Online]. Available: *arXiv:1902.10565*.

[66] T. Garipov, P. Izmailov, D. Podoprikhin, D. P. Vetrov, and A. G. Wilson, "Loss surfaces, mode connectivity, and fast ensembling of DNNs," in *Proc. 32nd Int. Conf. Neural Inf. Process. Syst. (NeurIPS 2018)*, pp. 8803–8812, Montreal, Canada, 2018. [Online]. Available: https://papers.nips.cc/paper/2018/file/be3087e74e9100d4bc4c6268cdbe8456-Paper.pdf

[67] L. N. Smith, "Cyclical learning rates for training neural networks," in *2017 IEEE Winter Conf. Appl. Comput. Vision (WACV 2017)*, California, USA, Mar. 24–31, 2017, pp. 464–472, doi: 10.1109/WACV.2017.58. [Online]. Available: *arXiv:1506.01186*.

[68] I-C. Wu, T.-R. Wu, A.-J. Liu, H. Guei, and T. Wei, "On strength adjustment for MCTS-based programs," in *Proc. AAAI Conf. Artif. Intell. (AAAI-19)*, vol. 33, no. 01, Honolulu, Hawaii, USA, pp. 1222–1229, doi: 10.1609/aaai.v33i01.33011222. [Online]. Available: https://ojs.aaai.org/index.php/AAAI/article/view/3917/3795

[69] A.-J. Liu, T.-R. Wu, I-C. Wu, H. Guei, and T.-H. Wei, "Strength Adjustment and Assessment for MCTS-Based Programs," *IEEE Comput. Intell. Mag.*, vol. 15, no. 3, pp. 60–73, Aug. 2020, doi: 10.1109/MCI.2020.2998315.

[70] I-C. Wu, T.-R. Wu, A.-J. Liu, H. Guei, and T.-H. Wei, "Method for adjusting the strength of turn-based game automatically," U.S. Patent US11247128B2, Feb. 22, 2022.

[71] I-C. Wu, T.-R. Wu, A.-J. Liu, H. Guei, and T.-H. Wei, "Method for automatically modifying strength of turn based game," R.O.C. (Taiwan) Patent TWI725662B, Apr. 21, 2021.

[72] Y. Tian, J. Ma, Q. Gong, S. Sengupta, Z. Chen, J. Pinkerton, and L. Zitnick, "Elf opengo: An analysis and open reimplementation of alphazero," in *Proc. 36th Int. Conf. Mach. Learn. (ICML 2019)*, California, USA, Jun. 10–15, 2019, pp. 6244–6253. [Online]. Available: *arXiv:1902.04522*.

[73] I. Danihelka, A. Guez, J. Schrittwieser, and D. Silver, "Policy improvement by planning with Gumbel". in *the 10th Int. Conf. Learn. Representations (ICLR 2022)*, Virtual, Apr. 25–29, 2022. [Online]. Available: https://openreview.net/pdf?id=bERaNdoegnO

[74] J. Downey. "Evolving Neural Networks to Play 2048." (May 12, 2014). Accessed: May 15, 2016. [Online Video]. Available: https://www.youtube.com/watch?v=jsVnuw5Bv0s

[75] T. Weston. "2048-Deep-Learning." Accessed: May 15, 2016. [Online]. Available: https://github.com/anubisthejackle/2048-Deep-Learning

[76] M. D. Zeiler, "Adadelta: an adaptive learning rate method," 2012, *arXiv:1212.5701*.

[77] Y. Jia, E. Shelhamer, J. Donahue, S. Karayev, J. Long, R. Girshick, S. Guadarrama, and T. Darrell, "Caffe: Convolutional architecture for fast feature embedding," in *Proc. 22nd ACM Int. Conf. Multimedia (MM'14)*, Florida, USA, 2014, pp. 675–678, doi: 10.1145/2647868.2654889. [Online]. Available: *arXiv:1408.5093*.

[78] W. Kool, H. Van Hoof, and M. Welling, "Stochastic beams and where to find them: The gumbel-top-k trick for sampling sequences without replacement," in *Proc. 36th Int. Conf. Mach. Learn. (ICML 2019)*, vol. 97, no. 1, pp. 3499–3508, Long Beach, California, USA, Jun. 10–15, 2019, [Online]. Available: http://proceedings.mlr.press/v97/kool19a/kool19a.pdf

[79] C.-Y. Kao, H. Guei, T.-R. Wu, and I-C. Wu, "Gumbel MuZero for the Game of 2048," presented at *the 27th Int. Conf. Technol. Appl. Artif. Intell. (TAAI 2022)*, Tainan, Taiwan, Dec. 1–3, 2022.

[80] Z. Karnin, T. Koren, and O. Somekh, "Almost optimal exploration in multi-armed bandits," in *Proc. 30th Int. Conf. Mach. Learn. (ICML 2013)*, vol. 28, no. 1, pp. 1238–1246, Atlanta, USA, Jun. 16–21, 2013.

[81] Y.-H. Cheng, "The Design of a MuZero Framework," M.S. thesis, Institute of Computer Science and Engineering, National Chiao Tung University, 2021.

[82] C.-Y. Kao, "Gumbel MuZero for the Game of 2048," M.S. thesis, Institute of Computer Science and Engineering, National Yang Ming Chiao Tung University, 2022.

[83] T.-J. Wei. "A Deep Learning AI for 2048." Accessed: Oct. 17, 2019. [Online]. Available: https://github.com/tjwei/2048-NN

[84] M. Samir. "2048 Deep Recurrent Reinforcement Learning." Accessed: Oct. 29, 2021. [Online]. Available: https://github.com/Mostafa-Samir/2048-RL-DRQN

[85] N. Virdee. "Trained A Convolutional Neural Network To Play 2048 using Deep-Reinforcement Learning." Accessed: Oct. 17, 2019. [Online]. Available: https://github.com/navjindervirdee/2048-deep-reinforcement-learning

[86] G. Wiese. "2048 Reinforcement Learning." Accessed: Oct. 17, 2019. [Online]. Available: https://github.com/georgwiese/2048-rl

[87] H. Guei, T.-H. Wei, and I-C. Wu, "Teaching Reinforcement Learning and Computer Games with 2048-Like Games," in *Proc. Annu. Conf. JSAI 33rd (JSAI 2019)*, Niigata, Japen, 2019, pp. 2J1E501–2J1E501, doi: 10.11517/pjsai.JSAI2019.0_2J1E501. [Online]. Available: https://www.jstage.jst.go.jp/article/pjsai/JSAI2019/0/JSAI2019_2J1E501/_pdf/-char/en

[88] Lazy Programmer Inc. "Artificial Intelligence: Reinforcement Learning in Python." Accessed: Oct. 17, 2019. [Online]. Available: https://www.udemy.com/course/artificial-intelligence-reinforcement-learning-in-python

[89] T. Simonini. "A Free course in Deep Reinforcement Learning from beginner to expert." Accessed: Oct. 17, 2019. [Online]. Available: https://simoninithomas.github.io/Deep_reinforcement_learning_Course

[90] The University of Alberta. "Reinforcement Learning Specialization." Accessed: Oct. 17, 2019. [Online]. Available: https://www.coursera.org/specializations/reinforcement-learning

[91] C. Archibald. "RISK AI Project." Accessed: Oct. 17, 2019. [Online]. Available: http://modelai.gettysburg.edu/2019/risk

[92] Packt Publishing. "Implementing AI to Play Games." Accessed: Oct. 17, 2019. [Online]. Available: https://www.udemy.com/course/implementing-ai-to-play-games

[93] P. Talaga. "Using Ultimate Tic Tac Toe to Motivate AI Game Agents." Accessed: Oct. 17, 2019. [Online]. Available: http://modelai.gettysburg.edu/2019/ultimatettt

[94] L. Lu, Y. Shin, Y. Su, and G. E. Karniadakis, "Dying relu and initialization: Theory and numerical examples," 2019, *arXiv:1903.06733*.

[95] A. L. Maas, A. Y. Hannun, and A. Y. Ng, "Rectifier nonlinearities improve neural network acoustic models," in *Proc. 30th Int. Conf. Mach. Learning (ICML 2013)*, vol. 30, no. 1, pp. 3, Atlanta, Georgia, USA, Jun. 2013, doi: 10.1.1.693.1422. [Online]. Available: http://robotics.stanford.edu/~amaas/papers/relu_hybrid_icml2013_final.pdf

# Appendix

## A  TC Converges With Insufficient Exploration

TC learning performs well on $n$-tuple networks for 2048. However, before this work, its effectiveness had only been confirmed mainly with smaller networks, e.g., the 5×6-tuple [5]. After Matsuzaki [8] proposed the designs of larger networks, we observed that TC learning with the 8×6-tuple resulted in unreliable learning performance caused by insufficient exploration.

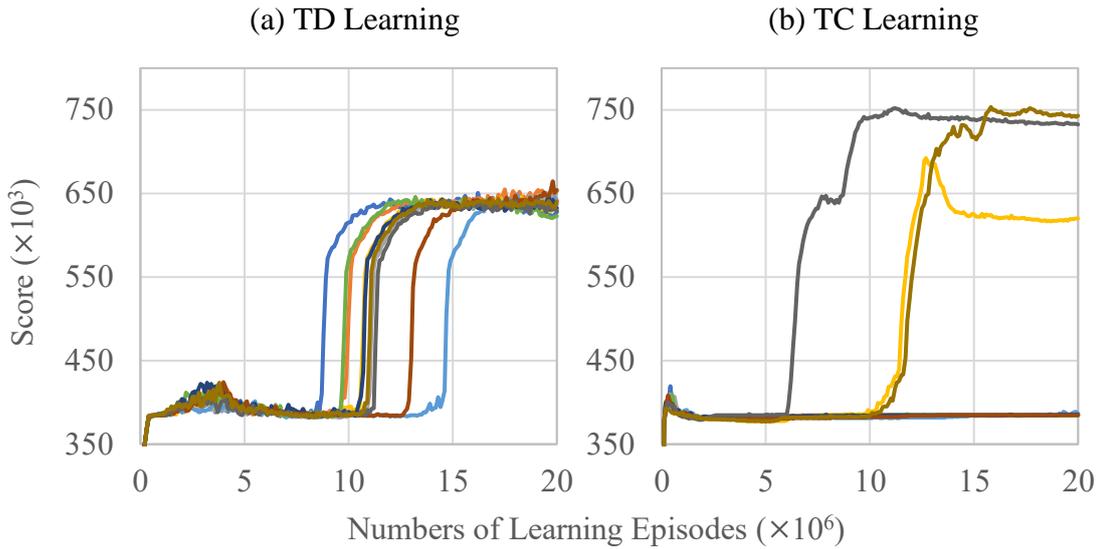

Figure 40. Maximum scores of TD and TC learning in the 2-stage 8×6-tuple network.

Figure 40 shows the learning processes of using TD and TC learning for training the 2nd stage of a 2-stage 8×6-tuple network. The 2nd stage started at states with 16384-tiles, and the goal is to achieve 32768-tiles. For statistical correctness, We run 10 trials for each method using different random seeds. TD learning used a fixed learning rate $\alpha = 0.1$; and TC learning used an initial learning rate $\alpha = 1.0$. As in the figure, in all 10 trials, TD learning eventually found solutions for reaching 32768-tiles, i.e., reaching a maximum score larger than 400k points. However, TC learning got stuck in local optimums in 7 out of 10 trials; only 3 achieved 32768-tiles. Since TC learning has an adaptive learning rate adjustment, it may learn better once it reaches 32768-tiles. However, TC learning converges very fast; likely, it cannot find a solution to 32768-tiles before the convergence.

# B  Negative Afterstate Values and Workarounds

Expectimax search is adopted to improve performance further. However, we noticed that as the expectimax search tree deepens, more and more leaf node values become negative. In extreme cases, more than 50% of leaf nodes had negative values, as shown in Figure 41. These negative values can seriously interfere with the search.

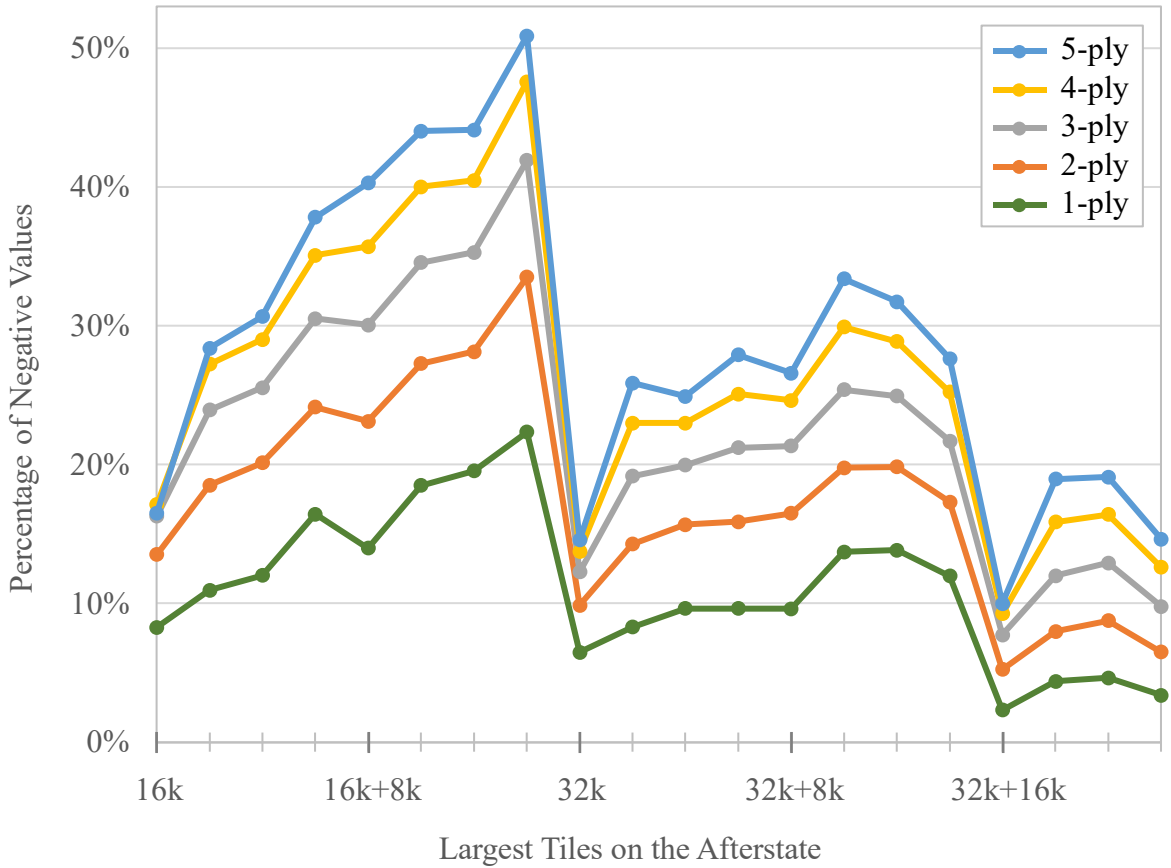

Figure 41. Percentages of negative afterstate values at different depths in an expectimax search tree.
Each horizontal axis tick represents an addition of 2048-tile.

A practical example of an interfered search is shown in Figure 42. In this example, the optimal action should be "up"; however, the leaf afterstate values of this branch were all negative. On the other hand, the action "down" was assigned a value of 0 as its successors were all terminal states. Finally, the expectimax search misidentified the action "down" as the best action, which caused the game to finish immediately after this erroneous decision.

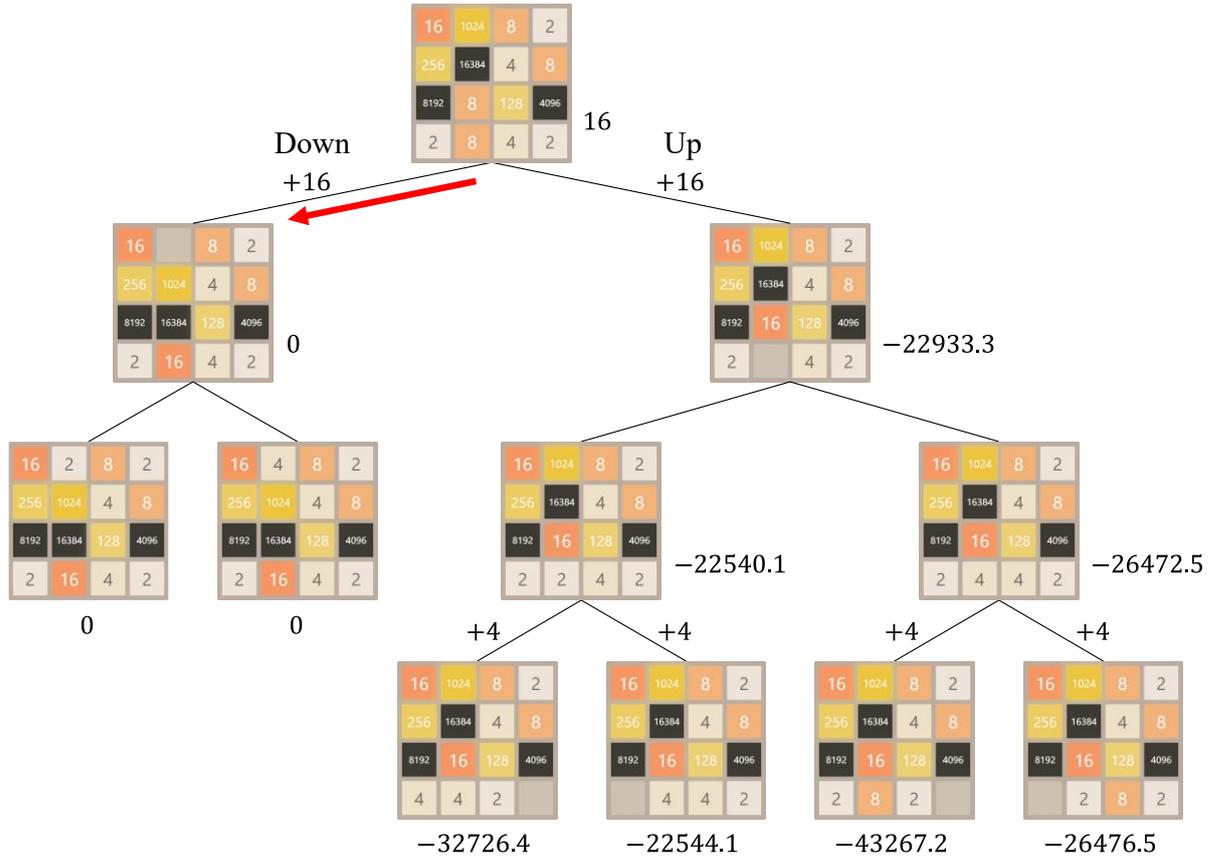

Figure 42. Example of an interfered search result caused by negative values. The optimal action for the root state should be "up"; however, the search tree misidentifies "down" as the optimal action. Note that terminal states are explicitly assigned 0 in the search tree, which causes action "down" to be assumed better in this example.

## B.1 The Cause of Negative Values

First, we analyze the cause of negative afterstate values in the $n$-tuple networks. At the beginning phase of an episode, an afterstate contains only *promising features*, namely features with only smaller tiles. As the TD target in the beginning phase is intrinsically higher, the promising feature weights are continuously increased during the training. On the other hand, at the ending phase of an episode, an afterstate may contain *unpromising features*, namely features with larger tiles. However, it may also contain promising features with higher feature weights, which higher the value of this afterstate. As the TD target in the ending phase is close to zero, the corresponding feature weights, for both promising and unpromising features, are adjusted with a negative TD error.

In short, the promising feature weights are increased in the beginning phase and decreased in the ending phase. However, the unpromising feature weights are only decreased in the ending phase. This procedure repeats during the training and exacerbates the extreme values. Finally, a negative afterstate value occurs when an afterstate contains more unpromising features than the average. Furthermore, negative afterstate values propagate to previous afterstates, resulting in even more negative afterstate values. Therefore, when using the original TD learning algorithm and the $n$-tuple networks, the negative afterstate values are inevitable. Table 40 shows extreme feature weights from an 8×6-tuple network trained by the optimistic method. As these negative values occur intrinsically during training $n$-tuple networks, we present two rectifications to mitigate the interference from extreme afterstate values.

Table 40. Examples of extreme feature weights in a well-trained 8×6-tuple network.

| Feature | Feature Weight |
|---|---|
| 16, 2, 0, 0 / 0, 2 | 703657 |
| 64, 2, 0, 0 / 0, 2 | 445896 |
| 32, 2, 0, 0 / 0, 2 | 443362 |
| 128, 2, 0, 0 / 0, 2 | 415132 |
| 256, 2, 0, 0 / 0, 2 | 379937 |
| 4, 0, 4, 16384 / 2, 2 | -575483 |
| 4, 0, 16, 16384 / 2, 2 | -575633 |
| 4, 0, 8192, 32768 / 2, 2 | -581457 |
| 4, 0, 8192, 16384 / 2, 2 | -582586 |
| 4, 0, 32768, 16384 / 2, 2 | -583979 |

The features correspond to the 6-tuple, as illustrated in Figure 5 (c).

## B.2 Rectification for Existing Networks

To rectify a trained $n$-tuple network that already with negative values, a straightforward workaround is to add an offset to the terminal state value, i.e., assign a negative value to the terminal state instead of 0. However, the negative afterstate values vary from different states to different search depths; there is no way to add a constant offset to the terminal state value.

Therefore, we rectify the negative values by using the modified expectimax search procedure $V'_{max}$ and $V'_{chance}$ as

$$V'_{max}(s_t, p) = \begin{cases} \max_{\forall a_t}\{r_t + V'_{chance}(s'_t, p-1)\} + 1, & \text{if } s_t \text{ is not terminal} \\ 0, & \text{if } s_t \text{ is terminal,} \end{cases} \quad (39)$$

$$V'_{chance}(s'_t, p) = \begin{cases} \mathbb{E}[V'(s_{t+1})], & \text{if } p > 0 \\ \max(V(s'_t), 0), & \text{if } p = 0, \end{cases} \quad (40)$$

$$\mathbb{E}[V'(s_{t+1})] = \sum_{s_{t+1}} \mathcal{P}(s'_t, s_{t+1}) V'_{max}(s_{t+1}, p). \quad (41)$$

In the modified version, all negative values from the TD afterstate value function $V(s'_t)$ are rectified with 0, and a constant step reward "+1" is added to nonterminal states, which allows the search tree to choose a deeper branch, especially when all branches are rectified. Note that the search result is not affected if all branches are fully expanded to the search depth limit without reaching terminal states or negative afterstates.

## B.3 Rectification for Training Networks

This section proposes changes to the learning algorithm to mitigate negative values during training. Before presenting the proposed changes, let us review the TD(0) afterstate learning framework with $n$-tuple network introduced in Chapter 2. For an $m \times n$-tuple network, there are $m$ different $n$-tuples $\phi_1, \ldots, \phi_m$ with their corresponding lookup tables $\text{LUT}_1, \ldots, \text{LUT}_m$. Given an afterstate $s'_t$, its estimation $V(s'_t)$ is calculated by summing all of the $m$ feature weights $\text{LUT}_i[\phi_i(s'_t)]$ as

$$V(s'_t) = \sum_{i=1}^{m} \text{LUT}_i[\phi_i(s'_t)]. \quad (42)$$

When adjusting $V(s'_t)$ by a TD error $\delta_t = r_{t+1} + V(s'_{t+1}) - V(s'_t)$, the adjustment is equally distributed to $m$ feature weights $\text{LUT}_i[\phi_i(s'_t)]$ with a learning rate $\alpha$ by

$$\text{LUT}_i[\phi_i(s'_t)] \leftarrow \text{LUT}_i[\phi_i(s'_t)] + \frac{\alpha}{m}(r_{t+1} + V(s'_{t+1}) - V(s'_t)). \quad (43)$$

*Rectified Linear Unit (ReLU)* is a common activation function for deep neural networks, which has been widely used in many works, e.g., [25], [35]–[38], [47]–[73]. ReLU uses a ramp function to rectify all negative values, i.e., $\max(x, 0)$, which is exactly what is needed. Unlike other activation functions such as sigmoid or hyperbolic tangent, ReLU is unbounded and behaves the same as the original linear function used by $n$-tuple network when values are positive. Therefore, ReLU can be easily combined with TD learning and $n$-tuple network without additional modifications to the network output range, by

$$V_{\text{ReLU}}(s'_t) = \begin{cases} V(s'_t), & \text{if } V(s'_t) \geq 0 \\ 0, & \text{if } V(s'_t) < 0, \end{cases} \qquad (44)$$

where $V_{\text{ReLU}}$ is the rectified value function, and $V(s'_t) = \sum_{i=1}^{m} \text{LUT}_i[\phi_i(s'_t)]$ is the original afterstate value function. For adjusting the feature weights, the original training algorithm in (43) is slightly rewritten as

$$\text{LUT}_i[\phi_i(s'_t)] \leftarrow \text{LUT}_i[\phi_i(s'_t)] + \frac{\alpha}{m}\left(r_{t+1} + V_{\text{ReLU}}(s'_{t+1}) - V(s'_t)\right). \qquad (45)$$

This modified version is named the *rectified TD learning*. In the rectified TD learning, the TD target is the rectified afterstate value $V_{\text{ReLU}}(s'_{t+1})$, which prevents the propagation of negative values (if present) from the next timestep $t+1$ to the current timestep $t$. Furthermore, a negative afterstate value $V(s'_t)$ can be recovered as its TD target is not less than 0. Note that the negative value is not recovered with the original TD error as in (43) since $V(s'_{t+1})$ is likely to be negative when $V(s'_t)$ is already negative.

Note that for the ReLU activation function, the gradient becomes 0 for negative afterstate values, i.e., the feature weights should not be adjusted if $V(s'_t)$ is already negative. Therefore, the feature weights are supposed to be adjusted as

$$\text{LUT}_i[\phi_i(s'_t)] \leftarrow \begin{cases} \text{LUT}_i[\phi_i(s'_t)] + \frac{\alpha}{m}\left(r_{t+1} + V_{\text{ReLU}}(s'_{t+1}) - V_{\text{ReLU}}(s'_t)\right), & \text{if } V(s'_t) > 0 \\ \text{LUT}_i[\phi_i(s'_t)], & \text{if } V(s'_t) < 0. \end{cases} \qquad (46)$$

The modified version using (46) for training is named the *ReLU TD learning*. Using the ReLU TD is much more dangerous due to the potential *dying ReLU problem* [94], in which the $n$-tuple networks updated with (46) were corrupted by being stuck in perpetual situations where $V(s'_t) < 0$ for most input. Other ReLU variants, e.g., *Leaky ReLU* [95], may mitigate the dying ReLU problem, but they do not rectify negative values precisely as desired. Therefore, we assume that using the rectified TD learning with (45) for $n$-tuple networks is more promising.

Preliminarily experiments were performed to evaluate the effectiveness of the rectified TD learning and the ReLU TD learning in Yeh's 4×6-tuple (Figure 4). Ten networks were trained for each method for statistical correctness. Each network was trained using 10M episodes. Each search depth was assessed using 1M episodes. As shown in Table 41, both rectified TD and ReLU TD learning performed comparably to the original TD learning. However, as the ReLU TD does not recover the values, the values may become extremely negative after training, which could eventually cause dying ReLU problems. Furthermore, although rectified TD learning is expected to have advantages when the search deepens, it is currently too expensive to run enough episodes with deep searches for statistical conclusions. Therefore, it is left open.

Table 41. Comparison of different rectification methods in the 4×6-tuple network.

| Method | Search Depth | | Average of the Lowest Values | | |
|---|---|---|---|---|---|
| | 1-Ply | 2-Ply | Lowest | Lowest 1‰ | Lowest 1% |
| Original TD | 180.7 ± 2 | 278.4 ± 1.6 | -74.3 ± 7.1 | 7.3 ± 0.2 | 33.6 ± 0.2 |
| Rectified TD | **181.5 ± 1.8** | **280.2 ± 2.3** | **-55.3 ± 4.1** | **12.1 ± 0.3** | **36.5 ± 0.2** |
| ReLU TD | 180.9 ± 3.9 | 276.7 ± 1.1 | -103.5 ± 5.9 | 3.1 ± 0.6 | 31.6 ± 0.6 |
| Original TC | **248.1 ± 1.4** | 308.4 ± 1.8 | -119.7 ± 11.6 | 1.4 ± 0.2 | 28.9 ± 0.3 |
| Rectified TC | 247.3 ± 1.6 | **310.1 ± 2.1** | **-82.7 ± 11.6** | **10.3 ± 0.1** | **38.3 ± 0.2** |
| ReLU TC | **248.1 ± 1.5** | 307.9 ± 1.7 | -210 ± 20.4 | -8.8 ± 0.3 | 22.9 ± 0.7 |

The unit is a thousand ($\times 10^3$). The search depth results are the average of the average scores; The average of the lowest values is the average of the average feature weights.

# C  Comparison of Exploration Techniques

The $\epsilon$-*greedy* and the *softmax* are standard exploration techniques. According to previous reports, these exploration techniques significantly inhibited the learning performance for 2048 on $n$-tuple networks [2], [5]. However, previous works have not evaluated these techniques on larger networks such as the 8×6-tuple. In order to compare different exploration techniques on the 8×6-tuple network, we used the TD+TC paradigm as the base learning algorithm, in which the exploration technique was only applied in the TD phase: (i) OI: $V_{init}$ = 0 or 320k; (ii) $\epsilon$-greedy: $\epsilon_{init}$ = 0.01 or 0.1; (iii) softmax: $T_{init}$ = 0.1 or 1. The standard TD+TD with $V_{init}$ = 0 was not only the baseline but also a proposed exploration technique. The exploration level of $\epsilon$-greedy and softmax was reduced every 1M training episodes, linearly starting from the initial value ($\epsilon_{init}$ or $T_{init}$) to zero. The learning used a total of 100M episodes with $\alpha$ = 0.1 and $P_{TC}$ = 10%; each configuration was tested twice.

Table 42. Performance of TD+TC learning with different exploration techniques in the 8×6-tuple network.

| Method | Average Score | 8192 [%] | 16384 [%] | 32768 [%] |
|---|---|---|---|---|
| OI, $V_{init}$ = 0 | 362 922 ± 745 | 97.06% ± 0.01% | **85.73% ± 0.35%** | 20.89% ± 0.29% |
| OI, $V_{init}$ = 320k | **370 874 ± 206** | **97.11% ± 0.13%** | 85.51% ± 0.11% | **23.77% ± 0.29%** |
| $\epsilon$-greedy, $\epsilon_{init}$ = 0.01 | 257 531 ± 656 | 93.86% ± 0.10% | 74.08% ± 0.23% | 0.00% ± 0.00% |
| $\epsilon$-greedy, $\epsilon_{init}$ = 0.1 | 221 984 ± 2773 | 89.27% ± 0.23% | 59.09% ± 1.46% | 0.00% ± 0.00% |
| Softmax, $T_{init}$ = 0.1 | 251 636 ± 122 | 92.85% ± 0.05% | 71.81% ± 0.06% | 0.00% ± 0.00% |
| Softmax, $T_{init}$ = 1 | 230 371 ± 5047 | 84.67% ± 0.19% | 57.57% ± 2.54% | 0.00% ± 0.00% |

Table 42 summarizes the results of applying TD+TC with these exploration techniques. Figure 43 and Figure 44 show the average and maximum scores during training, respectively. Again, OTD+TC performed the best among all exploration techniques. In the learning processes, as shown in Figure 43, learning with $\epsilon$-greedy or softmax completely inhibited the learning ability. By comparing two exploration levels within each method, we also observed that more exploration in $\epsilon$-greedy and softmax further decreased the learning performance. To summarize in short, our experiments confirmed again that $\epsilon$-greedy and softmax do not work well for the game of 2048, at least for the 8×6-tuple network.

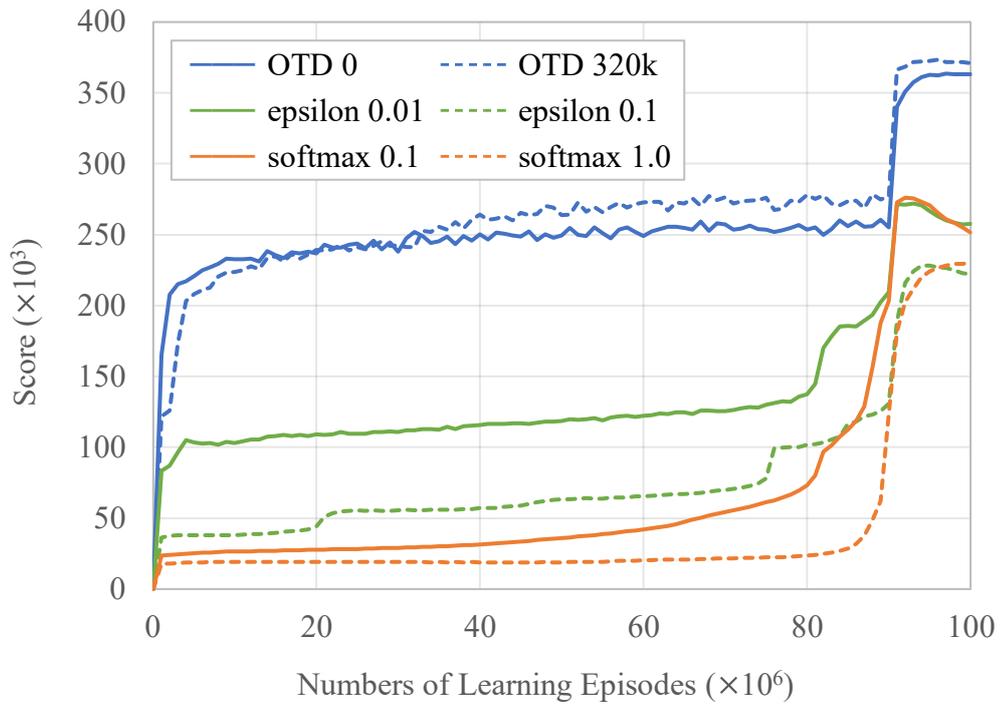

Figure 43. Average scores of TD+TC learning with different exploration techniques in the 8×6-tuple network.

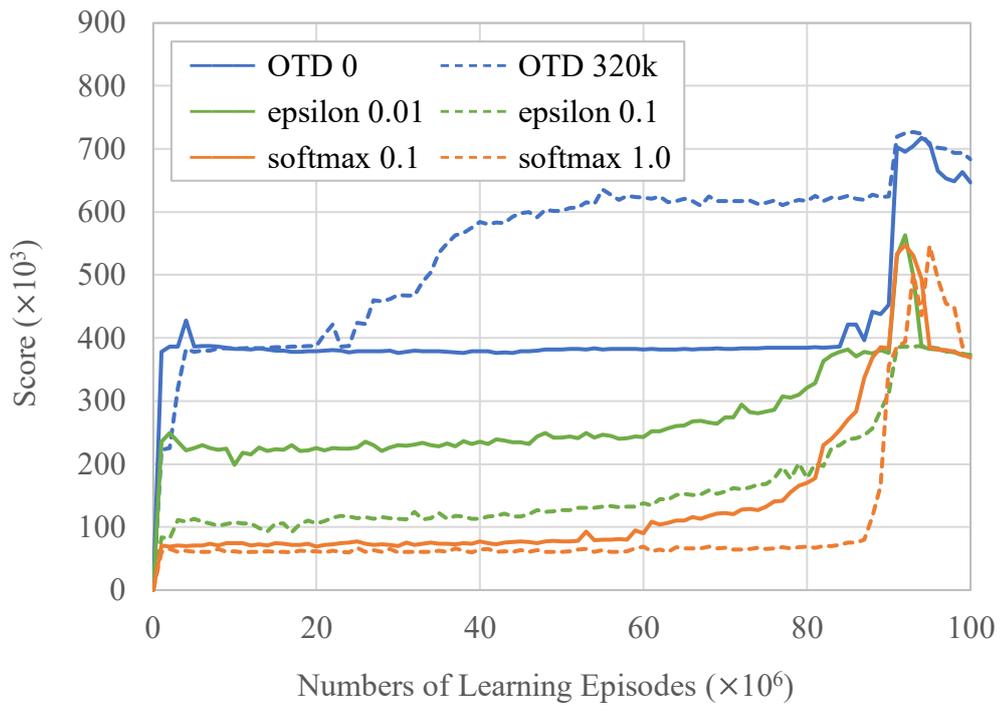

Figure 44. Maximum scores of TD+TC learning with different exploration techniques in the 8×6-tuple network.

# D  Alternatives to the Proposed Course

Various courses on reinforcement learning or computer game algorithms are already available. However, most of them focus on either just the former [88]–[90] or the latter [91]–[93]. In this section, we summarize the characteristics of these popular alternatives, and also provide a brief distinction between these alternatives and the proposed course.

Courses on reinforcement learning usually cover a wide range of RL methods, such as TD, Q-learning, SARSA, Policy Gradients, A2C, A3C, etc. These courses typically provide students with small tasks, such as implementing video game AI bots, to compare the benefits of these different RL methods. Take some available alternatives as examples. "*Artificial Intelligence: Reinforcement Learning in Python*" is a complete guide to AI and RL, which allows students to implement 17 different RL algorithms [88]. "*A Free course in Deep Reinforcement Learning from beginner to expert*" guides students to implement some most popular deep reinforcement algorithms on various video games [89]. "*Reinforcement Learning Specialization*" is a collection of four RL courses that helps students to master the concepts of RL and various RL algorithms, and understand how to apply AI to solve real-world problems [90].

Courses on computer game algorithms mainly focus on search techniques and their variants, such as minimax search, alpha-beta search, MCTS, etc. They may also include other AI-related topics such as decision trees, Bayesian classifiers, evolutionary algorithms, or even basic NN tutorials. For example, "*RISK AI Project*" uses the board game RISK as a topic to teach some major search methods and decision trees. It also comes with a final tournament like the proposed courses [91]. "*Implementing AI to Play Games*" also comes with popular search techniques; it also covers constraint satisfaction problems (CSP) and evolutionary algorithms [92]. "*Ultimate Tic Tac Toe AI Implementation*" is a complete project that teaches students different search strategies by extending a Tic Tac Toe program [93].

More specifically, these courses focus on their areas of expertise, and they may allow students to learn more about either reinforcement learning or computer game algorithms. They may provide tutorials for various algorithms, but not as complete assignments as the projects listed in this dissertation.

# E  Open-Source Materials

This section introduces the codes developed for this dissertation, including *TDL2048+*, *TDL2048-Demo*, and *2048-Framework*. All of them are open-source.

TDL2048+[10] is a highly optimized framework of temporal difference learning and $n$-tuple networks for 2048. It supports major algorithms related to 2048, e.g., TD(0), TD($\lambda$), $n$-step TD, TC, multistage training, optimistic initialization, ensemble learning, carousal shaping, restart, tile-downgrading, lock-free parallelism, expectimax search, and MCTS. Furthermore, it is the most efficient implementation for 2048. In this dissertation, the reinforcement learning methods related to $n$-tuple network were developed based on TDL2048+.

TDL2048-Demo[11] is an example of temporal difference learning, mainly for beginners to learn how reinforcement learning works in this game. Its lightweight design has less than 1000 lines of code, including a standard TD(0) algorithm, a simple bitboard, and a configurable $n$-tuple network. The material for the lightweight course introduced in Section 5.3 is modified from TDL2048-Demo by removing implementation related to TD(0).

2048-Framework[12] is the framework for 2048-like games. For development, it provides standard procedures for game-playing, statistics, and input/output. It also supports extending 2048-like games to two-player games. This framework does not include reinforcement learning algorithms and other specific game-playing algorithms. Therefore, it is used as the material for the comprehensive courses as introduced in Section 5.4.

---

[10] TDL2048+ is available at https://github.com/moporgic/TDL2048

[11] TDL2048-Demo is available at https://github.com/moporgic/TDL2048-Demo
 The material for the lightweight course is available at
 https://cgilab.nctu.edu.tw/~shared_data/2048/tutorial/2048-tutorial-latest.zip

[12] 2048-Framework has both C++ and Python implementations:
 C++ code available at https://github.com/moporgic/2048-Framework
 Python code available at https://github.com/moporgic/2048-Framework-Python

# Vita

Hung Guei was born in Taoyuan, Taiwan, in 1993. He received a B.S. degree in Computer Science at the Department of Computer Science and Information Engineering of National Central University, Taoyuan, Taiwan, in 2015. In the same year, he began to study for an M.S. degree in Computer Science at the Institute of Computer Science and Engineering of National Chiao Tung University[13], Hsinchu, Taiwan. Subsequently, he applied for direct admission to the Ph.D. degree in 2016. His research interests include computer games, artificial intelligence, reinforcement learning, machine learning, and distributed computing.

---

[13] Has been merged into National Yang Ming Chiao Tung University in February 2021.